\documentclass[numbers]{article}
\usepackage[preprint]{nips_2019}

\usepackage[utf8]{inputenc} 
\usepackage[T1]{fontenc}    
\usepackage{hyperref}       
\usepackage{url}            
\usepackage{booktabs}       
\usepackage{amsfonts}       
\usepackage{nicefrac}       
\usepackage{microtype}      
\usepackage{authblk}
\usepackage{graphicx}
\usepackage{graphics}
\usepackage{verbatim}
\usepackage{float}
\usepackage{lipsum}
\restylefloat{table}
\usepackage{float}
\usepackage{siunitx}

\usepackage[resetlabels]{multibib}
\newcites{Manuscript}{References}
\newcites{OnlineMethods}{References}
\newcites{Supplement}{References}

\newcommand\blfootnote[1]{%
  \begingroup
  \renewcommand\thefootnote{}\footnote{#1}%
  \addtocounter{footnote}{-1}%
  \endgroup
}

\title{Automated Design of Deep Learning Methods for Biomedical Image Segmentation}

\author[1,2$\dagger$] {\textbf{Fabian Isensee}}
\author[1$\dagger$] {\textbf{Paul F. Jaeger}}
\author[3$\ddagger$] {\textbf{Simon A. A. Kohl}}
\author[1,4] {\textbf{Jens Petersen}}
\author[1,5*] {\textbf{Klaus H. Maier-Hein}}

\affil [1] {Division of Medical Image Computing, German Cancer Research Center, Heidelberg}
\affil [2] {Faculty of Biosciences, University of Heidelberg, Heidelberg, Germany}
\affil [3] {DeepMind, London, United Kingdom}
\affil [4] {Faculty of Physics \&\ Astronomy, University of Heidelberg, Heidelberg, Germany}
\affil [5] {Pattern Analysis and Learning Group, Heidelberg University Hospital, Department of Radiation Oncology, Heidelberg, Germany}
\affil[*]{\texttt {k.maier-hein@dkfz.de} }

\begin{document}
\fontsize{10}{14}\selectfont
\maketitle
\begin{abstract}
Biomedical imaging is a driver of scientific discovery and core component of medical care, currently stimulated by the field of deep learning. While semantic segmentation algorithms enable 3D image analysis and quantification in many applications, the design of respective specialised solutions is non-trivial and highly dependent on dataset properties and hardware conditions. We propose nnU-Net, a deep learning framework that condenses the current domain knowledge and autonomously takes the key decisions required to transfer a basic architecture to different datasets and segmentation tasks. Without manual tuning, nnU-Net surpasses most specialised deep learning pipelines in 19 public international competitions and sets a new state of the art in the majority of the 49 tasks. The results demonstrate a vast hidden potential in the systematic adaptation of deep learning methods to different datasets. We make nnU-Net publicly available as an open-source tool that can effectively be used out-of-the-box, rendering state of the art segmentation accessible to non-experts and catalyzing scientific progress as a framework for automated method design.
\end{abstract}
 \blfootnote{$\dagger$ Equal contribution. $\ddagger$ Work started while doing a PhD at the German Cancer Research Center.}
\section{Introduction}
Semantic segmentation transforms raw biomedical image data into meaningful, spatially structured information and thus plays an essential role for scientific discovery in the field \citeManuscript{falk2019u,hollon2020near}. At the same time, semantic segmentation is an essential ingredient to numerous clinical applications \citeManuscript{aerts2014decoding,nestle2005comparison}, including applications of artificial intelligence in diagnostic support systems \citeManuscript{de2018clinically,ACDCTMI}, therapy planning support \citeManuscript{nikolov2018deep}, intra-operative assistance \citeManuscript{hollon2020near} or tumor growth monitoring \citeManuscript{kickingereder2019automated}. The high interest in automatic segmentation methods manifests in a thriving research landscape, accounting for $70\%$ of international image analysis competitions in the biomedical sector \citeManuscript{maier2018rankings}.\\\\ 
Despite the recent success of deep learning-based segmentation methods, their applicability to specific image analysis problems of end-users is often limited. The task-specific design and configuration of a method requires high levels of expertise and experience, with small errors leading to strong performance drops \citeManuscript{litjens2017survey}. Especially in 3D biomedical imaging, where dataset properties like imaging modality, image size, (anisotropic) voxel spacing or class ratio vary drastically, the pipeline design can be cumbersome, because experience on what constitutes a successful configuration may not translate to the dataset at hand. The numerous expert decisions involved in designing and training a neural network range from the exact network architecture to the training schedule and methods for data augmentation or post-processing. Each sub-component is controlled by essential hyperparameters like learning rate, batch size, or class sampling \citeManuscript{litjens2017survey}. An additional layer of complexity on the overall setup is posed by the hardware available for training and inference \citeManuscript{lecun20191}. Algorithmic optimization of the codependent design choices in this high dimensional space of hyperparameters is technically demanding and amplifies both the number of required training cases as well as compute resources by orders of magnitude \citeManuscript{elsken2019neural}. As a consequence, the end-user is commonly left with an iterative trial and error process during method design that is mostly driven by their individual experience, only scarcely documented and hard to replicate, inevitably evoking suboptimal segmentation pipelines and methodological findings that do not generalize to other datasets \citeManuscript{litjens2017survey,bergstra2012random}.\\\\ 
To further complicate things, there is an unmanageable number of research papers that propose architecture variations and extensions for performance improvement. This bulk of studies is incomprehensible to the non-expert and difficult to evaluate even for experts \citeManuscript{litjens2017survey}. Approximately 12000 studies cite the 2015 U-Net architecture on biomedical image segmentation \citeManuscript{ronneberger2015u}, many of which propose extensions and advances. We put forward the hypothesis that a basic U-Net is still hard to beat if the corresponding pipeline is designed adequately.\\\\
To this end, we propose nnU-Net (“no new net”), which makes successful 3D biomedical image segmentation accessible for biomedical research applications. nnU-Net automatically adapts to arbitrary datasets and enables out-of-the-box segmentation on account of two key contributions: 
\begin{enumerate}
    \item We formulate the pipeline optimization problem in terms of a data fingerprint (representing the key properties of a dataset) and a pipeline fingerprint (representing the key design choices of a segmentation algorithm).
    \item We make their relation explicit by condensing domain knowledge into a set of heuristic rules that robustly generate a high quality pipeline fingerprint from a corresponding data fingerprint while considering associated hardware constraints.
\end{enumerate}
In contrast to algorithmic approaches for method configuration that are formulated as a task-specific optimization problem, nnU-Net readily executes systematic rules to generate deep learning methods for previously unseen datasets without need for further optimization.\\\\
In the following, we demonstrate  the superiority of this concept by presenting a new state of the art in numerous international challenges through application of our algorithm without manual intervention. The strong results underline the significance of nnU-Net for users who require algorithms for semantic segmentation on their custom datasets: as an open source tool, nnU-Net can simply be downloaded and trained out-of-the box to generate state of the art segmentations without requiring manual adaptation or expert knowledge. We further demonstrate shortcomings in the design process of current biomedical segmentation methods. Specifically, we take an in-depth look at the 2019 Kidney and Kidney Tumor Segmentation (KiTS) semantic image segmentation challenge and demonstrate how important task-specific design and configuration of a method are in comparison to choosing one of the many architectural extensions and advances previously proposed on top of the U-Net. By automating this design and configuration process, nnU-Net fosters the ambition and the ability of researchers to validate novel ideas on larger numbers of datasets, while at the same time serving as an ideal reference method when demonstrating methodological improvements.
\section{Results}
\begin{figure}
\includegraphics[width=\textwidth]{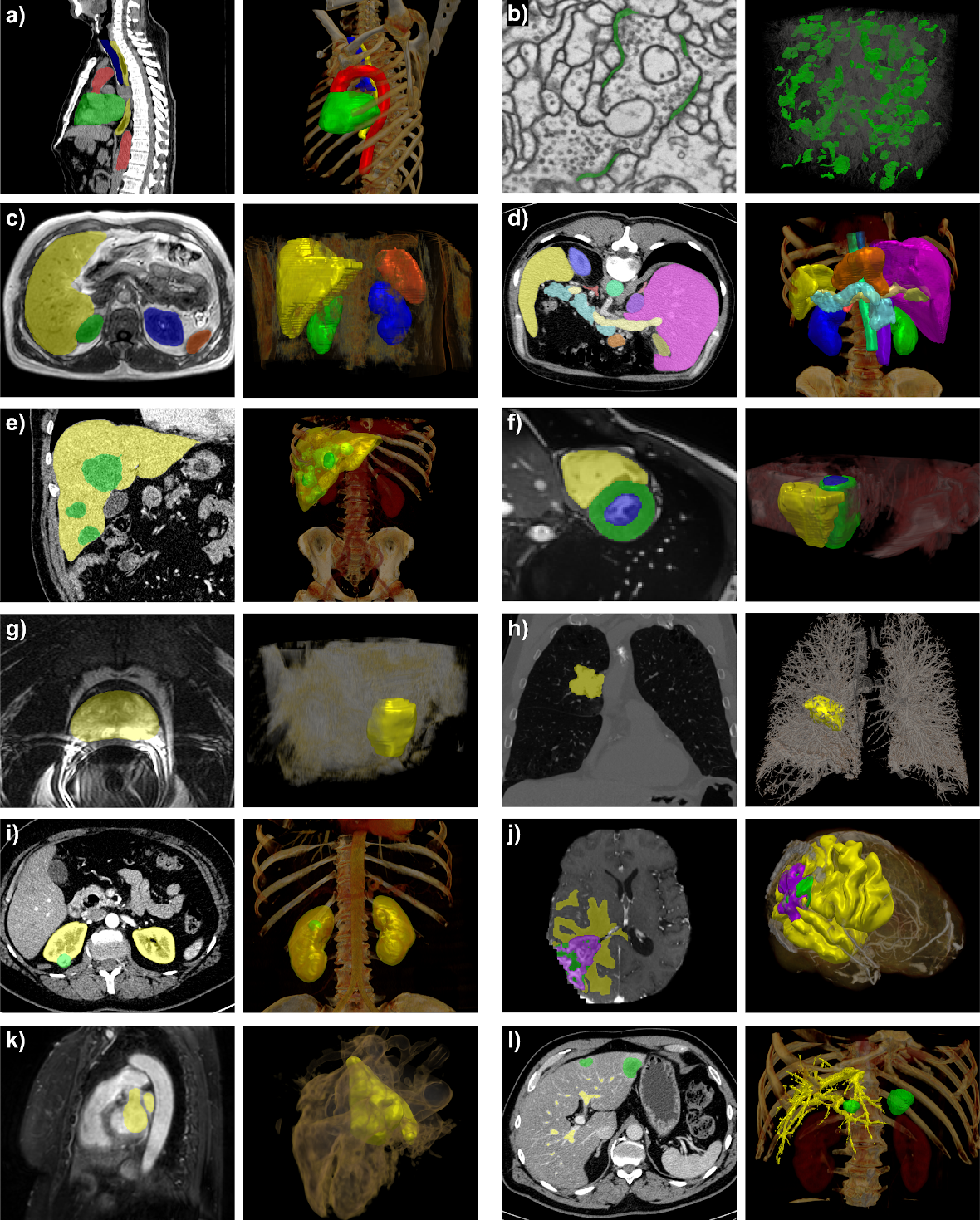}
\caption[]{\textbf{nnU-Net handles a broad variety of datasets and target image properties.} All examples originate from the test sets of different international segmentation challenges that nnU-Net was applied on. Target structures for each dataset are shown in 2D projected onto the raw data (left) and in 3D together with a volume rendering of the raw data (right). All visualizations are created with the MITK Workbench \citeManuscript{nolden2013medical}. a: heart (green), aorta (red), trachea (blue) and esophagus (yellow) in CT images (D18). b: synaptic clefts (green) in electron microscopy scans (D19). c: liver (yellow), spleen (orange), left/right kidney (blue/green) in T1 in-phase MRI (D16). d: thirteen abdominal organs in CT images (D11). e: liver (yellow) and liver tumors (green) in CT images (D14). f: right ventricle (yellow), left ventricular cavity (blue) myocardium of left ventricle (green) in cine MRI (D13). g: prostate (yellow) in T2 MRI (D12). h: lung nodules (yellow) in CT images (D6). i: kidneys (yellow) and kidney tumors (green) in CT images (D17). j: edema (yellow), enhancing tumor (purple), necrosis (green) in MRI (T1, T1 with contrast agent, T2, FLAIR) (D1). k: left ventricle (yellow) in MRI (D2). l: hepatic vessels (yellow) and liver tumors (green) in CT (D8). See Figure \ref{fig:dataset_diversity} for dataset references.}
\label{fig:qualitative_results}
\end{figure}

\begin{figure}
\includegraphics[width=\textwidth]{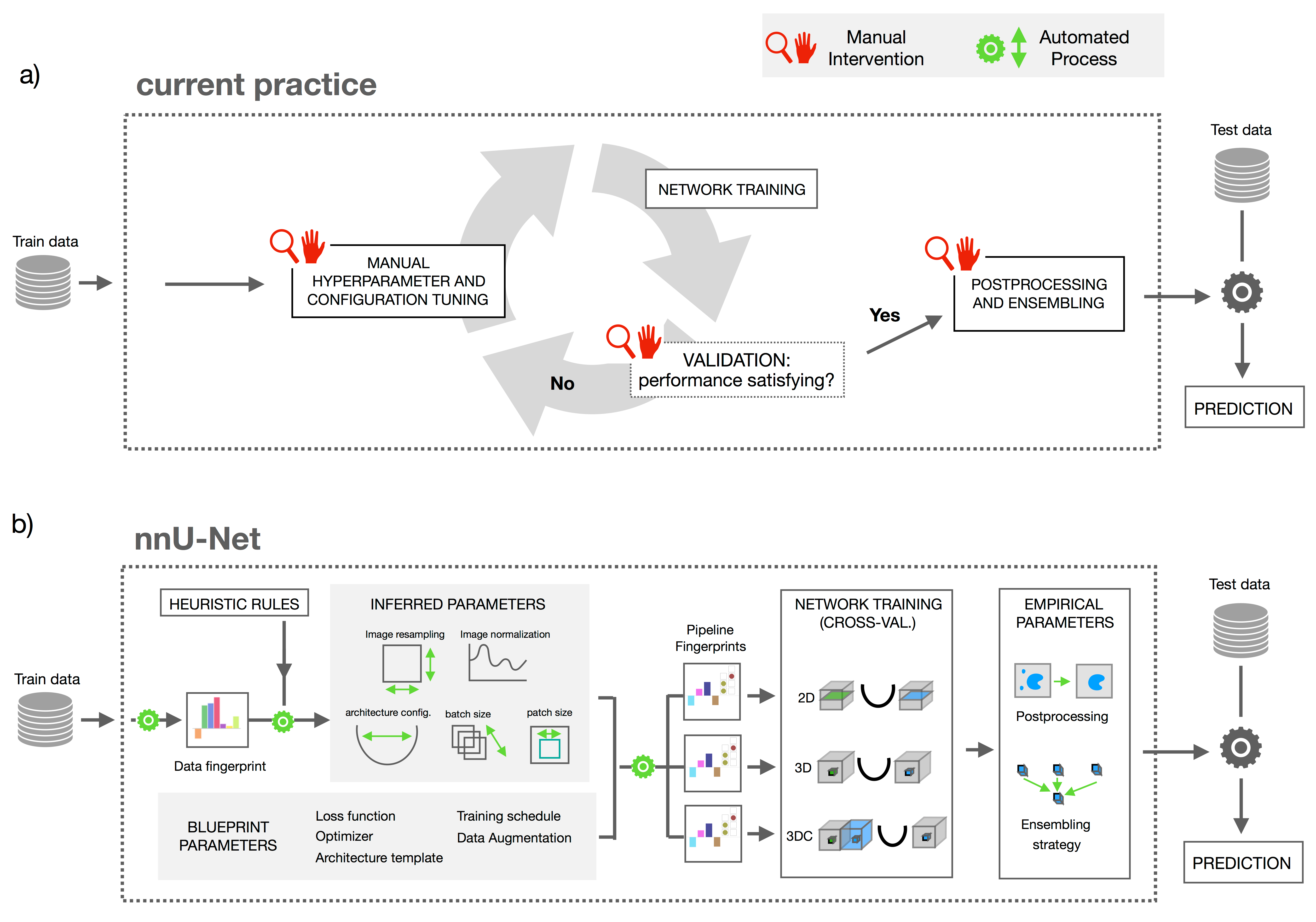}
\caption{\textbf{Manual and proposed automated configuration of deep learning methods.} a) Current practice of configuring a deep learning method for biomedical segmentation: An iterative trial and error process of manually choosing a set of hyperparameters and architecture configurations, training the model, and monitoring performance of the model on a validation set.  b) Proposed automated configuration by nnU-Net: Dataset properties are summarized in a “dataset fingerprint”. A set of heuristic rules operates on this fingerprint to infer the data-dependent hyperparameters of the pipeline. These are completed by blueprint parameters, the data-independent design choices to form “pipeline fingerprints”. Three architectures are trained based on these pipeline fingerprints in a 5-fold cross-validation. Finally, nnU-Net automatically selects the optimal ensemble of these architectures and performs postprocessing if required.}
\label{fig:nnunet_vs_current_practice}
\end{figure}

\begin{figure}
\includegraphics[width=\textwidth]{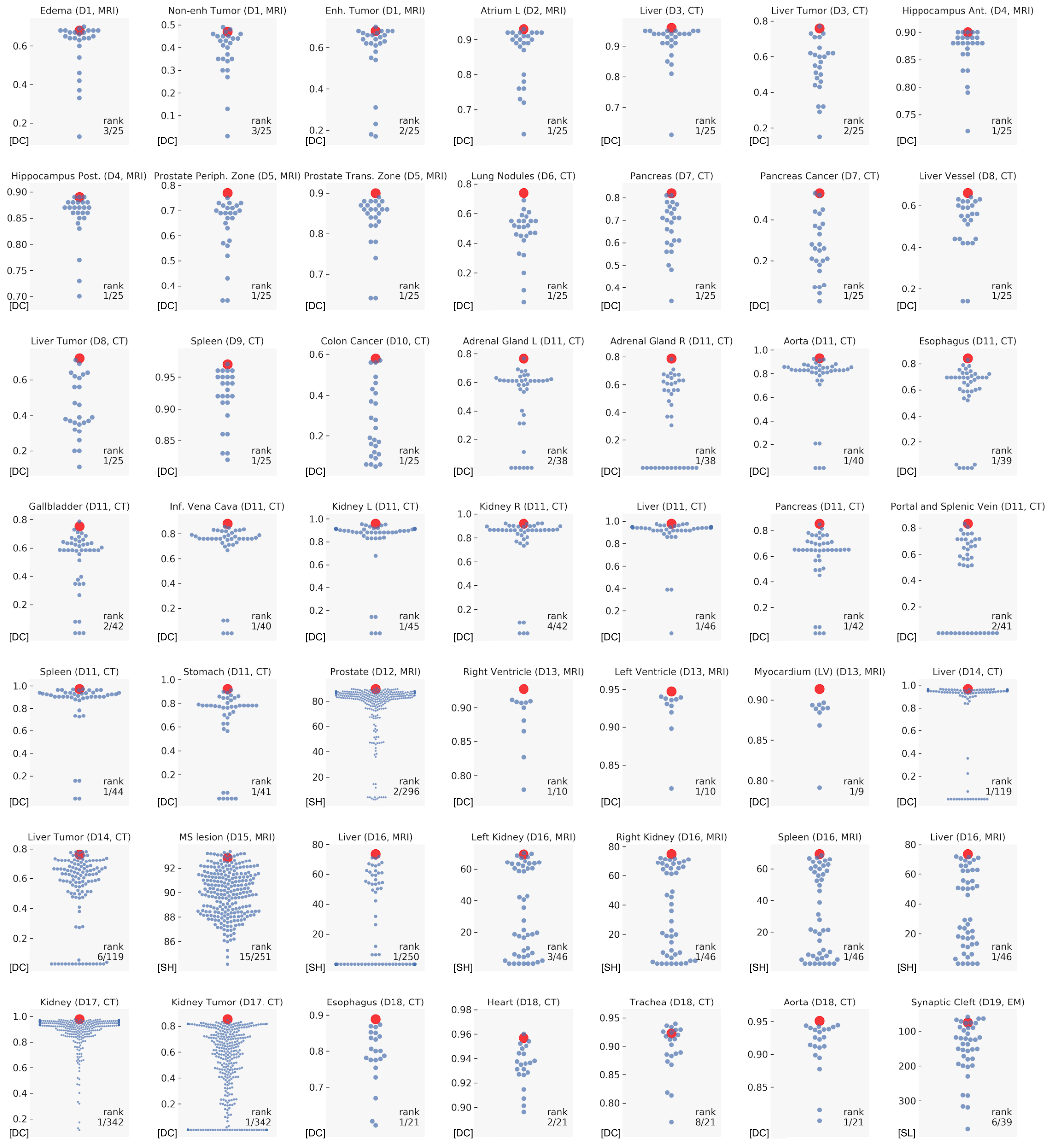}
\caption{\textbf{nnU-Net outperforms most specialized deep learning pipelines.} Quantitative results from all international challenges that nnU-Net competed in. For each segmentation task, results achieved by nnU-Net are highlighted in red, competing teams are shown in blue. For each segmentation task the respective rank is displayed in the bottom right corner as nnU-Net's rank / total number of submissions. Axis scales: [DC] Dice coefficient, [OH] other score (higher is better), [OL] other score (lower is better). All leaderboards were accessed on December 12th 2019.}
\label{fig:quantitative_results}
\end{figure}

\begin{figure}
\begin{center}
\includegraphics[width=0.75\textwidth]{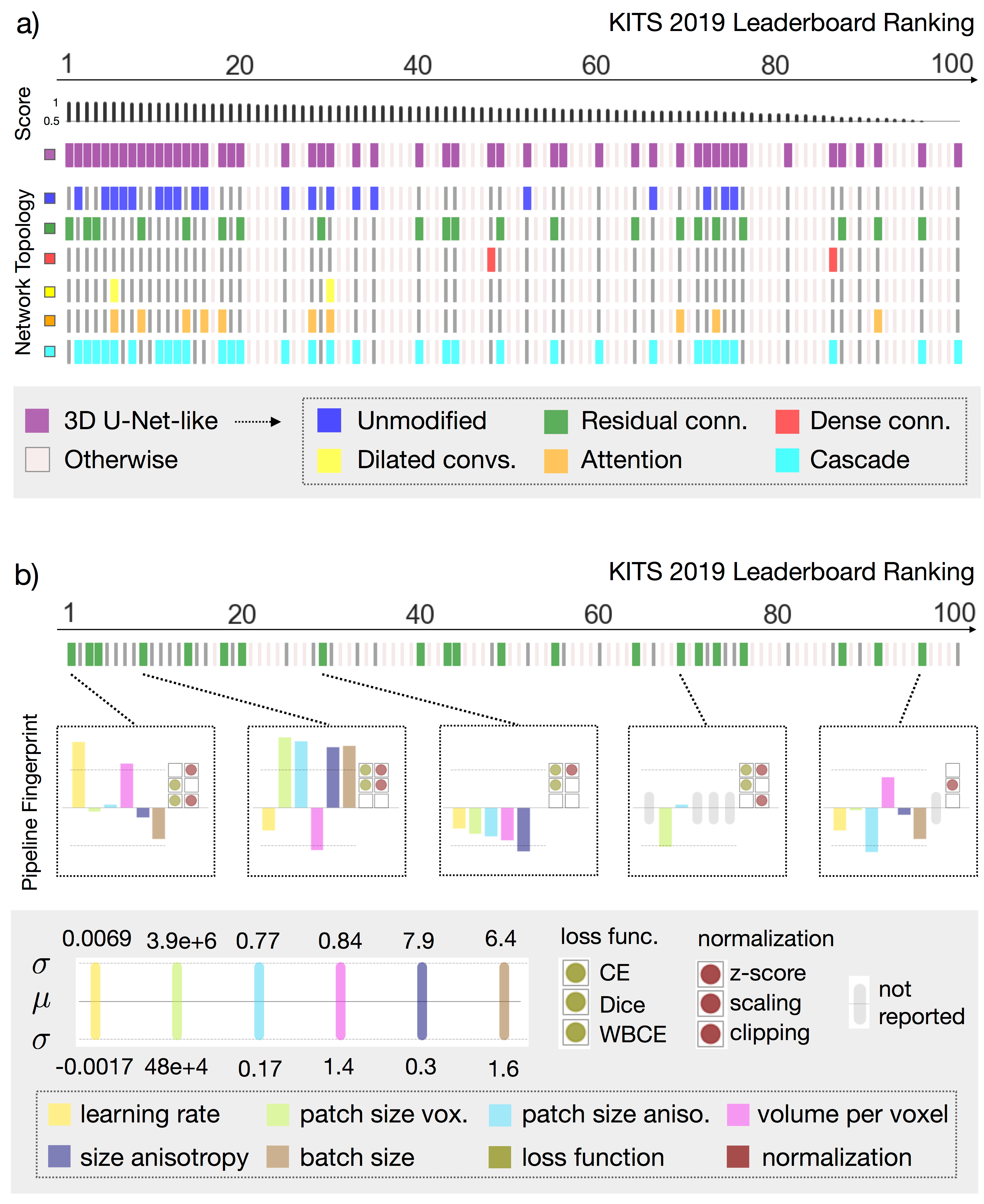}
\caption[]{\textbf{Pipeline fingerprints from KITS 2019 \citeManuscript{heller2019state} leaderboard entries.} a) Coarse categorization of leaderboard entries by architecture variation. All top 15 contributions are encoder-decoder architectures with skip-connections, 3D convolutions and output stride 1 (“3D U-Net-like”, purple). No clear pattern arises from further sub-groupings into different architectural variations. Also, none of the analyzed architectures guarantees good performance, indicating a large dependency of performance beyond architecture type. b) Finer-grained key parameters selected from the pipeline fingerprints of all non-cascade 3D-U-Net-like architectures with residual connections (displayed on z-score normalized scale). The contributions vary drastically in their rankings as well as their fingerprints. Still, there is no evident relation between single parameters and performance. Abbreviations: CE = Cross entropy loss function, Dice = Soft Dice loss function, WBCE = Weighted binary cross entropy loss function.}
\label{fig:kits_analysis}
\end{center}
\end{figure}

nnU-Net is a deep learning framework that enables 3D semantic segmentation in many biomedical imaging applications, without requiring the design of respective specialised solutions. Exemplary segmentation results generated by nnU-Net for a variety of datasets are shown in Figure \ref{fig:qualitative_results}.
\paragraph{nnU-Net automatically adapts to any new dataset}
Figure \ref{fig:nnunet_vs_current_practice}a shows the current practice of adapting segmentation pipelines to a new dataset. This process is expert-driven and involves manual trial-and-error experiments that are typically specific to the task at hand \citeManuscript{litjens2017survey}. As shown in Figure \ref{fig:nnunet_vs_current_practice}b, nnU-Net addresses the adaptation process systematically. Therefore, we define a \textit{dataset fingerprint} as a standardized dataset representation comprising key properties such as image sizes, voxel spacing information or class ratios, and a \textit{pipeline fingerprint} as the entirety of choices being made during method design. nnU-Net is designed to generate a successful pipeline fingerprint for a given dataset fingerprint. In nnU-Net, the pipeline fingerprint is divided into three groups: blueprint, inferred and empirical parameters. The blueprint parameters represent fundamental design choices (such as using a plain U-Net-like architecture template) as well as hyperparameters for which a robust default value can simply be picked (for example loss function, training schedule and data augmentation). The inferred parameters encode the necessary adaptations to a new dataset and include modifications to the exact network topology, patch size, batch size and image preprocessing. The link between a data fingerprint and the inferred parameters is established via execution of a set of heuristic rules, without the need for expensive re-optimization when applied to unseen datasets. Note that many of these design choices are co-dependent: The target image spacing, for instance, affects image size, which in return determines the size of patches the model should see during training, which affects the network topology and has to be counterbalanced by the size of training mini-batches in order to not exceed GPU memory limitations. nnU-Net strips the user of the burden to manually account for these co-dependencies. The empirical parameters are autonomously identified via cross-validation on the training cases. Per default, nnU-Net generates three different U-Net configurations: a 2D U-Net, a 3D U-Net that operates at full image resolution and a 3D U-Net cascade where the first U-Net operates on downsampled images and the second is trained to refine the segmentation maps created by the former at full resolution.  After cross-validation nnU-Net empirically chooses the best performing configuration or ensemble. Finally, nnU-Net empirically opts for “non-largest component suppression” as a postprocessing step if performance gains are measured. The output of nnU-Net’s automated adaptation and training process are fully trained U-Net models that can be deployed to make predictions on unseen images. We provide an in-depth description of the methodology behind nnU-Net in the online methods. The overarching design principles, i.e. our best-practice recommendations for method adaptation to new datasets, are summarized in Supplementary Information \ref{supplement:cheat_sheet}.  All segmentation pipelines generated by nnU-Net in the context of this manuscript are provided in Supplementary Information \ref{supplement:challenge_participation_summary}.
\paragraph{nnU-Net handles a wide variety of target structures and image properties}
We demonstrate the value of nnU-Net as an out-of-the-box segmentation tool by applying it to 10 international biomedical image segmentation challenges comprising 19 different datasets and 49 segmentation tasks across a variety of organs, organ substructures, tumors, lesions and cellular structures in magnetic resonance imaging (MRI), computed tomography scans (CT) as well as electron microscopy (EM) images. Challenges are international competitions that can be seen as the equivalent to clinical trials for algorithm benchmarking. Typically, they are hosted by individual researchers, institutes, or societies, aiming to assess the performance of multiple algorithms in a standardized environment \citeManuscript{maier2018rankings}. In all segmentation tasks, nnU-Net was trained from scratch using only the provided challenge data. While the methodology behind nnU-Net was developed on the 10 training sets provided by the Medical Segmentation Decathlon \citeManuscript{decathlonDataPaper}, the remaining datasets and tasks were used for independent testing, i.e. nnU-Net was simply applied without further optimization. Qualitatively, we observe that nnU-Net is able to handle a large disparity in dataset properties and diversity in target structures, i.e. generated pipeline configurations are in line with what human experts consider a reasonable or sensible setting  (see Supplementary Information \ref{supplement:example_pipelines_acdc}and \ref{supplement:example_pipelines_lits}). Examples for segmentation results generated by nnU-Net are presented in Figure \ref{fig:qualitative_results}. 
\paragraph{nnU-Net outperforms specialized pipelines in a range of diverse tasks}
Most international challenges use the Soerensen-Dice coefficient as a measure of overlap to quantify segmentation quality \citeManuscript{heller2019state,bilic2019liver, menze2014multimodal, ACDCTMI}. Here, perfect agreement results in a Dice coefficient of 1, whereas no agreement results in a score of 0. Other metrics used by some of the challenges include the Normalized Surface Dice (higher is better) \citeManuscript{de2018clinically} and the Hausdorff Distance (lower is better), both quantifying the distance between the borders of two segmentations. 
Figure \ref{fig:quantitative_results} provides an overview of the quantitative results achieved by nnU-Net and the competing challenge teams across all 49 segmentation tasks. Despite its generic nature, nnU-Net outperforms most existing semantic segmentation solutions, even though the latter were specifically optimized towards the respective task. Overall, nnU-Net sets a new state of the art in 29 out of 49 target structures and otherwise shows performances on par with or close to the top leaderboard entries. 
\paragraph{Details in pipeline configuration have more impact on performance than architectural variations}
To highlight how important the task-specific design and configuration of a method are in comparison to choosing one of the many architectural extensions and advances previously proposed on top of the U-Net, we put our results into context of current research by analyzing the participating algorithms in the recent Kidney and Kidney Tumor Segmentation (KiTS) 2019 challenge hosted by the Medical Image Computing and Computer Assisted Intervention (MICCAI) society \citeManuscript{heller2019state}. The MICCAI society has consistently been hosting at least $50\%$ of all annual biomedical image analysis challenges \citeManuscript{maier2018rankings}. With more than 100 competitors, the KiTS challenge was the largest competition at MICCAI 2019.
Our analysis of the KiTS leaderboard\footnote{\url{http://results.kits-challenge.org/miccai2019/}} (see Figure \ref{fig:kits_analysis}a) reveals several insights on the current landscape of deep learning based segmentation method design: First, the top-15 methods were offspring of the (3D) U-Net architecture from 2016, confirming its impact on the field of biomedical image segmentation. Second, the figure demonstrates that contributions using the same type of network result in performances spread across the entire leaderboard. Third, when looking closer into the top-15, none of the commonly used architectural modifications (e.g. residual connections \citeManuscript{milletari2016v,he2016deep}, dense connections \citeManuscript{jegou2017one,huang2017densely}, attention mechanisms \citeManuscript{oktay2018attention} or dilated convolutions \citeManuscript{chen2017deeplab,mckinley2018ensembles}) represent a necessary condition for good performance on the KiTS task. By example this shows that many of the previously introduced algorithm modifications may not generally be superior to a properly tuned baseline method. 

\begin{figure}
\includegraphics[width=\textwidth]{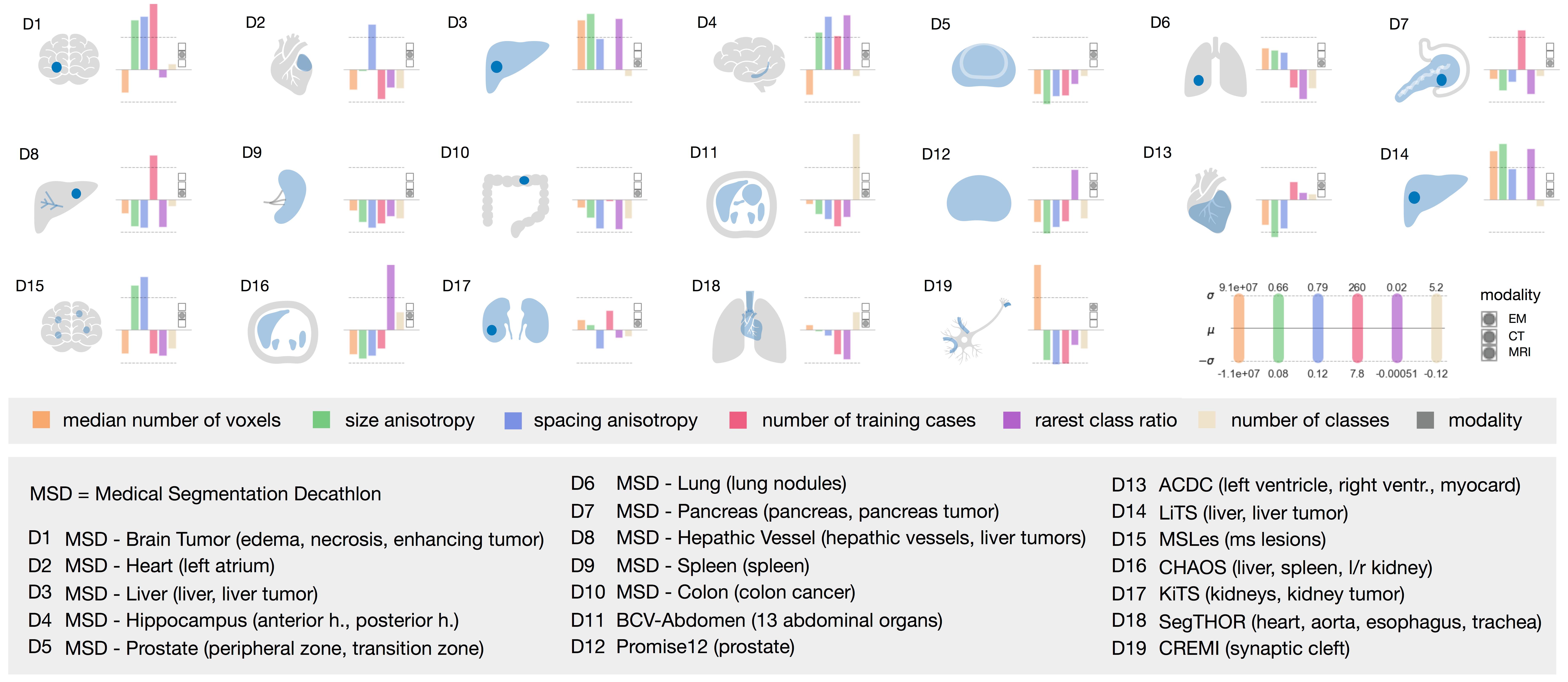}
\caption{\textbf{Data fingerprints across different challenge datasets.} The data fingerprints show the key properties (displayed on z-score normalized scale) for the 19 datasets used in the nnU-Net experiments (see Supplementary Material \ref{supplement:dataset_details} for detailed dataset descriptions). Datasets vary tremendously in their properties, requiring intense method adaptation to the individual dataset and underlining the need for evaluation on larger numbers of datasets when drawing general methodological conclusions. Abbreviations: EM = Electron Microscopy, CT = Computed Tomography, MRI = Magnetic Resonance Imaging.}
\label{fig:dataset_diversity}
\end{figure}

\begin{figure}
\includegraphics[width=\textwidth]{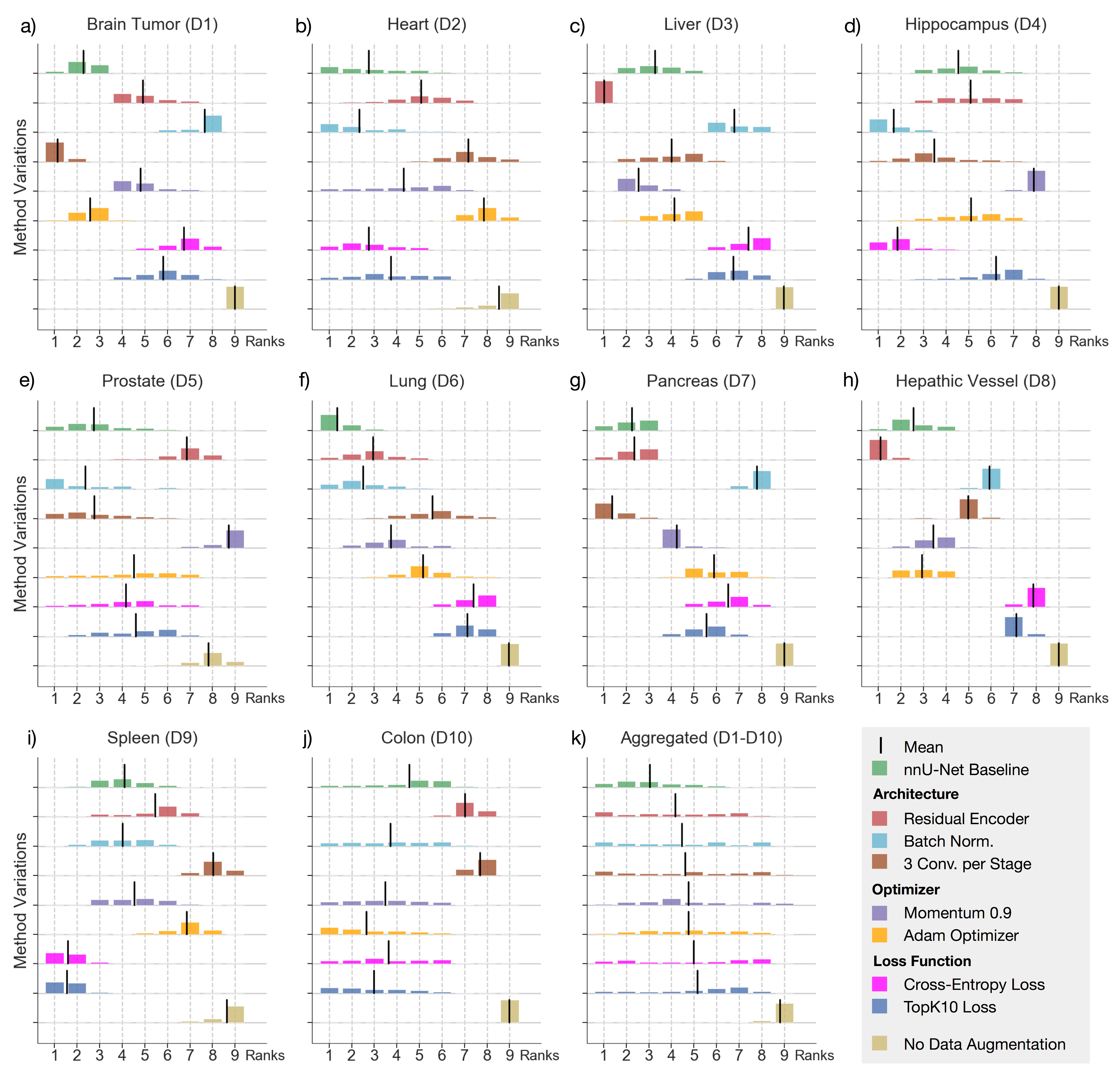}
\caption{\textbf{Evaluation of design decisions across multiple tasks.} (a-j) Evaluation of exemplary model variations on ten datasets of the medical segmentation decathlon (D1-D10, see Figure \ref{fig:dataset_diversity} for dataset references). The analysis is done for every dataset by aggregating validation splits of the five-fold cross-validation into one large validation set. 1000 virtual validation sets are generated via bootstrapping (drawn with replacement). Algorithms are ranked on each virtual validation set, resulting in a distribution over rankings. The results indicate that evaluation of methodological variations on too few datasets is prone to result in a misleading level of generality, since most performance changes are not consistent over datasets. (k) The aggregation of rankings across datasets yields insights into what design decisions robustly generalize.}
\label{fig:ablation_studies}
\end{figure}

Figure \ref{fig:kits_analysis}b underlines the importance of hyperparameter tuning by analyzing algorithms using the same architecture variant as the challenge-winning contribution, a 3D U-Net with residual connections. While one of these methods won the challenge, other contributions based on the same principle cover the entire range of evaluation scores and rankings. Key configuration parameters were selected from respective pipeline fingerprints and are shown for all non-cascaded residual U-Nets, illustrating the co-dependent design choices that each team made during pipeline design. The drastically varying configurations submitted by contestants indicate the underlying complexity of the high-dimensional optimization problem that is implicitly posed by designing a deep learning method for biomedical 3D image segmentation.

nnU-Net experimentally confirms the importance of good hyperparameters over architectural variations on the KiTS dataset by setting a new state of the art on the open leaderboard (which also includes the original challenge submissions analysed here) with a plain 3D U-Net architecture (see Figure \ref{fig:quantitative_results}). Our results from further international challenge participations confirm this observation across a variety of datasets. 
\paragraph{Different datasets require different pipeline configurations}
We extract the data fingerprints of 19 biomedical segmentation datasets. As displayed in Figure \ref{fig:dataset_diversity}, this documents an exceptional dataset diversity in biomedical imaging, and reveals the fundamental reason behind the lack of out-of-the-box segmentation algorithms: The complexity of method design is amplified by the fact that suitable pipeline settings either directly or indirectly depend on the data fingerprint under potentially complex relations. As a consequence, pipeline settings that are identified as optimal for one dataset (such as KiTS, see above) may not generalize to others, resulting in a need for (currently manual) re-optimization on each individual dataset. An example for configuration parameters depending on dataset properties is the image size which affects the size of patches that the model sees during training, which in turn affects the required network topology (i.e. number of downsampling steps, size of convolution filters, etc.). The network topology itself again influences several other hyperparameters in the pipeline. 
\paragraph{Multiple tasks enable robust design decisions}
nnU-Net is a framework that enables benchmarking of new modifications or extensions of methods across multiple datasets without having to manually reconfigure the entire pipeline for each dataset. To demonstrate this, and also to support some of the core design choices made in nnU-Net, we systematically tested the performance of common pipeline variations in the nnU-Net blueprint parameters on 10 different datasets (Figure \ref{fig:ablation_studies}): the application of two alternative loss functions (Cross-entropy and TopK10 \citeManuscript{wu2016bridging}), the introduction of residual connections in the encoder \citeManuscript{he2016identity}, using three convolutions per resolution instead of two (resulting in a deeper network architecture), two modifications of the optimizer (a reduced momentum term and an alternative optimizer (Adam \citeManuscript{adam})), batch norm \citeManuscript{ioffe2015batch} instead of instance norm \citeManuscript{ulyanov2016instance} and the omission of data augmentation. Ranking stability was estimated by bootstrapping as suggested by the challengeR tool \citeManuscript{wiesenfarth2019methods}. 

The volatility of the ranking between datasets demonstrates how single hyperparameter choices can affect segmentation performance depending on the dataset. The results clearly show that caution is required when drawing methodological conclusions from evaluations that are based on an insufficient number of datasets. While five out of the nine variants achieved rank 1 in at least one of the datasets, neither of them exhibits consistent improvements across the ten tasks. The original nnU-Net configuration shows the best generalization and ranks first when aggregating results of all datasets. 

In current research practice, evaluation is rarely performed on more than two datasets and even then the datasets come with largely overlapping properties (such as both being abdominal CT scans). As we showed here, such evaluation is unsuitable for drawing general methodological conclusions. We relate the lack of sufficiently broad evaluations to the manual tuning effort required when adapting existing pipelines to individual datasets. nnU-Net alleviates this shortcoming in two ways: As a framework that can be extended to enable effective evaluation of new concepts across multiple tasks, and as a plug-and-play, standardized and state-of-the-art baseline to compare against.

\paragraph{nnU-Net is freely available and can be used out-of-the-box}
nnU-Net is freely available as an open-source tool. It can be installed via Python Package Index (PyPI). The source code is publicly available on Github (\url{https://github.com/MIC-DKFZ/nnUNet}). A comprehensive documentation is available together with the source code. Pretrained models for all presented datasets are available for download at \url{https://zenodo.org/record/3734294}. 
\section{Discussion}
We presented nnU-Net, a deep learning framework for biomedical image analysis that automates model design for 3D semantic segmentation tasks. The method sets a new state of the art in the majority of tasks it was evaluated on, outperforming all respective specialized processing pipelines. The strong performance of nnU-Net is not achieved by a new network architecture, loss function or training scheme (hence the name nnU-Net - “no new net”), but by replacing the complex process of manual pipeline optimization with a systematic approach based on explicit and interpretable heuristic rules. Requiring zero user-intervention, nnU-Net is the first segmentation tool that can be applied out-of-the-box to a very large range of biomedical imaging datasets and is thus the ideal tool for users who require access to semantic segmentation methods and do not have the expertise, time, or compute resources required to manually adapt existing solutions to their problem.\\\\ 
Our analysis on the KITS leaderboard as well as nnU-Net's performance across 19 datasets confirms our initial hypothesis that common architectural modifications proposed by the field during the last 5 years may not necessarily be required to achieve state-of-the-art segmentation performance. Instead, we observed that contributions using the same type of network result in performances spread across the entire leaderboard. This observation is in line with Litjens et al., who, in their review from 2017, found that "many researchers use the exact same architectures [...] but have widely varying results" \citeManuscript{litjens2017survey}. There are several possible reasons for why performance improvements based on architectural extensions proposed by the literature may not hold beyond the dataset they were proposed on: many of them are evaluated on a limited amount of datasets, often as low as a single one. In practice this largely limits their success on unseen datasets with varying properties, because the quality of the hyperparameter configuration often overshadows the effect of the evaluated architectural modification. This finding is in line with an observation by Litjens et al., who concluded that "the exact architecture is not the most important determinant in getting a good solution" \citeManuscript{litjens2017survey}. Moreover, as shown above, it can be difficult to tune existing baselines to a given dataset. This obstacle can unknowingly, but nonetheless unduly, make a new approach look better than the baseline, resulting in biased literature.\\\\
In this work, we demonstrated that nnU-Net is able to alleviate this bottleneck of current research in biomedical image segmentation in two ways: On the one hand, nnU-Net serves as a framework for methodological modifications enabling simple evaluation on an arbitrary number of datasets. On the other hand, nnU-Net represents the first standardized method that does not require manual task-specific adaptation and as such can readily serve as a strong baseline on any new 3D segmentation task.\\\\
The research performed in “AutoML” \citeManuscript{hutter2011sequential,cubuk2019autoaugment} or “Neural architecture search” \citeManuscript{elsken2019neural} has similarities to our approach in that this line of research seeks to strip the ML user or researcher of the burden to manually find good hyperparameters. In contrast to nnU-Net however, AutoML aims to learn hyperparameters directly from the data. This comes with practical difficulties such as enormous requirements with respect to compute and data resources.  Additionally, AutoML methods need to optimize the hyperparameters for each new task. The same disadvantages apply to “Grid Search” \citeManuscript{bergstra2012random}, where extensive trial and error sweeps in the hyperparameter landscape are performed  to empirically find good configurations for a specific task.  In contrast, nnU-Net transforms domain knowledge into inductive biases, thus shortcuts the high dimensional optimization of hyperparameters and minimizes required computational and data resources. As elaborated above, these heuristics are developed on the basis of 10 different datasets of the Medical Segmentation Decathlon. The diversity within these 10 datasets has proven sufficient to achieve robustness to the variability encountered in all the remaining challenge participations. This is quite remarkable given the underlying complexity of method design and strongly confirms the suitability of condensing the process in a few generally applicable rules that are simply executed when given a new dataset fingerprint and do not require any further task-specific actions. The formal definition and also publishing of these explicit rules is a step towards systematicity and interpretability in the task of hyperparameter selection, which has previously been considered a “highly empirical exercise”, for which “no clear recipe can be given.” \citeManuscript{litjens2017survey}.\\\\
Despite its strong performance across 49 diverse tasks, there might be segmentation tasks for which nnU-Net’s automatic adaptation is suboptimal. For example, nnU-Net was developed with a focus on the Dice coefficient as performance metric. Some tasks, however, might require highly domain specific target metrics for performance evaluation, which could  influence method design. Also, yet unconsidered dataset properties could exist which may cause suboptimal segmentation performance. One example is the synaptic cleft segmentation task of the CREMI challenge (https://cremi.org). While nnU-Net's performance is highly competitive (rank 6/39), manual adaptation of the loss function as well as electron microscopy-specific preprocessing may be necessary to surpass state-of-the-art performance \citeManuscript{heinrich2018synaptic}. In principle, there are two ways of handling cases that are not yet optimally covered by nnU-Net: For potentially re-occurring cases, nnU-Net's heuristics could be extended accordingly; for highly domain specific cases, nnU-Net should be seen as a good starting point for necessary modifications.\\\\
In summary, nnU-Net sets a new state of the art in various semantic segmentation challenges and displays strong generalization characteristics without need for any manual intervention, such as the tuning of hyper-parameters. As pointed out by Litjens et al. and quantitatively confirmed here, hyper-parameter optimization constitutes a major difficulty for past and current research in biomedical image segmentation. nnU-Net automates the otherwise often unsystematic and cumbersome procedure and may thus help alleviate this burden. We propose to leverage nnU-Net as an out-of-the box tool for state-of-the-art segmentation, a framework for large-scale evaluation of novel ideas without manual effort, and as a standardized baseline method to compare ideas against without the need for task-specific optimization.
\section{Acknowledgements}
This work was co-funded by the National Center for Tumor Diseases (NCT) in Heidelberg and the Helmholtz Imaging Platform (HIP) of the German Cancer Consortium (DKTK). We thank our colleagues at DKFZ who were involved in the various challenge contributions, especially Andre Klein, David Zimmerer, Jakob Wasserthal, Gregor Koehler, Tobias Norajitra and Sebastian Wirkert who contributed to the Decathlon submission. We also thank the MITK team who supported us in producing all medical dataset visualizations. We are also thankful to all the challenge organizers, who provided an important basis for our work. We want to especially mention Nicholas Heller, who enabled the collection of all the details from the KiTS challenge through excellent challenge design, and Emre Kavur from the CHAOS team, who generated comprehensive leaderboard information for us. We thank Manuel Wiesenfarth for his helpful advice concerning the ranking of methods and the visualization of rankings. Last but not least, we thank Olaf Ronneberger and Lena Maier-Hein for their important feedback on this manuscript.

\bibliographystyle{abbrv}

\newpage

\section*{Methods}
\label{onlinemethods}
A quick overview of the nnU-Net design principles can be found in the Supplemental Material \ref{supplement:cheat_sheet}. This section provides detailed information on how these principles are implemented.
\paragraph{Dataset fingerprints} 
As a first processing step, nnU-Net crops the provided training cases to their nonzero region. While this had no effect on most datasets in our experiments, it reduced the image size of brain datasets such as D1 (Brain Tumor) and D15 (MSLes) substantially and thus improved computational efficiency. Based on the cropped training data, nnU-Net creates a dataset fingerprint that captures all relevant parameters and properties: image sizes (i.e. number of voxels per spatial dimension) before and after cropping, image spacings (i.e. the physical size of the voxels), modalities (read from metadata) and number of classes for all images as well as the total number of training cases. Furthermore, the fingerprint includes the mean, standard deviation as well as the 0.5 and 99.5 percentiles of the intensity values in the foreground regions, i.e. the voxels belonging to any of the class labels, computed over all training cases. 
\paragraph{Pipeline fingerprints}
nnU-Net automizes the design of deep learning methods for biomedical image segmentation by generating a so-called pipeline fingerprint that contains all relevant information. Importantly, nnU-Net reduces the design choices to the really essential ones and automatically infers these choices using a set of heuristic rules. These rules condense the domain knowledge and operate on the above-described data fingerprint and the project-specific hardware constraints. These inferred parameters are complemented by blueprint parameters, which are data-independent, and empirical parameters, which are optimized during training. 
\paragraph{Blueprint parameters}
Architecture template: All U-Net architectures configured by nnU-Net originate from the same template. This template closely follows the original U-Net \cite{OM_ronneberger2015u} and its 3D counterpart \cite{OM_cciccek20163d}. According to our hypothesis that a well-configured plain U-Net is still hard to beat, none of our U-Net configurations make use of recently proposed architectural variations such as residual connections \cite{OM_he2016deep,OM_he2016identity}, dense connections \cite{OM_huang2017densely,OM_jegou2017one}, attention mechanisms \cite{OM_oktay2018attention}, squeeze and excitation \cite{OM_hu2018squeeze} or dilated convolutions \cite{OM_chen2017deeplab}. Minor changes with respect to the original architecture were made: To enable large patch sizes, the batch size of the networks in nnU-Net is small. In fact, most 3D U-Net configurations were trained with a batch size of only 2 (see Supplementary Material Figure \ref{fig:supplement_architecture_configuration}a). Batch normalization \cite{OM_ioffe2015batch}, which is often used to speed up or stabilize the training, does not perform well with small batch sizes \cite{OM_wu2018group,OM_singh2019filter}. We therefore use instance normalization \cite{OM_ulyanov2016instance} for all U-Net models. Furthermore, we replace ReLU with leaky ReLUs \cite{OM_maas2013rectifier} (negative slope 0.01). Networks are trained with deep supervision: additional auxiliary losses are added in the decoder to all but the two lowest resolutions, allowing gradients to be injected deeper into the network and facilitating the training of all layers in the network. All U-Nets employ the very common configuration of two blocks per resolution step in both encoder and decoder, with each block consisting of a convolution, followed by instance normalization and a leaky ReLU nonlinearity. Downsampling is implemented as strided convolution (motivated by representational bottleneck, see \cite{OM_szegedy2016rethinking})  and upsampling as convolution transposed. As a tradeoff between performance and memory consumption, the initial number of feature maps is set to 32 and doubled (halved) with each downsampling (upsampling) operation. To limit the final model size, the number of feature maps is additionally capped at 320 and 512 for 3D and 2D U-Nets, respectively.

\textit{Training schedule:} Based on experience and as a trade-off between runtime and reward, all networks are trained for 1000 epochs with one epoch being defined as iteration over 250 minibatches. Stochastic gradient descent with nesterov momentum ($\mu = 0.99$) and an initial learning rate of 0.01 is used for learning network weights. The learning rate is decayed throughout the training following the ‘poly’ learning rate policy \cite{OM_chen2017deeplab}:  $(1 - \mathrm{epoch} /\mathrm{epoch}_{\mathrm {max}})^{0.9}$. The loss function is the sum of cross-entropy and Dice loss \cite{OM_drozdzal2016importance}. For each deep supervision output, a corresponding downsampled ground truth segmentation mask is used for loss computation. The training objective is the sum of the losses at all resolutions: $L = w_1 \cdot L_1 + w_2 \cdot L_2 + … $ . Hereby, the weights halve with each decrease in resolution, resulting in $w_2 = 1/2 \cdot w_1; w_3 = 1/4 \cdot w_1$, etc. and are normalized to sum to 1.
Samples for the mini batches are chosen from random training cases. Oversampling is implemented to ensure robust handling of class imbalances: $66.7\%$ of samples are from random locations within the selected training case while $33.3\%$ of patches are guaranteed to contain one of the foreground classes that are present in the selected training sample (randomly selected). The number of foreground patches is rounded with a forced minimum of 1 (resulting in 1 random and 1 foreground patch with batch size 2). A variety of data augmentation techniques are applied on the fly during training: rotations, scaling, Gaussian noise, Gaussian blur, brightness, contrast, simulation of low resolution, gamma and mirroring. Details are provided in Supplementary Information \ref{supplement:dataaugmentation}.

\textit{Inference:} Images are predicted with a sliding window approach, where the window size equals the patch size used during training. Adjacent predictions overlap by half the size of a patch. The accuracy of segmentation decreases towards the borders of the window. To suppress stitching artifacts and reduce the influence of positions close to the borders, a Gaussian importance weighting is applied, increasing the weight of the center voxels in the softmax aggregation. Test time augmentation by mirroring along all axes is applied.
\paragraph{Inferred Parameters}
\textit{Intensity normalization:} There are two different image intensity normalization schemes supported by nnU-Net. The default setting for all modalities except CT images is z-scoring. For this option, during training and inference, each image is normalized independently by subtracting its mean, followed by division with its standard deviation. If cropping resulted in an average size decrease of $25\%$ or more, a mask for central non-zero voxels is created and the normalization is applied within that mask only, ignoring the surrounding zero voxels. For computed tomography (CT) images, nnU-Net employs a different scheme, as intensity values are quantitative and reflect physical properties of the tissue. It can therefore be beneficial to retain this information by using a global normalization scheme that is applied to all images. To this end, nnU-Net uses the 0.5 and 99.5 percentiles of the foreground voxels for clipping as well as the global foreground mean a standard deviation for normalization on all images.

\textit{Resampling:} In some datasets, particularly in the medical domain, the voxel spacing (the physical space the voxels represent) is heterogeneous. Convolutional neural networks operate on voxel grids and ignore this information. To cope with this heterogeneity, nnU-Net resamples all images to the same target spacing (see paragraph below) using either third order spline, linear or nearest neighbor interpolation. The default setting for image data is third order spline interpolation. For anisotropic images (maximum axis spacing / minimum axis spacing > 3), in-plane resampling is done with third order spline whereas out of plane interpolation is done with nearest neighbor. Treating the out of plane axis differently in anisotropic cases suppresses resampling artifacts, as large contour changes between slices are much more common. Segmentation maps are resampled by converting them to one hot encodings. Each channel is then interpolated with linear interpolation and the segmentation mask is retrieved by an argmax operation. Again, anisotropic cases are interpolated using “nearest neighbor” on the low resolution axis.  

\textit{Target spacing:} The selected target spacing is a crucial parameter. Larger spacings result in smaller images and thus a loss of details whereas smaller spacings result in larger images preventing the network from accumulating sufficient contextual information since the patch size is limited by the given GPU memory budget. Although this tradeoff is in part addressed by the 3D U-Net cascade (see below), a sensible target spacing for low and full resolution is still required. For the 3D full resolution U-Net, nnU-Net uses the median value of the spacings found in the training cases computed independently for each axis as default target spacing. For anisotropic datasets, this default can result in severe interpolation artifacts or in a substantial loss of information due to large variances in resolution across the training data. Therefore, the target spacing of the lowest resolution axis is selected to be the 10th percentile of the spacings found in the training cases if both voxel and spacing anisotropy (i.e. the ratio of lowest spacing axis to highest spacing axis) are larger than 3. For the 2D U-Net, nnU-Net generally operates on the two axes with the highest resolution. If all three axes are isotropic, the two trailing axes are utilized for slice extraction. The target spacing is the median spacing of the training cases (computed independently for each axis). For slice-based processing, no resampling along the out-of-plane axis is required. 

\textit{Adaptation of network topology, patch size and batch size:} Finding an appropriate U-Net architecture configuration is crucial for good segmentation performance. nnU-Net prioritizes large patch sizes while remaining within a predefined GPU memory budget. Larger patch sizes allow for more contextual information to be aggregated and thus typically increase segmentation performance. They come, however, at the cost of a decreased batch size which results in noisier gradients during backpropagation. To improve the stability of the training, we require a minimum batch size of 2 and choose a large momentum term for network training (see blueprint parameters). Image spacing is also considered in the adaptation process: Downsampling operations may operate only on specific axes and convolutional kernels in the 3D U-Nets can operate on certain image planes only (pseudo-2D). The network topology for all U-Net configurations is chosen on basis of the median image size after resampling as well as the target spacing the images were resampled to. A flow chart for the adaptation process is presented in the Supplements in Figure \ref{fig:supplement_architecture_configuration}. The adaptation of the architecture template, which is described in more detail in the following, is computationally inexpensive. Due to the GPU memory consumption estimate being based on feature map sizes, no GPU is required to run the adaptation process. 

\textit{Initialization:} The patch size is initialized as the median image shape after resampling. If the patch size is not divisible by $2^{n_{d}}$ for each axis, where $n_d$ is the number of downsampling operations, it is padded accordingly.

\textit{Architecture topology:} The architecture is configured by determining the number of downsampling operations along each axis depending on the patch size and voxel spacing. Downsampling is performed until further downsampling would reduce the feature map size to smaller than 4 voxels or the feature map spacings become anisotropic. The downsampling strategy is determined by the voxel spacing: high resolution axes are downsampled separately until their resolution is within factor 2 of the lower resolution axis. Subsequently, all axes are downsampled simultaneously. Downsampling is terminated for each axis individually, once the respective feature map constraint is triggered. The default kernel size for convolutions is $3 \times 3 \times 3$ and $3 \times 3$ for 3D U-Net and 2D U-Net, respectively. If there is an initial resolution discrepancy between axes (defined as a spacing ratio larger than 2), the kernel size for the out-of-plane axis is set to 1 until the resolutions are within a factor of 2. Note that the convolutional kernel size then remains at 3 for all axes.

\textit{Adaptation to GPU memory budget:} The largest possible patch size during configuration is limited by the amount of GPU memory. Since the patch size is initialized to the median image shape after resampling, it is initially too large to fit into the GPU for most datasets. nnU-Net estimates the memory consumption of a given architecture based on the size of the feature maps in the network, comparing it to reference values of known memory consumption. The patch size is then reduced in an iterative process while updating the architecture configuration accordingly in each step until the required budget is reached (see Figure \ref{fig:supplement_architecture_configuration} in the Supplements). The reduction of the patch size is always applied to the largest axis relative to the median image shape of the data. The reduction in one step amounts to $2^{n_{d}}$ voxels of that axis, where $n_d$ is the number of downsampling operations.

\textit{Batch size:} As a final step, the batch size is configured. If a reduction of patch size was performed the batch size is set to 2. Otherwise, the remaining GPU memory headroom is utilized to increase the batch size until the GPU is fully utilized. To prevent overfitting, the batch size is capped such that the total number of voxels in the minibatch do not exceed $5\%$ of the total number of voxels of all training cases. 
Examples for generated U-Net architectures are presented in Supplementary Information \ref{fig:supplement_acdc_example} and \ref{fig:supplement_lits_example}.

\textit{Configuration of 3D U-Net cascade:} Running a segmentation model on downsampled data increases the size of patches in relation to the image and thus enables the network to accumulate more contextual information. This comes at the cost of a reduction in details in the generated segmentations and may also cause errors if the segmentation target is very small or characterized by its texture. In a hypothetical scenario with unlimited GPU memory,  it is thus generally favored to train models at full resolution with a patch size that covers the entire image. The 3D U-Net cascade approximates this approach by first running a 3D U-Net on downsampled images and then training a second, full resolution 3D U-Net to refine the segmentation maps of the former. This way, the “global”, low resolution network utilizes maximal contextual information to generate its segmentation output, which then serves as an additional input channel that guides the second, “local” U-Net. The cascade is triggered only for datasets where the patch size of the 3d full resolution U-Net covers less than $12.5\%$ of the median image shape. If this is the case, the target spacing for the downsampled data and the architecture of the associated 3D low resolution U-Net are configured jointly in an iterative process. The target spacing is initialized as the target spacing of the full resolution data. In order for the patch size to cover a large proportion of the input image, the target spacing is then increased stepwise by $1\%$ while updating the architecture configuration accordingly in each step until the patch size of the resulting network topology surpasses $25\%$ of the current median image shape. If the current spacing is anisotropic (factor 2 difference between lowest and highest resolution axis), only the spacing of the higher resolution axes is increased. The configuration of the second 3D U-Net of the cascade is identical to the standalone 3D U-Net for which the configuration process is described above (except that the upsampled segmentation maps of the first U-Net are concatenated to its input). Figure \ref{fig:supplement_architecture_configuration}b in the Supplements provides an overview of this optimization process.
\paragraph{Empirical parameters}
Ensembling and selection of U-Net configuration(s): nnU-Net automatically determines which (ensemble of) configuration(s) to use for inference based on the average foreground Dice coefficient computed via cross-validation on the training data. The selected model(s) can be either a single U-Net (2D, 3D full resolution, 3D low resolution or the full resolution U-Net of the cascade) or an ensemble of any two of these configurations. Models are ensembled by averaging softmax probabilities.

\textit{Postprocessing:} Connected component-based postprocessing is commonly used in medical image segmentation \cite{OM_bilic2019liver,OM_heller2019state}. Especially in organ segmentation it often helps to remove spurious false positive detections by removing all but the largest connected component. nnU-Net follows this assumption and automatically benchmarks the effect of suppressing smaller components on the cross-validation results. First, all foreground classes are treated as one component. If suppression of all but the largest region improves the average foreground Dice coefficient and does not reduce the Dice coefficient for any of the classes, this procedure is selected as the first postprocessing step. Finally, nnU-Net builds on the outcome of this step and decides whether the same procedure should be performed for individual classes.
\paragraph{Implementation details}
nnU-Net is implemented in Python utilizing the PyTorch \cite{OM_paszke2019pytorch} framework. The Batchgenerators library \cite{OM_isensee_fabian_2020_3632567} is used for data augmentation. For reduction of computational burden and GPU memory footprint, mixed precision training is implemented with Nvidia Apex/Amp (\url{https://github.com/NVIDIA/apex}). For use as a framework, the source code is available on GitHub (\url{https://github.com/MIC-DKFZ/nnUNet}). Users who seek to use nnU-Net as a standardized benchmark or to run inference with our pretrained models can install nnU-Net via PyPI. For a full description of how to use nnU-Net, please refer to the online documentation available on the GitHub page.

\section*{Reporting summary} Further information on research design is available in the Nature Research Reporting Summary linked to this article.
\section*{Code availability} The nnU-Net repository is available at: \url{https://github.com/mic-dkfz/nnunet}. Pre-traiend models for all datasets utilized in this study are available for download at \url{https://zenodo.org/record/3734294}.
\section*{Data availability}
All 19 datasets used in this study are publicly available. References for web access as well as key data properties can be found in the Supplementary Material \ref{supplement:dataset_details} and \ref{supplement:challenge_participation_summary}.

\bibliographystyle{abbrv}


\renewcommand\thefigure{\thesection.\arabic{figure}}    
\renewcommand\thetable{\thesection.\arabic{table}}  

\newpage
\appendix
\LARGE{\textbf{Supplementary Information}}\\\\\\
\fontsize{10}{14}\selectfont
    This document contains supplementary information for the manuscript 'Automated Design of Deep Learning Methods for Biomedical Image Segmentation'.

\section{Dataset details}
\setcounter{figure}{0} 
\setcounter{table}{0}
    \label{supplement:dataset_details}
    
    Table \ref{tab:supplement_table_with_datasets} provides an overview of the datasets used in this manuscript including respective references for data access. The numeric values presented here are computed based on the training cases for each of these datasets. They are the basis of the dataset fingerprints presented in Figure \ref{fig:dataset_diversity}. 

    \begin{table}[h]
        \tiny
        \label{tab:supplement_table_with_datasets}
        \begin{tabular}{lllllllll}
            ID & Dataset Name & \begin{tabular}[c]{@{}l@{}}Associated\\ Challenges\end{tabular} & Modalities & \begin{tabular}[c]{@{}l@{}}Median Shape\\ (Spacing [mm])\end{tabular} & \begin{tabular}[c]{@{}l@{}}N \\ Classes \end{tabular} & \begin{tabular}[c]{@{}l@{}}Rarest \\ Class Ratio \end{tabular} & \begin{tabular}[c]{@{}l@{}}N Training\\ Cases\end{tabular} & Segmentation Tasks \\[8pt] \hline \\
            D1 & Brain Tumour & \citeSupplement{SM_decathlonDataPaper}, \citeSupplement{SM_menze2014multimodal} & \begin{tabular}[c]{@{}l@{}}MRI (T1, T1c, \\ T2, FLAIR)\end{tabular} & \begin{tabular}[c]{@{}l@{}}138x169x138 \\ (1, 1, 1)\end{tabular} & 3 & $7.3 10^{-3}$ & 484 & \begin{tabular}[c]{@{}l@{}}edema, active tumor,\\ necrosis\end{tabular} \\[8pt]
            D2 & Heart & \citeSupplement{SM_decathlonDataPaper} & MRI & \begin{tabular}[c]{@{}l@{}}115x320x232 \\ (1.37, 1.25, 1.25)\end{tabular} & 1 & $4.0 10^{-3}$ & 20 & left ventricle \\[8pt]
            D3 & Liver & \citeSupplement{SM_decathlonDataPaper}, \citeSupplement{SM_bilic2019liver} & CT & \begin{tabular}[c]{@{}l@{}}432x512x512 \\ (1, 0.77, 0.77)\end{tabular} & 2 & $2.6 10^{-2}$ & 131 & liver, liver tumors \\[8pt]
            D4 & Hippocampus & \citeSupplement{SM_decathlonDataPaper} & MRI & \begin{tabular}[c]{@{}l@{}}36x50x35 \\ (1, 1, 1)\end{tabular} & 2 & $2.7 10^{-2}$ & 260 & \begin{tabular}[c]{@{}l@{}}anterior and posterior \\ hippocampus\end{tabular} \\[8pt]
            D5 & Prostate & \citeSupplement{SM_decathlonDataPaper} & \begin{tabular}[c]{@{}l@{}}MRI\\ (T2, ADC)\end{tabular} & \begin{tabular}[c]{@{}l@{}}20x320x319 \\ (3.6, 0.62, 0.62)\end{tabular} & 2  & $5.4 10^{-3}$ & 32 & \begin{tabular}[c]{@{}l@{}}peripheral and \\ transition zone\end{tabular} \\[8pt]
            D6 & Lung & \citeSupplement{SM_decathlonDataPaper} & CT & \begin{tabular}[c]{@{}l@{}}252x512x512 \\ (1.24, 0.79, 0.79)\end{tabular} & 1 & $3.9 10^{-4}$ & 63 & lung nodules \\[8pt]
            D7 & Pancreas & \citeSupplement{SM_decathlonDataPaper} & CT & \begin{tabular}[c]{@{}l@{}}93x512x512 \\ (2.5, 0.80, 0.80)\end{tabular} & 2 & $2.0 10^{-3}$ & 282 & \begin{tabular}[c]{@{}l@{}}pancreas, pancreas \\ cancer\end{tabular} \\[8pt]
            D8 & HepaticVessel & \citeSupplement{SM_decathlonDataPaper} & CT & \begin{tabular}[c]{@{}l@{}}49x512x512 \\ (5, 0.80, 0.80)\end{tabular} & 2 & $1.1 10^{-3}$ & 303 & \begin{tabular}[c]{@{}l@{}}hepatic vessels, \\ tumors\end{tabular} \\[8pt]
            D9 & Spleen & \citeSupplement{SM_decathlonDataPaper} & CT & \begin{tabular}[c]{@{}l@{}}90x512x512 \\ (5, 0.79, 0.79)\end{tabular} & 1 & $4.7 10^{-3}$ & 41 & spleen \\[8pt]
            D10 & Colon & \citeSupplement{SM_decathlonDataPaper} & CT & \begin{tabular}[c]{@{}l@{}}95x512x512 \\ (5, 0.78, 0.78)\end{tabular} & 1 & $5.6 10^{-4}$ & 126 & colon cancer \\[8pt]
            D11 & AbdOrgSeg & \citeSupplement{SM_BCVAbdomenChallenge} & CT & \begin{tabular}[c]{@{}l@{}}128x512x512 \\ (3, 0.76, 0.76)\end{tabular} & 13 & $4.4 10^{-3}$ & 30 & \begin{tabular}[c]{@{}l@{}}13 abdominal \\ organs\end{tabular} \\[8pt]
            D12 & Promise & \citeSupplement{SM_promise12Challenge} & MRI & \begin{tabular}[c]{@{}l@{}}24x320x320 \\ (3.6, 0.61, 0.61)\end{tabular} & 1 & $2.0 10^{-2}$ & 50 & prostate \\[8pt]
            D13 & ACDC & \citeSupplement{SM_ACDCTMI} & cine MRI & \begin{tabular}[c]{@{}l@{}}9x256x216 \\ (10, 1.56, 1.56)\end{tabular} & 3 & $1.2 10^{-2}$ & \begin{tabular}[c]{@{}l@{}}200 \\ (100x2) * \end{tabular} & \begin{tabular}[c]{@{}l@{}}left ventricle, right \\ ventricle,\\ myocardium\end{tabular} \\[8pt]
            D14 & LiTS ** & \citeSupplement{SM_bilic2019liver} & CT & \begin{tabular}[c]{@{}l@{}}432x512x512 \\ (1, 0.77, 0.77)\end{tabular} & 2 & $2.6 10^{-2}$ & 131 & liver, liver tumors \\[8pt]
            D15 & MSLesion & \citeSupplement{SM_mslesionchallenge} & \begin{tabular}[c]{@{}l@{}}MRI (FLAIR, \\ MPRAGE, PD, \\ T2)\end{tabular} & \begin{tabular}[c]{@{}l@{}}137x180x137 \\ (1, 1, 1)\end{tabular} & 1 & $1.7 10^{-3}$ & \begin{tabular}[c]{@{}l@{}}42 \\ (21x2) * \end{tabular} & \begin{tabular}[c]{@{}l@{}}multiple sclerosis \\ lesions \end{tabular} \\[8pt]
            D16 & CHAOS &  \citeSupplement{SM_kavur2020chaos} & MRI & \begin{tabular}[c]{@{}l@{}}30x204x256 \\ (9, 1.66, 1.66)\end{tabular} & 4 & $3.3 10^{-2}$ & \begin{tabular}[c]{@{}l@{}}60 \\ (20 + 20x2) * \end{tabular} & \begin{tabular}[c]{@{}l@{}}liver, spleen, left and \\ right kidney\end{tabular} \\[8pt]
            D17 & KiTS & \citeSupplement{SM_heller2019state} & CT & \begin{tabular}[c]{@{}l@{}}107x512x512 \\ (3, 0.78, 0.78)\end{tabular} & 2 & $7.5 10^{-3}$ & 206 & \begin{tabular}[c]{@{}l@{}}kidney, kidney \\ tumor\end{tabular} \\[8pt]
            D18 & SegTHOR & \citeSupplement{SM_initialsegthorpaper} & CT & \begin{tabular}[c]{@{}l@{}}178x512x512 \\ (2.5, 0.98, 0.98)\end{tabular} & 4 & $4.6 10^{-4}$ & 40 & \begin{tabular}[c]{@{}l@{}}heart, aorta, \\ esophagus, trachea\end{tabular} \\[8pt]
            D19 & CREMI & \citeSupplement{SM_heinrich2018synaptic} & \begin{tabular}[c]{@{}l@{}}Electron \\ Microscopy\end{tabular} & \begin{tabular}[c]{@{}l@{}}125x1250x1250 \\ (40, 4, 4)\end{tabular} & 1 & $5.2 10^{-3}$ & 3 & synaptic clefts \\
            \vspace{0.1cm}\\
            \multicolumn{6}{l}{* multiple annotated examples per training case} \\
            \multicolumn{6}{l}{** almost identical to Decathlon Liver; Decathlon changed the training cases and test set slightly} 
        \end{tabular}
        \caption{Overview over the challenge datasets used in this manuscript.}
    \end{table}

\section{nnU-Net Design Principles}
\setcounter{figure}{0} 
\setcounter{table}{0}
    \label{supplement:cheat_sheet}
    
    Here we present a brief overview of the design principles of nnU-Net on a conceptual level. Please refer to the online methods for a more detailed information on how these guidelines are implemented.
    
    \subsection{Blueprint Parameters}
    \begin{itemize}
        \item Architecture Design decisions:
        \begin{itemize}
            \item U-Net like architectures enable state of the art segmentation when the pipeline is well-configured. According to our experience, sophisticated architectural variations are not required to achieve state of the art performance.
            \item Our architectures only use plain convolutions, instance normalization and Leaky nonlinearities. The order of operations in each computational block is conv - instance norm - leaky ReLU.
            \item We use two computational blocks per resolution stage in both encoder and decoder.
            \item Downsampling is done with strided convolutions (the convolution of the first block of the new resolution has stride >1), upsampling is done with convolutions transposed. We should note that we did not observe substantial disparities in segmentation accuracy between this approach and alternatives (e.g. max pooling, bi/trilinear upsampling). 
        \end{itemize}
        
        \item Selecting the best U-Net configuration:
        It is difficult to estimate which U-Net configuration performs best on what dataset. To address this, nnU-Net designs three separate configurations and automatically chooses the best one based on cross-validation (see inferred parameters). Predicting which configurations should be trained on which dataset is a future research direction.
        \begin{itemize}
            \item 2D U-Net: Runs on full resolution data. Expected to work well on anisotropic data, such as D5 (Prostate MRI) and D13 (ACDC, cine MRI) (for dataset references see Table \ref{tab:supplement_table_with_datasets}).
            \item 3D full resolution U-Net:  Runs on full resolution data. Patch size is limited by availability of GPU memory. Is overall the best performing configuration (see results in \ref{supplement:challenge_participation_summary}). For large data, however, the patch size may be too small to aggregate sufficient contextual information. 
            \item 3D U-Net cascade: Specifically targeted towards large data. First, coarse segmentation maps are learned by a 3D U-Net that operates on low resolution data. These segmentation maps are then refined by a second 3D U-Net that operates on full resolution data.
        \end{itemize}
        
        \item Training Scheme
        \begin{itemize}
            \item All trainings run for a fixed length of 1000 epochs, where each epoch is defined as 250 training iterations (using the batch size configured by nnU-Net). Shorter trainings than this default empirically result in diminished segmentation performance.
            \item As for the opimizer, stochastic gradient descent with a high initial learning rate (0.01) and a large nesterov momentum (0.99) empirically provided the best results. The learning rate is reduced during the training using the 'polyLR' schedule as described in \citeSupplement{SM_chen2017deeplab}, which is an almost linear decrease to 0.
            \item Data augmentation is essential to achieve state of the art performance. It is important to run the augmentations on the fly and with associated probabilities to obtain a never ending stream of unique examples (see Section \ref{supplement:dataaugmentation} for details).
            \item Data in the biomedical domain suffers from class imbalance. Rare classes could end up being ignored because they are underrepresented during training. Oversampling foreground regions addresses this issue reliably. It should, however, not be overdone so that the network also sees all the data variability of the background.
            \item The Dice loss function is well suited to address the class imbalance, but comes with its own drawbacks. Dice loss optimizes the evaluation metric directly, but due to the patch based training, in practice merely approximates it. Furthermore, oversampling of classes skews the class distribution seen during training. Empirically, combining the Dice loss with a cross-entropy loss improved training stability and segmentation accuracy. Therefore, the two loss terms are simply averaged.
        \end{itemize}
        \item Inference
        \begin{itemize}
            \item Validation sets of all folds in the cross-validation are predicted by the single model trained on the respective training data. The 5 models resulting from training on 5 individual folds are subsequently used as an ensemble for predicting test cases.
            \item Inference is done patch based with the same patch size as used during training. Fully convolutional inference is not recommended because it causes issues with zero-padded convolutions and instance normalization.
            \item To prevent stitching artifacts, adjacent predictions are done with a distance of patch\_size / 2. Predictions towards the border are less accurate, which is why we use Gaussian importance weighting for softmax aggregation (the center voxels are weighted higher then the border voxels).
        \end{itemize}
    \end{itemize}

    \subsection{Inferred Parameters}
    These parameters are not fixed across datasets, but configured on-the-fly by nnU-Net according to the data fingerprint (low dimensional representation of dataset properties) of the task at hand.
    \begin{itemize}
        \item Dynamic Network adaptation:
        \begin{itemize}
            \item The network architecture needs to be adapted to the size and spacing of the input patches seen during training. This is necessary to ensure that the receptive field of the network covers the entire input.
            \item We perform downsampling until the feature maps are relatively small (minimum is $4 \times 4( \times 4))$ to ensure sufficient context aggregation.
            \item Due to having a fixed number of blocks per resolution step in both the encoder and decoder, the network depth is coupled to its input patch size. The number of convolutional layers in the network (excluding segmentation layers) is $(5*k + 2)$ where $k$ is the number of downsampling operations (5 per downsampling stems from 2 convs in the encoder, 2 in the decoder plus the convolution transpose).
            \item Additional loss functions are applied to all but the two lowest resolutions of the decoder to inject gradients deep into the network. 
            \item For anisotropic data, pooling is first exclusively performed in-plane until the resolution matches between the axes. Initially, 3D convolutions use a kernel size of 1 (making them effectively 2D convolutions) in the out of plane axis to prevent aggregation of information across distant slices. Once an axes becomes too small, downsampling is stopped individually for this axis.
        \end{itemize}

        \item Configuration of the input patch size:
        \begin{itemize}
            \item Should be as large as possible while still allowing a batch size of 2 (under a given GPU memory constraint). This maximizes the context available for decision making in the network.
            \item Aspect ratio of patch size follows the median shape (in voxels) of resampled training cases.
        \end{itemize}
        
        \item Batch size:
        \begin{itemize}
            \item Batch size is configured with a minimum of 2 to ensure robust optimization, since noise in gradients increases with fewer sample in the minibatch.
            \item If GPU memory headroom is available after patch size configuration, the batch size is increased until GPU memory is maxed out.
        \end{itemize}
        
        \item Target spacing and resampling:
        \begin{itemize}
            \item For isotropic data, the median spacing of training cases (computed independently for each axis) is set as default. Resampling with third order spline (data) and linear interpolation (one hot encoded segmentation maps such as training annotations) give good results.
            \item For anisotropic data, the target spacing in the out of plane axis should be smaller than the median, resulting in higher resolution in order to reduce resampling artifacts. To achieve this we set the target spacing as the 10th percentile of the spacings found for this axis in the training cases. Resampling across the out of plane axis is done with nearest neighbor for both data and one-hot encoded segmentation maps.
        \end{itemize}
        
        \item Intensity normalization:
        \begin{itemize}
            \item Z-score per image (mean substraction and division by standard deviation) is a good default.
            \item We deviate from this default only for CT images, where a global normalization scheme is determined based on the intensities found in foreground voxels across all training cases.
        \end{itemize}
    \end{itemize}
    
    \subsection{Empirical Parameters}
    Some parameters cannot be inferred by simply looking at the dataset fingerprint of the training cases. These are determined empirically by monitoring validation performance after training.
    \begin{itemize}
        \item Model selection: While the 3D full resolution U-Net shows overall best performance, selection of the best model for a specific task at hand can not be predicted with perfect accuracy. Therefore, nnU-Net generates three U-Net configurations and automatically picks the best performing method (or ensemble of methods) after cross-validation.

        \item Postprocessing: Often, particularly in medical data, the image contains only one instance of the target structure. This prior knowledge can often be exploited by running connected component analysis on the predicted segmentation maps and removing all but the largest component. Whether to apply this postprocessing is determined by monitoring validation performance after cross-validation. Specifically, postprocessing is triggered for individual classes where the Dice score is improved by removing all but the largest component.
    \end{itemize}

\section{Analysis of exemplary nnU-Net-generated pipelines}
\setcounter{figure}{0} 
\setcounter{table}{0}

    In this section we briefly introduce the pipelines generated by nnU-Net for D13 (ACDC) and D14 (LiTS) to create an intuitive understanding of nnU-Nets design principles and the motivation behind them.
\subsection{ACDC}
    \label{supplement:example_pipelines_acdc}
    
    \begin{figure}
    \centering
    \includegraphics[width=\textwidth]{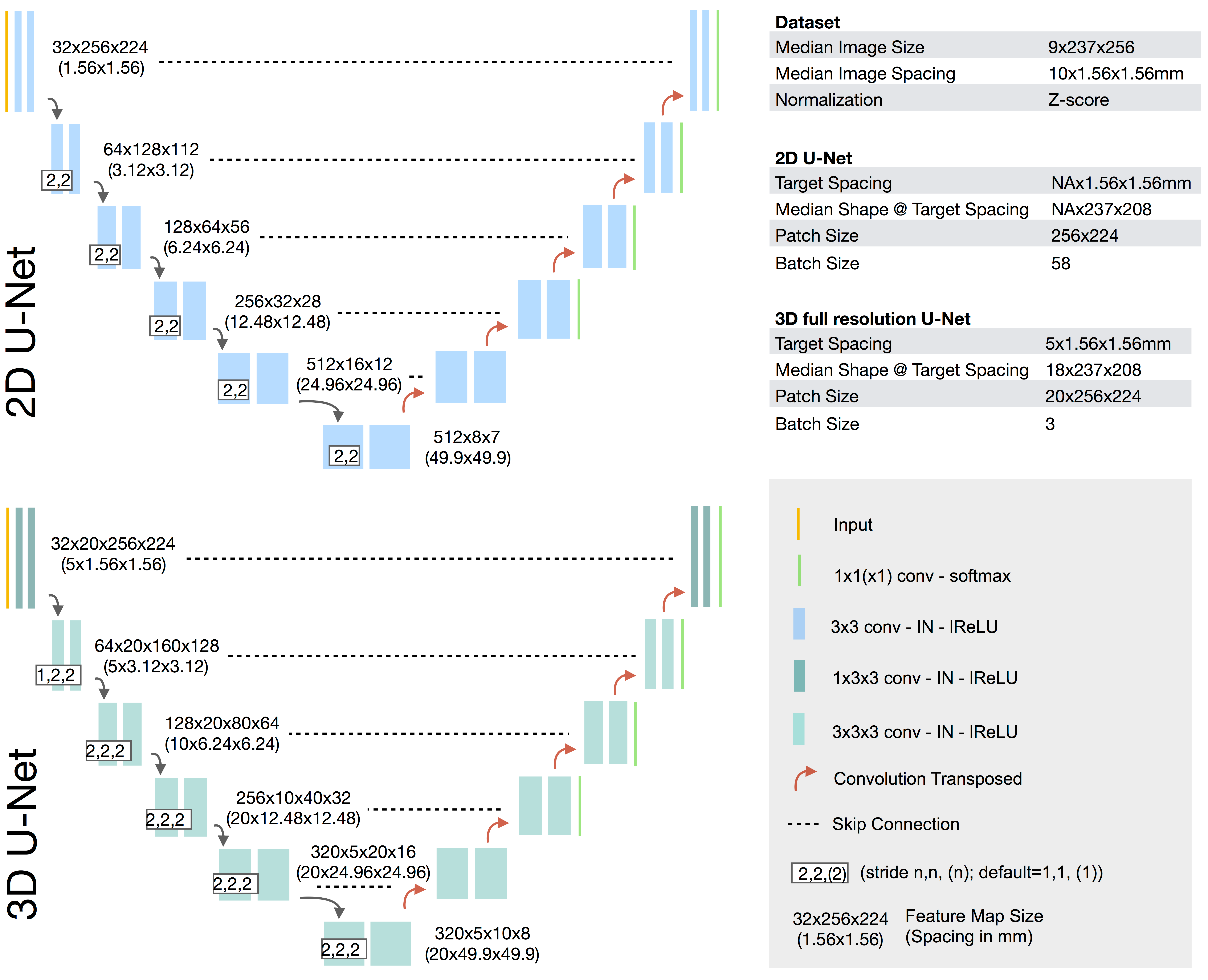}
    \caption{Network architectures generated by nnU-Net for the ACDC dataset (D13)}
    \label{fig:supplement_acdc_example}
    \end{figure}
     
    Figure \ref{fig:supplement_acdc_example} provides a summary of the pipelines that were automatically generated by nnU-Net for this dataset.
    
    \paragraph{Dataset Description} The Automated Cardiac Diagnosis Challenge (ACDC) \citeSupplement{SM_ACDCTMI} was hosted by MICCAI in 2017. Since then it is running as an open challenge with data and current leaderboard available at \url{https://acdc.creatis.insa-lyon.fr}. In the segmentation part of the challenge, participating teams were asked to generate algorithms for segmenting the right ventricle, the left myocardium and the left ventricular cavity from cine MRI. For each patient, reference segmentations for two time steps within the cardiac cycle were provided. With 100 training patients, this amounts to a total of 200 annotated images. One key property of cine MRI is that slice acquisition takes place across multiple cardiac cycles and breath holds. This results in a limited number of slices and thus a low out of plane resolution as well as the possibility for slice misalignments. Figure \ref{fig:supplement_acdc_example} provides a summary of the pipelines that were automatically generated by nnU-Net for this dataset. 
    The typical image shape (here the median image size is computed for each axis independently) is $9 \times 237 \times 256$ voxels at a spacing of $10 \times 1.56 \times  \SI{1.56}{\milli\metre}$.
    \paragraph{Intensity Normalization} With the images being MRI, nnU-Net normalizes all images individually by subtracting their mean and dividing by their standard deviation. 
    \paragraph{2D U-Net} As target spacing for the in-plane resolution, $1.56 \times \SI{1.56}{\milli\metre}$ is determined. This is identical for the 2D and the 3D full resolution U-Net. Due to the 2D U-Net operating on slices only, the out of plane resolution for this configuration is not altered and remains heterogeneous within the training set. The 2D U-Net is configured as described in the Online Methods \ref{onlinemethods} to have a patch size of $256 \times 224$ voxels, which fully covers the typical image shape after in-plane resampling ($237 \times 208$). 
    3D U-Net The size and spacing anisotropy of this dataset causes the out-of-plane target spacing of the 3D full resolution U-Net to be selected as 5mm, corresponding to the 10th percentile of the spacings found in the training cases. In datasets such as ACDC, the segmentation contour can change substantially between slices due to the large slice to slice distance. Choosing the target spacing to be lower results in more images that are upsampled for U-Net training and then downsampled for the final segmentation export. Preferring this variant over the median causes more images to be downsampled for training and then upsampled for segmentation export and therefore reduces interpolation artifacts substantially. Also note that resampling the out of plane axis is done with nearest neighbor interpolation.The median image shape after resampling for the 3D full resolution U-Net is $18 \times 237 \times 208$ voxels. As described in the Online Methods \ref{onlinemethods} nnU-Net configures a patch size of $20 \times 256 \times 224$ for network training, which fits into the memory budget with a batch size of 3. Note how the convolutional kernel sizes in the 3D U-Net start with ($1 \times 3 \times 3$) which is effectively a 2D convolution for the initial layers (see also Figure \ref{fig:supplement_acdc_example}). The reasoning behind this is that due to the large discrepancy in voxel spacing, too many changes are expected across slices and the aggregation of imaging information may therefore not be beneficial. Similarly, pooling is done in-plane only (conv kernel stride (1, 2, 2)) until the spacing between in-plane and out-of-plane axes are within a factor of 2. Only after the spacings approximately match the pooling and the convolutional kernel sizes become isotropic.
    \paragraph{3D U-Net} cascade Since the 3D U-Net already covers the whole median image shape, the U-Net cascade is not necessary and therefore omitted. 
    \paragraph{Training and Postprocessing} During training, spatial augmentations for the 3D U-Net (such as scaling and rotation) are done in-plane only to prevent resampling of imaging information across slices which would cause interpolation artifacts. Each U-Net configuration is trained in a five-fold cross-validation on the training cases. Note that we interfere with the splits in order to ensure that patients are properly stratified (since there are two images per patient). Thanks to the cross-validation, nnU-Net can use the entire training set for validation and ensembling. To this end, the validation splits of each of the five fold are aggregated. nnU-Net evaluates the performance (ensemble of models or single configuration) by averaging the Dice scores over all foreground classes and cases, resulting in a single scalar value. Detailed results are omitted here for brevity (they are presented in Supplementary Information \ref{supplement:challenge_participation_summary}). Based on this evaluation scheme, the 2D U-Net obtains a score of 0.9165, the 3D full resolution a score of 0.9181 and the ensemble of the two a score of 0.9228. Therefore the ensemble is selected for predicting the test cases. Postprocessing is configured on the segmentation maps of the ensemble. Removing all but the largest connected component was found beneficial for the right ventricle and the left ventricular cavity.

\subsection{LiTS}
    \label{supplement:example_pipelines_lits}

    \begin{figure}
    \centering
    \includegraphics[width=\textwidth]{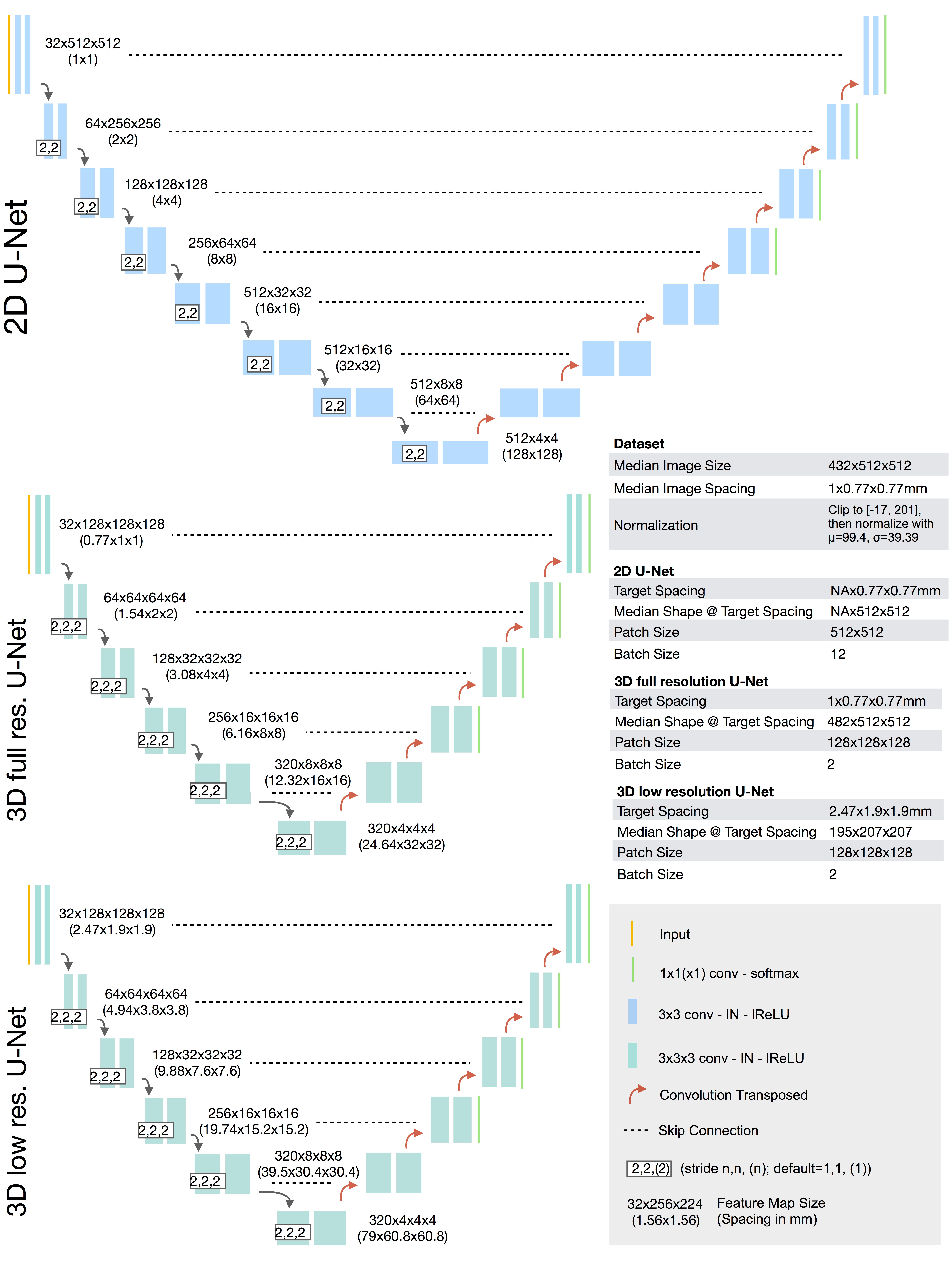}
    \caption{Network architectures generated by nnU-Net for the LiTS dataset (D14)}
    \label{fig:supplement_lits_example}
    \end{figure}
    
    Figure \ref{fig:supplement_lits_example} provides a summary of the pipelines that were automatically generated by nnU-Net for this dataset.
     \paragraph{Dataset Description} The Liver and Liver Tumor Segmentation challenge (LiTS) \citeSupplement{SM_bilic2019liver} was hosted by MICCAI in 2017. Due to the large, high quality dataset it provides, the challenge plays an important role in concurrent research. The challenge is hosted at \url{https://competitions.codalab.org/competitions/17094}. 
    The segmentation task in LiTS is the segmentation of the liver and liver tumors in abdominal CT scans. The challenge provides 131 training cases with reference annotations. The test set has a size of 70 cases and the reference annotations are known only to the challenge organizers. The median image shape of the training cases is $432 \times 512 \times 512$ voxels with a corresponding voxel spacing of $1 \times 0.77 \times \SI{0.77}{\milli\metre}$.
    \paragraph{Intensity Normalization} Voxel intensities in CT scans are linked to quantitative physical properties of the tissue. The intensities are therefore expected to be consistent between scanners. nnU-Net leverages this consistency by applying a global intensity normalization scheme (as opposed to ACDC in Supplementary Information \ref{supplement:example_pipelines_acdc}, where cases are normalized individually using their mean and standard deviation). To this end, nnU-Net extracts intensity information as part of the dataset fingerprint: the intensities of the voxels belonging to any of the foreground classes (liver and liver tumor) are collected across all training cases. Then, the mean and standard deviations of these values as well as their 0.5 and 99.5 percentiles are computed. Subsequently, all images are normalized by clipping them to the 0.5 and 99.5 percentiles, followed by subtraction of the global mean and division by the global standard deviation.
    \paragraph{2D U-Net} The target spacing for the 2D U-Net is determined to be $NA \times 0.77 \times \SI{0.77}{\milli\metre}$, which corresponds to the median voxel spacing encountered in the training cases. Note that the 2D U-Net operates on slices only, so the out of plane axis is left untouched. Resampling the training cases results in a median image shape of $NA \times 512 \times 512$ voxels (we indicate by NA that this axis is not resampled). Since this is the median shape, cases in the training set can be smaller or larger than that. The 2D U-Net is configured to have an input patch size of $512 \times 512$ voxels and a batch size of 12. 
    \paragraph{3D U-Net} The target spacing for the 3D U-Net is determined to be $1 \times 0.77 \times \SI{0.77}{\milli\metre}$, which corresponds to the median voxel spacing. Because the median spacing is nearly isotropic, nnU-Net does not use the 10th percentile for the out of plane axis as was the case for ACDC (see Supplementary Information \ref{supplement:example_pipelines_acdc}). The resampling strategy is decided on a per-image basis. Isotropic cases (maximum axis spacing / minimum axis spacing < 3) are resampled with third order spline interpolation for the image data and linear interpolation for the segmentations. Note that segmentation maps are always converted into a one hot representation prior to resampling which is converted back to a segmentation map after the interpolation. For anisotropic images, nnU-Net resamples the out-of-plane axis separately, as was done in ACDC.\\\\
    After resampling, the median image shape is $482 \times 512 \times 512$. nnU-Net prioritizes a large patch size over a large batch size (note that these are coupled under a given GPU memory budget) to capture as much contextual information as possible. The 3D U-Net is thus configured to have a patch size of $ 128 \times 128 \times 128$ voxels and a batch size of 2, which is the minimum allowed according to nnU-Net heuristics. Since The input patches have nearly isotropic spacing, all convolutional kernel sizes and downsampling strides are isotropic ($3 \times 3 \times 3$ and $2 \times 2 \times 2$, respectively). 
    \paragraph{3D U-Net cascade} Although nnU-Net prioritizes large input patches, the patch size of the 3D full resolution U-Net is too small to capture sufficient contextual information (it only covers 1/60 of the voxels of the median image shape after resampling). This can cause misclassifications of voxels because the patches are too ‘zoomed in’, making for instance the distinction between the spleen and the liver particularly hard. 
    The 3D U-Net cascade is designed to tackle this problem by first training a 3D U-Net on downsampled data and then refining the low-resolution segmentation output with a second U-Net that operates as full resolution. Using the process described in the Online Methods \ref{onlinemethods} as well as Figure \ref{fig:supplement_architecture_configuration} b), the target spacing for the low resolution U-Net is determined to be $2.47 \times 1.9 \times \SI{1.9}{\milli\metre}$, resulting in a median image shape of $195 \times 207 \times 207$ voxels. The 3D low resolution operates on $128 \times 128 \times 12$8 patches with a batch size of 2. Note that while this setting is identical to the 3D U-Net configuration here, this is not necessarily the case for other datasets. If the 3D full resolution U-Net data was anisotropic, nnU-Net would prioritize to downsample the higher resolution axes first resulting in a deviating network architecture, patch size and batch size.
    After five-fold cross-validation of the 3D low resolution U-Net, the segmentation maps of the respective validation sets are upsampled to the target spacing of the 3D full resolution U-Net. The full resolution U-Net of the cascade (which has an identical configuration to the regular 3D full resolution U-Net) is then trained to refine the coarse segmentation maps and correct any errors it encounters. This is done by concatenating a one hot encoding of the upsampled segmentations to the input of the network.
    
    \paragraph{Training and Postprocessing} All network configurations are trained as five fold cross-validation. nnU-Net again evaluates all configurations by computing the average Dice score across all foreground classes, resulting in a scalar metric per configuration. Based on this evaluation scheme, the scores are 0.7625 for the 2D U-Net, 0.8044 for the 3D full resolution U-Net, 0.7796 for the 3D low resolution U-Net and 0.8017 for the full resolution 3D U-Net of the cascade. The best combination of two models was identified as the one between the low and full resolution U-Nets with a score of 0.8111. Postprocessing is configured on the segmentation maps of this ensemble. Removing all but the largest connected component was found beneficial for the combined foreground region (union of liver and liver tumor label) as well as for the liver label alone, as both resulted in small performance gains when empirically testing it on the training data.

\section{Details on nnU-Net's Data Augmentation}
\setcounter{figure}{0} 
\setcounter{table}{0}
    \label{supplement:dataaugmentation}
    A variety of data augmentation techniques is applied during training. All augmentations are computed on the fly on the CPU using background workers. The data augmentation pipeline is implemented with the publicly available \textit{batchgenerators} framework \footnote{\url{https://github.com/MIC-DKFZ/batchgenerators}}. nnU-Net does not vary the parameters of the data augmentation pipeline between datasets.\\\\ 
    Sampled patches are initially larger than the patch size used for training. This results in less out of boundary values (here 0) being introduced during data augmentation when rotation and scaling is applied. As a part of the rotation and scaling augmentation, patches are center-cropped to the final target patch size. To ensure that the borders of original images appear in the final patches, preliminary crops may initially extend outside the boundary of the image.\\\\
    Spatial augmentations (rotation, scaling, low resolution simulation) are applied in 3D for the 3D U-Nets and applied in 2D when training the 2D U-Net or a 3D U-Net with anisotropic patch size. A patch size is considered anisotropic if the largest edge length of the patch size is at least three times larger than the smallest. \\\\
    To increase the variability in generated patches, most augmentations are varied with parameters drawn randomly from predefined ranges. In this context, $x \sim U(a, b)$ indicates that $x$ was drawn from a uniform distribution between $a$ and $b$. Furthermore, all augmentations are applied stochastically according to a predefined probability.\\\\
    The following augmentations are applied by nnU-Net (in the given order):
    \begin{enumerate}
    \item \textbf{Rotation and Scaling.} Scaling and rotation are applied together for improved speed of computation. This approach reduces the amount of required data interpolations to one. Scaling and rotation are applied with a probability of 0.2 each (resulting in probabilities of 0.16 for only scaling, 0.16 for only rotation and 0.08 for both being triggered). If processing isotropic 3D patches, the angles of rotation (in degrees) $\alpha_x$, $\alpha_y$ and $\alpha_z$ are each drawn from $U(-30, 30)$. If a patch is anisotropic or 2D, the angle of rotation is sampled from $U(-180, 180)$. If the 2D patch size is anisotropic, the angle is sampled from $U(-15, 15)$. 
    Scaling is implemented via multiplying coordinates with a scaling factor in the voxel grid. Thus, scale factors smaller than one result in  a "zoom out" effect while values larger one result in a "zoom in" effect. The scaling factor is sampled from $U(0.7, 1.4)$ for all patch types.
    \item \textbf{Gaussian Noise.} Zero centered additive Gaussian noise is added to each voxel in the sample independently. This augmentation is applied with a probability of 0.15. The variance of the noise is drawn from $U(0, 0.1)$ (note that the voxel intensities in all samples are close to zero mean and unit variance due to intensity normalization).
    \item \textbf{Gaussian Blur.} Blurring is applied with a probability of 0.2 per sample. If this augmentation is triggered in a sample, blurring is applied with a probability of 0.5 for each of the associated modalities (resulting in a combined probability of only 0.1 for samples with a single modality). The width (in voxels) of the Gaussian kernel $\sigma$ is sampled from $U(0.5, 1.5)$ independently for each modality.
    \item \textbf{Brightness.} Voxel intensities are multiplied by $x\sim U(0.7, 1.3)$ with a probability of 0.15.
    \item \textbf{Contrast.} Voxel intensities are multiplied by $x\sim U(0.65, 1.5)$ with a probability of 0.15. Following multiplication, the values are clipped to their original value range.
    \item \textbf{Simulation of low resolution.} This augmentation is applied with a probability of 0.25 per sample and 0.5 per associated modality. Triggered modalities are downsampled by a factor of $U(1, 2)$ using nearest neighbor interpolation and then sampled back up to their original size with cubic interpolation. For 2D patches or anisotropic 3D patches, this augmentation is applied only in 2D leaving the out of plane axis (if applicable) in its original state.
    \item \textbf{Gamma augmentation.} This augmentation is applied with a probability of 0.15. The patch intensities are scaled to a factor of $[0, 1]$ of their respective value range. Then, a  nonlinear intensity transformation is applied per voxel: $i_{new} = i_{old}^\gamma$ with $\gamma \sim U(0.7, 1.5)$. The voxel intensities are subsequently scaled back to their original value range. With a probability of 0.15, this augmentation is applied with the voxel intensities being inverted prior to transformation: $(1 - i_{new}) = (1 - i_{old})^\gamma$.
    \item \textbf{Mirroring.} All patches are mirrored with a probability of 0.5 along all axes. 
    \end{enumerate}
    For the full resolution U-Net of the U-net cascade, nnU-Net additionally applies the following augmentations to the segmentation masks generated by the low resolution 3D U-net. Note that the segmentations are stored as one hot encoding.
    \begin{enumerate}
    \item \textbf{Binary Operators.} With probability 0.4, a binary operator is applied to all labels in the predicted masks. This operator is randomly chosen from [dilation, erosion, opening, closing]. The structure element is a sphere with radius $r \sim U(1, 8)$. The operator is applied to the labels in random order. Hereby, the one hot encoding property is retained. Dilation of one label, for example, will result in removal of all other labels in the dilated area.
    \item \textbf{Removal of Connected Components.} With probability 0.2, connected components that are smaller than 15\% of the patch size are removed from the one hot encoding.
    \end{enumerate}

\section{Network Architecture Configuration}
\setcounter{figure}{0} 
\setcounter{table}{0}
    \label{supplement:architecture_configuration}
    
\begin{figure}
    \centering
    \includegraphics[width=\textwidth]{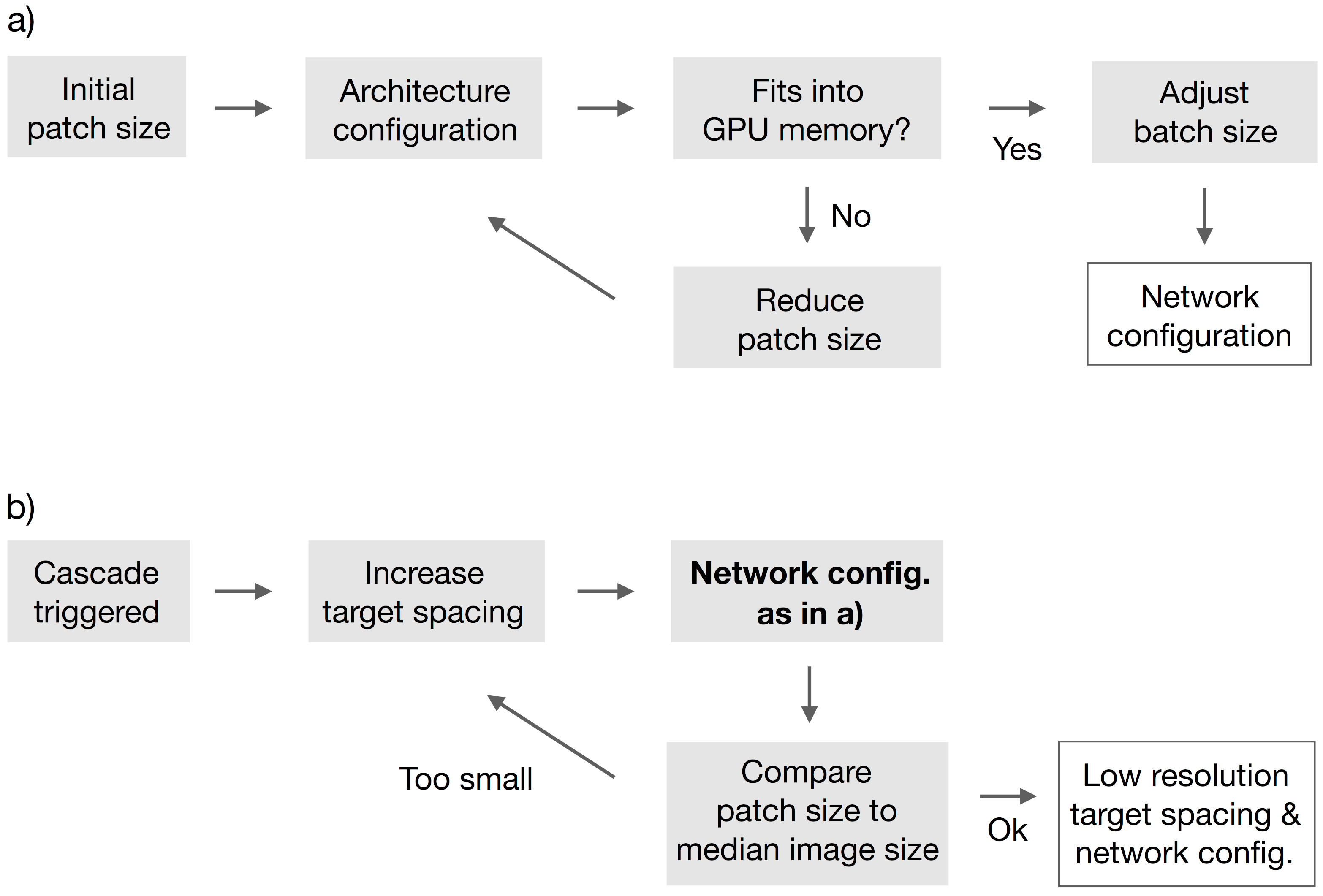}
    \caption{Workflow for network architecture configuration. a) the configuration of a U-Net architecture given an input patch size and corresponding voxel spacing. Due to discontinuities in GPU memory consumption (due to changes in number of pooling operations and thus network depth), the architecture configuration cannot be solved analytically. b) Configuration of the 3D low resolution U-Net of the U-Net cascade. The input patch size of the 3D lowres U-Net must cover at least 1/4 of the median shape of the resampled trainig cases to ensure sufficient contextual information. Higher resolution axes are downsampled first, resulting in a potentially different aspect ratio of the data relative to the full resolution data. Due to the patch size following this aspect ratio, the network architecture of the low resolution U-Net may differ from the full resolution U-Net. This requires reconfiguration of the network architecture as depicted in a) for each iteration. All computations are based on memory consumption estimates resulting in fast computation times (sub 1s for configuring all network architectures).}
    \label{fig:supplement_architecture_configuration}
    \end{figure}

    Figure \ref{fig:supplement_architecture_configuration} serves as a visual aid for the iterative process of architecture configuration described in the online methods. 

\section{Summary of nnU-Net Challenge Participations}
\setcounter{figure}{0} 
\setcounter{table}{0}
    \label{supplement:challenge_participation_summary}
    In this section we provide details of all challenge participations.\\\
    In some participations, manual intervention regarding the format of input data or the cross-validation data splits was required for compatibility with nnU-Net. For each dataset, we disclose all manual interventions in this section. The most common cause for manual intervention was training cases that were related to each other (such as multiple time points of the same patient) and thus required to be separated for mutual exclusivity between data splits. A detailed description of how to perform this intervention is further provided along with the source code.\\\\
    For each dataset, we run all applicable nnU-Net configurations (2D, 3D fullres, 3D lowres, 3D cascade) in 5-fold cross-validation. All models are trained from scratch without pretraining and trained only on the provided training data of the challenge without external training data. Note that other participants may be using external data in some competitions. For each dataset, nnU-Net subsequently identifies the ideal configuration(s) based on cross-validation and ensembling. Finally, The best configuration is used to predict the test cases.\\\\
    The pipeline generated by nnU-Net is provided for each dataset in the compact representation described in Section \ref{supplement:architecture_decoding}. We furthermore provide a table containing detailed cross-validation as well as test set results.\\\\
    All leaderboards were last accessed on December 12th, 2019.
    
\subsection{Challenge Inclusion Criteria}
    When selecting challenges for participation, our goal was to apply nnU-Net to as many different datasets as possible to demonstrate its robustness and flexibility. We applied the following criteria to ensure a rigorous and sound testing environment:
    \begin{enumerate}
        \item The task of the challenge is semantic segmentation in any 3D imaging modality with images of any size.
        \item Training cases are provided to the challenge participants.
        \item Test cases are separate, with the ground truth not being available to the challenge participants.
        \item Comparison to results from other participants is possible (e.g. through standardized evaluation with an online platform and a public leaderboard).
    \end{enumerate}
    
    The competitions outlined below are the ones who qualified under these criteria and were thus selected for evaluation of nnU-Net. To our knowledge, CREMI \footnote{\url{https://cremi.org/leaderboard/}} is the only competition from the biological domain that meets these criteria.
    
\subsection{Compact Architecture Representation}
    \label{supplement:architecture_decoding}

    \begin{figure}
    \centering
    \includegraphics[width=1\textwidth]{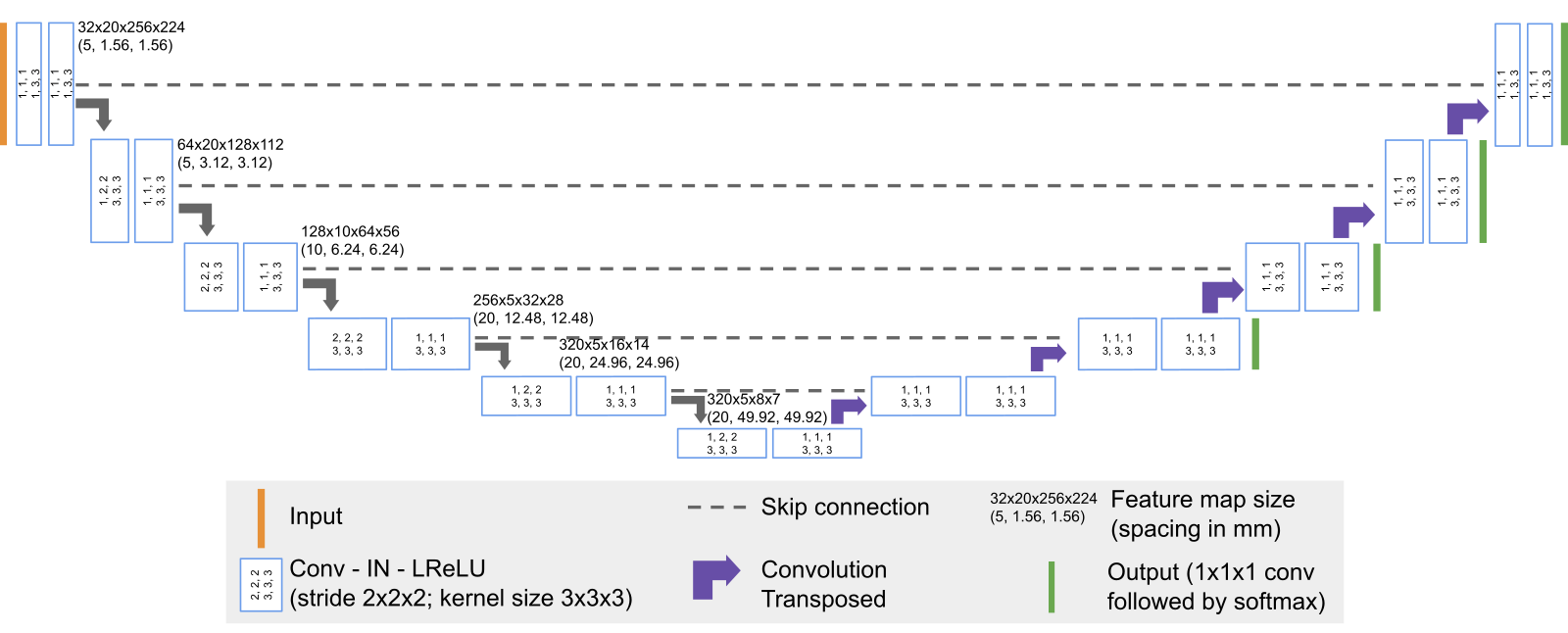}
    \caption{Decoding the architecture. We provide all generated architectures in a compact representation from which they can be fully reconstructed if desired. The architecture displayed here can be represented by means of kernel sizes {[}{[}1, 3, 3{]}, {[}3, 3, 3{]}, {[}3, 3, 3{]}, {[}3, 3, 3{]}, {[}3, 3, 3{]}, {[}3, 3, 3{]}{]} and strides {[}{[}1, 2, 2{]}, {[}2, 2, 2{]}, {[}2, 2, 2{]}, {[}1, 2, 2{]}, {[}1, 2, 2{]}{]} (see description in the text).}
    \label{fig:supplement_architecture_decoding}
    \end{figure}
    In the following sections, network architectures generated by nnU-Net will be presented in a compact representation consisting of two lists: one for the convolutional kernel sizes and one for the downsampling strides. As we describe in this section, this representation can be used to fully reconstruct the entire network architecture. The condensed representation is chosen to prevent an excessive amount of figures.\\\\
    Figure \ref{supplement:architecture_decoding} exemplary shows the 3D full resolution U-Net for the ACDC dataset (D13). The architecture has 6 resolution stages. Each resolution stage in both encoder and decoder consists of two computational blocks. Each block is a sequence of (conv - instance norm - leaky ReLU), as described in \ref{onlinemethods}. In this figure, one such block is represented by an outlined blue box. Within each box, the stride of the convolution is indicated by the first three numbers (1,1,1 for the uppermost left box) and the kernel size of the convolution is indicated by the second set of numbers (1,3,3 for the uppermost left box). Using this information, along with the template with which our architectures are designed, we can fully describe the presented architecture with the following lists:
    
    \begin{itemize}
        \item \textbf{Convolutional Kernel Sizes}: The kernel sizes of this architecture are {[}{[}1, 3, 3{]}, {[}3, 3, 3{]}, {[}3, 3, 3{]}, {[}3, 3, 3{]}, {[}3, 3, 3{]}, {[}3, 3, 3{]}{]}. Note that this list contains 6 elements, matching the 6 resolutions encountered in the encoder. Each element in this list gives the kernel size of the convolutional layers at this resolution (here this is three digits due to the convolutions being three dimensional). Within one resolution, both blocks use the same kernel size. The convolutions in the decoder mirror the encoder (dropping the last entry in the list due to the bottleneck).
        \item \textbf{Downsampling strides}: The strides for downsampling here are {[}{[}1, 2, 2{]}, {[}2, 2, 2{]}, {[}2, 2, 2{]}, {[}1, 2, 2{]}, {[}1, 2, 2{]}{]}. Each downsampling step in the encoder is represented by one entry. A stride of 2 results in a downsampling of factor 2 along that axis which a stride of 1 leaves the size unchanged. Note how the stride initially is {[}1, 2, 2{]} due to the spacing discrepancy. This changes the initial spacing of $5 \times 1.56 \times  \SI{1.56}{\milli\metre}$ to a spacing of $5 \times 3.12 \times \SI{3.12}{\milli\metre}$ in the second resolution step. The downsampling strides only apply to the first convolution of each resolution stage in the encoder. The second convolution always has a stride of {[}1, 1, 1{]}. Again, the decoder mirrors the encoder, but the stride is used as output stride of the convolution transposed (resulting in appropriate upscaling of feature maps). Outputs of all convolutions transposed have the same shape as the skip connection originating from the encoder.
    \end{itemize}
    Segmentation outputs for auxiliary losses are added to all but the two lowest resolution steps.

 \subsection{Medical Segmentation Decathlon}
     \paragraph{Challenge summary}
     The Medical Segmentation Decathlon\footnote{\url{http://medicaldecathlon.com/}} \citeSupplement{SM_decathlonDataPaper} is a competition that spans 10 different segmentation tasks. These tasks are selected to cover a large proportion of the dataset variability in the medical domain. The overarching goal of the competition was to encourage researchers to develop algorithms that can work with these datasets out of the box without manual intervention. Each of the tasks comes with respective training and test data. A detailed description of datasets can be found on the challenge homepage.
     Originally, the challenge was divided into two phases: In phase I, 7 datasets were provided to the participants for algorithm development. In phase II, the algorithms were applied to three additional and previously unseen datasets without further changes. Challenge evaluation was performed for the two phases individually and winners were determined based on their performance on the test cases. 
     
     \paragraph{Initial version of nnU-Net}
     A preliminary version of nnU-Net was developed as part of our entry in this competition, where it achieved the first rank in both phases (see \url{http://medicaldecathlon.com/results.html}). We subsequently made the respective challenge report available on arXiv \citeSupplement{SM_isensee2018nnu}.\\\\
     nnU-Net has since been refined using all ten tasks of the Medical Segmentation Decathlon. The current version of nnU-Net as presented in this publication was again submitted to the open leaderboard (\url{https://decathlon-10.grand-challenge.org/evaluation/results/}), and achieved the first rank outperforming the initial nnU-Net as well as other methods that held the state of the art since the original competition \citeSupplement{SM_yu2019c2fnas}.
     
     \paragraph{Application of nnU-Net to the Medical Segmentation Decathlon}
     nnU-Net was applied to all ten tasks of the Medical Segmentation Decathlon without any manual intervention.
\subsubsection*{BrainTumour (D1)}
    \textbf{Normalization:} Each image is normalized independently by subtracting its mean and dividing by its standard deviation.
 
    \begin{table}[H]
        \renewcommand{\arraystretch}{1.2}
        \centering
        \footnotesize
        \scalebox{0.9}{
        \begin{tabular}{llll}
            \textbf{}  & \textbf{2D U-Net}   & \textbf{3D full resolution U-Net}  & \textbf{3D low resolution U-Net}  \\
            \toprule 
            Target spacing (mm): & NA x 1 x 1  & 1 x 1 x 1  & -  \\
            \begin{tabular}[c]{@{}l@{}}Median image shape at \\ target spacing:\end{tabular} & NA x 169 x 138   & 138 169 138  & -  \\
            Patch size:  & 192 x 160   & 128 x 128 x 128  & -  \\
            Batch size:  & 107   & 2   & -   \\
            Downsampling strides: & \begin{tabular}[c]{@{}l@{}}{[}{[}2, 2{]}, {[}2, 2{]}, {[}2, 2{]}, {[}2, 2{]},\\ {[}2, 2{]}\end{tabular} & \begin{tabular}[c]{@{}l@{}}{[}{[}2, 2, 2{]}, {[}2, 2, 2{]}, {[}2, 2, 2{]}, \\ {[}2, 2, 2{]}, {[}2, 2, 2{]}{]}\end{tabular} & - \\
            Convolution kernel sizes: & \begin{tabular}[c]{@{}l@{}}{[}{[}3, 3{]}, {[}3, 3{]}, {[}3, 3{]}, {[}3, 3{]}, \\ {[}3, 3{]}, {[}3, 3{]} \end{tabular} & \begin{tabular}[c]{@{}l@{}}{[}{[}3, 3, 3{]}, {[}3, 3, 3{]}, {[}3, 3, 3{]}, \\ {[}3, 3, 3{]}, {[}3, 3, 3{]}, {[}3, 3, 3{]}{]} \end{tabular} & - \\[4pt]
        \end{tabular}}
        \caption{Network configurations generated by nnU-Net for the BrainTumour dataset from the Medical Segmentation Decathlon (D1). For more information on how to decode downsampling strides and kernel sizes into an architecture, see \ref{supplement:architecture_decoding}}
    \end{table}
    
    \begin{table}[H]
        \renewcommand{\arraystretch}{1.2}
        \footnotesize
        \centering
        \begin{tabular}{lllll}
            & edema& non-enhancing tumor& enhancing tumour& mean \\
            \toprule 
            2D& 0.7957& 0.5985& 0.7825& 0.7256\\
            3D\_fullres *& 0.8101& 0.6199& 0.7934& 0.7411\\
            Best Ensemble & 0.8106& 0.6179& 0.7926& 0.7404\\
            Postprocessed & 0.8101& 0.6199& 0.7934& 0.7411\\[4pt] 
            Test set & 0.68& 0.47& 0.68 & 0.61\\[4pt] 
        \end{tabular}
        \caption{Decathlon BrainTumour (D1) results. Note that all reported Dice scores (except the test set) were computed using five fold cross-validation on the training cases. * marks the best performing model selected for subsequent postprocessing (see "Postprocessed") and test set submission (see "Test set") Note that the Dice scores for the test set are computed with the online platform and only two significant digits are reported. Best ensemble on this dataset was the combination of the 2D U-Net and the 3D full resolution U-Net.}
    \end{table}
\subsubsection*{Heart (D2)}
    \textbf{Normalization:} Each image is normalized independently by subtracting its mean and dividing by its standard deviation.
    
    \begin{table}[H]
        \renewcommand{\arraystretch}{1.2}
        \centering
        \footnotesize
        \scalebox{0.9}{
        \begin{tabular}{llll}
            \textbf{}  & \textbf{2D U-Net}   & \textbf{3D full resolution U-Net} & \textbf{3D low resolution U-Net}   \\
            \toprule 
            Target spacing (mm): & NA x 1.25 x 1.25  & 1.37 x 1.25 x 1.25 & -  \\
            \begin{tabular}[c]{@{}l@{}}Median image shape at \\ target spacing:\end{tabular} & NA x 320 x 232   & 115 x 320 x 232 & - \\
            Patch size:  & 320 x 256   & 80 x 192 x 160  & - \\
            Batch size:  & 40   & 2   & -  \\
            Downsampling strides: & \begin{tabular}[c]{@{}l@{}}{[}{[}2, 2{]}, {[}2, 2{]}, {[}2, 2{]}, {[}2, 2{]},\\ {[}2, 2{]}, {[}2, 1{]}{]}\end{tabular} & \begin{tabular}[c]{@{}l@{}}{[}{[}2, 2, 2{]}, {[}2, 2, 2{]}, {[}2, 2, 2{]}, \\ {[}2, 2, 2{]}, {[}1, 2, 2{]}{]}\end{tabular} & - \\
            Convolution kernel sizes: & \begin{tabular}[c]{@{}l@{}}{[}{[}3, 3{]}, {[}3, 3{]}, {[}3, 3{]}, {[}3, 3{]}, \\ {[}3, 3{]}, {[}3, 3{]}, {[}3, 3{]}, {]}\end{tabular} & \begin{tabular}[c]{@{}l@{}}{[}{[}3, 3, 3{]}, {[}3, 3, 3{]}, {[}3, 3, 3{]}, \\ {[}3, 3, 3{]}, {[}3, 3, 3{]}, {[}3, 3, 3{]}{]} \end{tabular}  & -  \\[4pt]
        \end{tabular}}
        \caption{Network configurations generated by nnU-Net for the Heart dataset from the Medical Segmentation Decathlon (D2). For more information on how to decode downsampling strides and kernel sizes into an architecture, see \ref{supplement:architecture_decoding}}
    \end{table}
    
    \begin{table}[H]
        \renewcommand{\arraystretch}{1.2}
        \footnotesize
        \centering
        \begin{tabular}{lll}
            & left atrium& mean\\
            \toprule 
            2D& 0.9090& 0.9090\\
            3D\_fullres *& 0.9328& 0.9328\\
            Best Ensemble& 0.9268& 0.9268\\
            Postprocessed & 0.9329 & 0.9329 \\[4pt] 
            Test set & 0.93 & 0.93\\[4pt] 
        \end{tabular}
        \caption{Decathlon Heart (D2) results. Note that all reported Dice scores (except the test set) were computed using five fold cross-validation on the training cases. * marks the best performing model selected for subsequent postprocessing (see "Postprocessed") and test set submission (see "Test set"). Note that the Dice scores for the test set are computed with the online platform and only two significant digits are reported. Best ensemble on this dataset was the combination of the 2D U-Net and the 3D full resolution U-Net.}
    \end{table}
\subsubsection*{Liver (D3)}
    \textbf{Normalization:} Clip to $[-17, 201]$, then subtract $99.40$ and finally divide by $39.36$.
    
    \begin{table}[H]
        \renewcommand{\arraystretch}{1.2}
        \centering
        \footnotesize
        \scalebox{0.9}{
        \begin{tabular}{llll}
            \textbf{}  & \textbf{2D U-Net}   & \textbf{3D full resolution U-Net}  & \textbf{3D low resolution U-Net}  \\
            \toprule 
            Target spacing (mm): & NA x 0.7676 x 0.7676  & 1 x 0.7676 x 0.7676  & 2.47 x 1.90 x 1.90  \\
            \begin{tabular}[c]{@{}l@{}}Median image shape at \\ target spacing:\end{tabular} & NA x 512 x 512   & 482 x 512 x 512  & 195 x 207 x 207  \\
            Patch size:  & 512 x 512   & 128 x 128 x 128  & 128 x 128 x 128  \\
            Batch size:  & 12   & 2   & 2   \\
            Downsampling strides: & \begin{tabular}[c]{@{}l@{}}{[}{[}2, 2{]}, {[}2, 2{]}, {[}2, 2{]}, {[}2, 2{]},\\ {[}2, 2{]}, {[}2, 2{]}, {[}2, 2{]}{]}\end{tabular} & \begin{tabular}[c]{@{}l@{}}{[}{[}2, 2, 2{]}, {[}2, 2, 2{]}, {[}2, 2, 2{]}, \\ {[}2, 2, 2{]}, {[}2, 2, 2{]}{]}\end{tabular} & \begin{tabular}[c]{@{}l@{}}{[}{[}2, 2, 2{]}, {[}2, 2, 2{]}, {[}2, 2, 2{]}, \\ {[}2, 2, 2{]}, {[}2, 2, 2{]}{]}\end{tabular} \\
            Convolution kernel sizes: & \begin{tabular}[c]{@{}l@{}}{[}{[}3, 3{]}, {[}3, 3{]}, {[}3, 3{]}, {[}3, 3{]}, \\ {[}3, 3{]}, {[}3, 3{]}, {[}3, 3{]}, {[}3, 3{]}{]}\end{tabular} & \begin{tabular}[c]{@{}l@{}}{[}{[}3, 3, 3{]}, {[}3, 3, 3{]}, {[}3, 3, 3{]}, \\ {[}3, 3, 3{]}, {[}3, 3, 3{]}, {[}3, 3, 3{]}{]}\end{tabular} & \begin{tabular}[c]{@{}l@{}}{[}{[}3, 3, 3{]}, {[}3, 3, 3{]}, {[}3, 3, 3{]}, \\ {[}3, 3, 3{]}, {[}3, 3, 3{]}, {[}3, 3, 3{]}{]}\\[4pt] \end{tabular}
        \end{tabular}}
        \caption{Network configurations generated by nnU-Net for the Liver dataset from the Medical Segmentation Decathlon (D3). For more information on how to decode downsampling strides and kernel sizes into an architecture, see \ref{supplement:architecture_decoding}}
    \end{table}
    
    \begin{table}[H]
        \renewcommand{\arraystretch}{1.2}
        \footnotesize
        \centering
        \begin{tabular}{llll}
            & liver& cancer& mean\\
            \toprule 
            2D& 0.9547& 0.5637& 0.7592\\
            3D\_fullres& 0.9571& 0.6372& 0.7971\\
            3D\_lowres& 0.9563& 0.6028& 0.7796\\
            3D cascade& 0.9600& 0.6386& 0.7993\\
            Best Ensemble*& 0.9613& 0.6564& 0.8088\\
            Postprocessed& 0.9621& 0.6600& 0.8111\\[4pt]  
            Test set & 0.96& 0.76& 0.86\\[4pt] 
        \end{tabular}
        \caption{Decathlon Liver (D3) results. Note that all reported Dice scores (except the test set) were computed using five fold cross-validation on the training cases. * marks the best performing model selected for subsequent postprocessing (see "Postprocessed") and test set submission (see "Test set"). Note that the Dice scores for the test set are computed with the online platform and only two significant digits are reported. Best ensemble on this dataset was the combination of the 3D low resolution U-Net and the 3D full resolution U-Net.}
    \end{table}
\subsubsection*{Hippocampus (D4)}
    \textbf{Normalization:} Each image is normalized independently by subtracting its mean and dividing by its standard deviation.
    
    \begin{table}[H]
        \renewcommand{\arraystretch}{1.2}
        \centering
        \footnotesize
        \scalebox{0.9}{
        \begin{tabular}{llll}
        \textbf{}  & \textbf{2D U-Net}   & \textbf{3D full resolution U-Net}  & \textbf{3D low resolution U-Net}  \\
        \toprule 
        Target spacing (mm): & NA x 1 x 1  & 1 x 1 x 1  & -  \\
        \begin{tabular}[c]{@{}l@{}}Median image shape at \\ target spacing:\end{tabular} & NA x 50 x 35   & 36 x 50 x 35  & -  \\
            Patch size:  & 56 x 40   & 40 x 56 x 40  & -  \\
            Batch size:  & 366   & 9   & -   \\
            Downsampling strides: & \begin{tabular}[c]{@{}l@{}}{[}{[}2, 2{]}, {[}2, 2{]}, {[}2, 2{]}{]}\end{tabular} & \begin{tabular}[c]{@{}l@{}}{[}{[}2, 2, 2{]}, {[}2, 2, 2{]}, {[}2, 2, 2{]}{]}\end{tabular} & - \\
            Convolution kernel sizes: & \begin{tabular}[c]{@{}l@{}}{[}{[}3, 3{]}, {[}3, 3{]}, {[}3, 3{]}, \\ {[}3, 3{]}{]}\end{tabular} & \begin{tabular}[c]{@{}l@{}}{[}{[}3, 3, 3{]}, {[}3, 3, 3{]}, {[}3, 3, 3{]},\\ {[}3, 3, 3{]}{]}\end{tabular} & - \\ [4pt]
        \end{tabular}}
        \caption{Network configurations generated by nnU-Net for the Hippocampus dataset from the Medical Segmentation Decathlon (D4). For more information on how to decode downsampling strides and kernel sizes into an architecture, see \ref{supplement:architecture_decoding}}
    \end{table}
    
    \begin{table}[H]
        \renewcommand{\arraystretch}{1.2}
        \footnotesize
        \centering
        \begin{tabular}{llll}
            & Anterior& Posterior& mean\\
            \toprule 
            2D& 0.8787& 0.8595& 0.8691\\
            3D\_fullres *& 0.8975& 0.8807& 0.8891\\
            Best Ensemble & 0.8962& 0.8790& 0.8876\\
            Postprocessed & 0.8975& 0.8807& 0.8891\\
            Test set & 0.90& 0.89& 0.895\\[4pt] 
        \end{tabular}
        \caption{Decathlon Hippocampus (D4) results. Note that all reported Dice scores (except the test set) were computed using five fold cross-validation on the training cases. * marks the best performing model selected for subsequent postprocessing (see "Postprocessed") and test set submission (see "Test set"). Note that the Dice scores for the test set are computed with the online platform and only two significant digits are reported. Best ensemble on this dataset was the combination of the 2D U-Net and the 3D full resolution U-Net.}
    \end{table}

\subsubsection*{Prostate (D5)}
    \textbf{Normalization:} Each image is normalized independently by subtracting its mean and dividing by its standard deviation.
    
    \begin{table}[H]
        \renewcommand{\arraystretch}{1.2}
        \centering
        \footnotesize
        \scalebox{0.9}{
        \begin{tabular}{llll}
            \textbf{}  & \textbf{2D U-Net}   & \textbf{3D full resolution U-Net}  & \textbf{3D low resolution U-Net}  \\
            \toprule 
            Target spacing (mm): & NA x 0.62 x 0.62  & 3.6 x 0.62 x 0.62  & -  \\
            \begin{tabular}[c]{@{}l@{}}Median image shape at \\ target spacing:\end{tabular} & NA x 320 x 319   & 20 x 320 x 319  & -  \\
            Patch size:  & 320 x 320   & 20 x 320 x 256  & -  \\
            Batch size:  & 32   & 2   & -   \\
            Downsampling strides: & \begin{tabular}[c]{@{}l@{}}{[}{[}2, 2{]}, {[}2, 2{]}, {[}2, 2{]}, {[}2, 2{]},\\ {[}2, 2{]}, {[}2, 2{]}{]}\end{tabular} & \begin{tabular}[c]{@{}l@{}}{[}{[}1, 2, 2{]}, {[}1, 2, 2{]}, {[}2, 2, 2{]}, \\ {[}2, 2, 2{]}, {[}1, 2, 2{]}, {[}1, 2, 2{]}{]}\end{tabular} & - \\
            Convolution kernel sizes: & \begin{tabular}[c]{@{}l@{}}{[}{[}3, 3{]}, {[}3, 3{]}, {[}3, 3{]}, {[}3, 3{]}, \\ {[}3, 3{]}, {[}3, 3{]}, {[}3, 3{]}{]}\end{tabular} & \begin{tabular}[c]{@{}l@{}}{[}{[}1, 3, 3{]}, {[}1, 3, 3{]}, {[}3, 3, 3{]},\\ {[}3, 3, 3{]},  {[}3, 3, 3{]}, {[}3, 3, 3{]},\\ {[}3, 3, 3{]}{]}\end{tabular} & - \\ [4pt]
        \end{tabular}}
        \caption{Network configurations generated by nnU-Net for the Prostate dataset from the Medical Segmentation Decathlon (D5). For more information on how to decode downsampling strides and kernel sizes into an architecture, see \ref{supplement:architecture_decoding}}
    \end{table}
    
    \begin{table}[H]
        \renewcommand{\arraystretch}{1.2}
        \footnotesize
        \centering
        \begin{tabular}{llll}
             & PZ& TZ& mean\\
            \toprule 
            2D& 0.6285& 0.8380& 0.7333\\
            3D\_fullres& 0.6663& 0.8410& 0.7537\\
            Best Ensemble *& 0.6611 & 0.8575& 0.7593\\
            Postprocessed & 0.6611 & 0.8577& 0.7594\\
            Test set & 0.77 & 0.90 & 0.835\\[4pt] 
        \end{tabular}
        \caption{Decathlon Prostate (D5) results. Note that all reported Dice scores (except the test set) were computed using five fold cross-validation on the training cases. * marks the best performing model selected for subsequent postprocessing (see "Postprocessed") and test set submission (see "Test set"). Note that the Dice scores for the test set are computed with the online platform and only two significant digits are reported. Best ensemble on this dataset was the combination of the 2D U-Net and the 3D full resolution U-Net.}
    \end{table}

\subsubsection*{Lung (D6)}
    \textbf{Normalization:} Clip to $[-1024, 325]$, then subtract $-158.58$ and finally divide by $324.70$.

    \begin{table}[H]
        \renewcommand{\arraystretch}{1.2}
        \centering
        \footnotesize
        \scalebox{0.9}{
        \begin{tabular}{llll}
            \textbf{}  & \textbf{2D U-Net}   & \textbf{3D full resolution U-Net}  & \textbf{3D low resolution U-Net}  \\
            \toprule 
            Target spacing (mm): & NA x 0.79 x 0.79 & 1.24 x 0.79 x 0.79 & 2.35 x 1.48 x 1.48    \\
            \begin{tabular}[c]{@{}l@{}}Median image shape at \\ target spacing:\end{tabular} & NA x 512 x 512  & 252 x 512 x 512 & 133 x 271 x 271    \\
            Patch size:  & 512 x 512   & 80 x 192 x 160  & 80 x 192 x 160   \\
            Batch size:  & 12   & 2   & 2   \\
            Downsampling strides: & \begin{tabular}[c]{@{}l@{}}{[}{[}2, 2{]}, {[}2, 2{]}, {[}2, 2{]}, {[}2, 2{]},\\ {[}2, 2{]}, {[}2, 2{]}, {[}2, 2{]}{]}\end{tabular} & \begin{tabular}[c]{@{}l@{}}{[}{[}2, 2, 2{]}, {[}2, 2, 2{]}, {[}2, 2, 2{]}, \\ {[}2, 2, 2{]}, {[}1, 2, 2{]}{]}\end{tabular} & \begin{tabular}[c]{@{}l@{}}{[}{[}2, 2, 2{]}, {[}2, 2, 2{]}, {[}2, 2, 2{]}, \\ {[}2, 2, 2{]}, {[}1, 2, 2{]}{]}\end{tabular} \\
            Convolution kernel sizes: & \begin{tabular}[c]{@{}l@{}}{[}{[}3, 3{]}, {[}3, 3{]}, {[}3, 3{]}, {[}3, 3{]}, \\ {[}3, 3{]}, {[}3, 3{]}, {[}3, 3{]}, {[}3, 3{]}{]}\end{tabular} & \begin{tabular}[c]{@{}l@{}}{[}{[}3, 3, 3{]}, {[}3, 3, 3{]}, {[}3, 3, 3{]}, \\ {[}3, 3, 3{]}, {[}3, 3, 3{]}, {[}3, 3, 3{]}{]}\end{tabular} & \begin{tabular}[c]{@{}l@{}}{[}{[}3, 3, 3{]}, {[}3, 3, 3{]}, {[}3, 3, 3{]}, \\ {[}3, 3, 3{]}, {[}3, 3, 3{]}, {[}3, 3, 3{]}{]}\\[4pt] \end{tabular}
        \end{tabular}}
        \caption{Network configurations generated by nnU-Net for the Lung dataset from the Medical Segmentation Decathlon (D6). For more information on how to decode downsampling strides and kernel sizes into an architecture, see \ref{supplement:architecture_decoding}}
    \end{table}
 
  \begin{table}[H]
     \renewcommand{\arraystretch}{1.2}
     \footnotesize
     \centering
     \begin{tabular}{lll}
         & cancer& mean\\
         \toprule 
         2D& 0.4989& 0.4989\\
         3D\_fullres& 0.7211& 0.7211\\
         3D\_lowres& 0.7109& 0.7109\\
         3D cascade& 0.6980& 0.6980\\
         Best Ensemble*& 0.7241& 0.7241\\
         Postprocessed & 0.7241& 0.7241\\[4pt]  
         Test set & 0.74& 0.74\\[4pt] 
     \end{tabular}
     \caption{Decathlon Lung (D6) results. Note that all reported Dice scores (except the test set) were computed using five fold cross-validation on the training cases. * marks the best performing model selected for subsequent postprocessing (see "Postprocessed") and test set submission (see "Test set"). Note that the Dice scores for the test set are computed with the online platform and only two significant digits are reported. Best ensemble on this dataset was the combination of the 3D low resolution U-Net and the 3D full resolution U-Net.}
 \end{table}

\subsubsection*{Pancreas (D7)}
    \textbf{Normalization:} Clip to $[-96.0, 215.0]$, then subtract $77.99$ and finally divide by $75.40$.

    \begin{table}[H]
          \renewcommand{\arraystretch}{1.2}
         \centering
         \footnotesize
         \scalebox{0.9}{
         \begin{tabular}{llll}
             \textbf{}  & \textbf{2D U-Net}   & \textbf{3D full resolution U-Net}  & \textbf{3D low resolution U-Net}  \\
             \toprule 
             Target spacing (mm): & NA x 0.8 x 0.8  & 2.5 x 0.8 x 0.8  & 2.58 x 1.29 x 1.29  \\
             \begin{tabular}[c]{@{}l@{}}Median image shape at \\ target spacing:\end{tabular} & NA x 512 x 512   & 96 x 512 x 512  & 93 x 318 x 318  \\
             Patch size:  & 512 x 512   & 40 x 224 x 224  & 64 x 192 x 192  \\
             Batch size:  & 12   & 2   & 2   \\
             Downsampling strides: & \begin{tabular}[c]{@{}l@{}}{[}{[}2, 2{]}, {[}2, 2{]}, {[}2, 2{]}, {[}2, 2{]},\\ {[}2, 2{]}, {[}2, 2{]}, {[}2, 2{]}{]}\end{tabular} & \begin{tabular}[c]{@{}l@{}}{[}{[}1, 2, 2{]}, {[}2, 2, 2{]}, {[}2, 2, 2{]}, \\ {[}2, 2, 2{]}, {[}1, 2, 2{]}{]}\end{tabular} & \begin{tabular}[c]{@{}l@{}}{[}{[}1, 2, 2{]}, {[}2, 2, 2{]}, {[}2, 2, 2{]}, \\ {[}2, 2, 2{]}, {[}2, 2, 2{]}{]}\end{tabular} \\
             Convolution kernel sizes: & \begin{tabular}[c]{@{}l@{}}{[}{[}3, 3{]}, {[}3, 3{]}, {[}3, 3{]}, {[}3, 3{]}, \\ {[}3, 3{]}, {[}3, 3{]}, {[}3, 3{]}, {[}3, 3{]}{]}\end{tabular} & \begin{tabular}[c]{@{}l@{}}{[}{[}1, 3, 3{]}, {[}3, 3, 3{]}, {[}3, 3, 3{]}, \\ {[}3, 3, 3{]}, {[}3, 3, 3{]}, {[}3, 3, 3{]}{]}\end{tabular} & \begin{tabular}[c]{@{}l@{}}{[}{[}3, 3, 3{]}, {[}3, 3, 3{]}, {[}3, 3, 3{]}, \\ {[}3, 3, 3{]}, {[}3, 3, 3{]}, {[}3, 3, 3{]}{]}\\[4pt] \end{tabular}
         \end{tabular}}
         \caption{Network configurations generated by nnU-Net for the Pancreas dataset from the Medical Segmentation Decathlon (D7). For more information on how to decode downsampling strides and kernel sizes into an architecture, see \ref{supplement:architecture_decoding}}
     \end{table}
 
    \begin{table}[H]
    \renewcommand{\arraystretch}{1.2}
    \footnotesize
    \centering
    \begin{tabular}{llll}
         & pancreas& cancer& mean\\
         \toprule 
         2D& 0.7738 & 0.3501& 0.5619\\
         3D\_fullres & 0.8217& 0.5274& 0.6745\\
         3D\_lowres & 0.8118& 0.5286& 0.6702\\
         3D cascade & 0.8101& 0.5380& 0.6741\\
         Best Ensemble *& 0.8214& 0.5428& 0.6821\\
         Postprocessed& 0.8214& 0.5428& 0.6821\\[4pt] 
         Test set & 0.82& 0.53& 0.675\\[4pt] 
    \end{tabular}
    \caption{Decathlon Pancreas (D7) results. Note that all reported Dice scores (except the test set) were computed using five fold cross-validation on the training cases. * marks the best performing model selected for subsequent postprocessing (see "Postprocessed") and test set submission (see "Test set"). Note that the Dice scores for the test set are computed with the online platform and only two significant digits are reported. Best ensemble on this dataset was the combination of the 3D full resolution U-Net and the 3D U-Net cascade.}
    \end{table}
    
\subsubsection*{Hepatic Vessel (D8)}
    \textbf{Normalization:} Clip to $[-3, 243]$, then subtract $104.37$ and finally divide by $52.62$.

    \begin{table}[H]
        \renewcommand{\arraystretch}{1.2}
        \centering
        \footnotesize
        \scalebox{0.9}{
        \begin{tabular}{llll}
            \textbf{}  & \textbf{2D U-Net}   & \textbf{3D full resolution U-Net}  & \textbf{3D low resolution U-Net}  \\
            \toprule 
            Target spacing (mm): & NA x 0.8 x 0.8  & 1.5 x 0.8 x 0.8  & 2.42 x 1.29 x 1.29  \\
            \begin{tabular}[c]{@{}l@{}}Median image shape at \\ target spacing:\end{tabular} & NA x 512 x 512   & 150 x 512 x 512  & 93 x 318 x 318  \\
            Patch size:  & 512 x 512   & 64 x 192 x 192  & 64 x 192 x 192  \\
            Batch size:  & 12   & 2   & 2   \\
            Downsampling strides: & \begin{tabular}[c]{@{}l@{}}{[}{[}2, 2{]}, {[}2, 2{]}, {[}2, 2{]}, {[}2, 2{]},\\ {[}2, 2{]}, {[}2, 2{]}, {[}2, 2{]}{]}\end{tabular} & \begin{tabular}[c]{@{}l@{}}{[}{[}2, 2, 2{]}, {[}2, 2, 2{]}, {[}2, 2, 2{]}, \\ {[}2, 2, 2{]}, {[}1, 2, 2{]}{]}\end{tabular} & \begin{tabular}[c]{@{}l@{}}{[}{[}2, 2, 2{]}, {[}2, 2, 2{]}, {[}2, 2, 2{]}, \\ {[}2, 2, 2{]}, {[}1, 2, 2{]}{]}\end{tabular} \\
            Convolution kernel sizes: & \begin{tabular}[c]{@{}l@{}}{[}{[}3, 3{]}, {[}3, 3{]}, {[}3, 3{]}, {[}3, 3{]}, \\ {[}3, 3{]}, {[}3, 3{]}, {[}3, 3{]}, {[}3, 3{]}{]}\end{tabular} & \begin{tabular}[c]{@{}l@{}}{[}{[}3, 3, 3{]}, {[}3, 3, 3{]}, {[}3, 3, 3{]}, \\ {[}3, 3, 3{]}, {[}3, 3, 3{]}, {[}3, 3, 3{]}{]}\end{tabular} & \begin{tabular}[c]{@{}l@{}}{[}{[}3, 3, 3{]}, {[}3, 3, 3{]}, {[}3, 3, 3{]}, \\ {[}3, 3, 3{]}, {[}3, 3, 3{]}, {[}3, 3, 3{]}{]}\\[4pt] \end{tabular}
        \end{tabular}}
        \caption{Network configurations generated by nnU-Net for the HepaticVessel dataset from the Medical Segmentation Decathlon (D8). For more information on how to decode downsampling strides and kernel sizes into an architecture, see \ref{supplement:architecture_decoding}}
    \end{table}
 
    \begin{table}[H]
    \renewcommand{\arraystretch}{1.2}
    \footnotesize
    \centering
    \begin{tabular}{llll}
         & Vessel& Tumour& mean\\
         \toprule 
         2D& 0.6180& 0.6359& 0.6269\\
         3D\_fullres & 0.6456& 0.7217& 0.6837\\
         3D\_lowres & 0.6294& 0.7079& 0.6687\\
         3D cascade & 0.6424& 0.7138& 0.6781\\
         Best Ensemble *& 0.6485& 0.7250& 0.6867\\
         Postprocessed& 0.6485& 0.7250& 0.6867\\[4pt] 
         Test set & 0.66& 0.72& 0.69\\[4pt] 
    \end{tabular}
    \caption{Decathlon HepaticVessel (D8) results. Note that all reported Dice scores (except the test set) were computed using five fold cross-validation on the training cases. * marks the best performing model selected for subsequent postprocessing (see "Postprocessed") and test set submission (see "Test set"). Note that the Dice scores for the test set are computed with the online platform and only two significant digits are reported. Best ensemble on this dataset was the combination of the 3D full resolution U-Net and the 3D low resolution U-Net.}
    \end{table}

\subsubsection*{Spleen (D9)}
    \textbf{Normalization:} Clip to $[-41, 176]$, then subtract $99.29$ and finally divide by $39.47$.

    \begin{table}[H]
        \renewcommand{\arraystretch}{1.2}
        \centering
        \footnotesize
        \scalebox{0.9}{
        \begin{tabular}{llll}
            \textbf{}  & \textbf{2D U-Net}   & \textbf{3D full resolution U-Net}  & \textbf{3D low resolution U-Net}  \\
            \toprule 
            Target spacing (mm): & NA x 0.79 x 0.79  & 1.6 x 0.79 x 0.79  & 2.77 x 1.38 x 1.38  \\
            \begin{tabular}[c]{@{}l@{}}Median image shape at \\ target spacing:\end{tabular} & NA x 512 x 512   & 187 x 512 x 512  & 108 x 293 x 293  \\
            Patch size:  & 512 x 512   & 64 x 192 x 160  & 64 x 192 x 192  \\
            Batch size:  & 12   & 2   & 2   \\
            Downsampling strides: & \begin{tabular}[c]{@{}l@{}}{[}{[}2, 2{]}, {[}2, 2{]}, {[}2, 2{]}, {[}2, 2{]},\\ {[}2, 2{]}, {[}2, 2{]}, {[}2, 2{]}{]}\end{tabular} & \begin{tabular}[c]{@{}l@{}}{[}{[}1, 2, 2{]}, {[}2, 2, 2{]}, {[}2, 2, 2{]}, \\ {[}2, 2, 2{]}, {[}2, 2, 2{]}{]}\end{tabular} & \begin{tabular}[c]{@{}l@{}}{[}{[}2, 2, 2{]}, {[}2, 2, 2{]}, {[}2, 2, 2{]}, \\ {[}2, 2, 2{]}, {[}1, 2, 2{]}{]}\end{tabular} \\
            Convolution kernel sizes: & \begin{tabular}[c]{@{}l@{}}{[}{[}3, 3{]}, {[}3, 3{]}, {[}3, 3{]}, {[}3, 3{]}, \\ {[}3, 3{]}, {[}3, 3{]}, {[}3, 3{]}, {[}3, 3{]}{]}\end{tabular} & \begin{tabular}[c]{@{}l@{}}{[}{[}1, 3, 3{]}, {[}3, 3, 3{]}, {[}3, 3, 3{]}, \\ {[}3, 3, 3{]}, {[}3, 3, 3{]}, {[}3, 3, 3{]}{]}\end{tabular} & \begin{tabular}[c]{@{}l@{}}{[}{[}3, 3, 3{]}, {[}3, 3, 3{]}, {[}3, 3, 3{]}, \\ {[}3, 3, 3{]}, {[}3, 3, 3{]}, {[}3, 3, 3{]}{]}\\[4pt] \end{tabular}
        \end{tabular}}
        \caption{Network configurations generated by nnU-Net for the Spleen dataset from the Medical Segmentation Decathlon (D9). For more information on how to decode downsampling strides and kernel sizes into an architecture, see \ref{supplement:architecture_decoding}}
    \end{table}
    
    \begin{table}[H]
        \renewcommand{\arraystretch}{1.2}
        \footnotesize
        \centering
        \begin{tabular}{lll}
            & spleen& mean\\
            \toprule 
            2D& 0.9492& 0.9492\\
            3D\_fullres& 0.9638& 0.9638\\
            3D\_lowres& 0.9683& 0.9683\\
            3D cascade& 0.9714& 0.9714\\
            Best Ensemble *& 0.9723& 0.9723\\
            Postprocessed & 0.9724& 0.9724\\[4pt]  
            Test set & 0.97& 0.97\\[4pt] 
        \end{tabular}
        \caption{Decathlon Spleen (D9) results. Note that all reported Dice scores (except the test set) were computed using five fold cross-validation on the training cases. * marks the best performing model selected for subsequent postprocessing (see "Postprocessed") and test set submission (see "Test set"). Note that the Dice scores for the test set are computed with the online platform and only two significant digits are reported. Best ensemble on this dataset was the combination of the 3D U-Net cascade and the 3D full resolution U-Net.}
    \end{table}

\subsubsection*{Colon (D10)}
    \textbf{Normalization:} Clip to $[-30.0, 165.82]$, then subtract $62.18$ and finally divide by $32.65$.

    \begin{table}[H]
        \renewcommand{\arraystretch}{1.2}
        \centering
        \footnotesize
        \scalebox{0.9}{
        \begin{tabular}{llll}
            \textbf{}  & \textbf{2D U-Net}   & \textbf{3D full resolution U-Net}  & \textbf{3D low resolution U-Net}  \\
            \toprule 
            Target spacing (mm): & NA x 0.78 x 0.78  & 3 x 0.78 x 0.78  & 3.09 x 1.55 x 1.55  \\
            \begin{tabular}[c]{@{}l@{}}Median image shape at \\ target spacing:\end{tabular} & NA x 512 x 512   & 150 x 512 x 512  & 146 x 258 x 258  \\
            Patch size:  & 512 x 512   & 56 x 192 x 160  & 96 x 160 x 160  \\
            Batch size:  & 12   & 2   & 2   \\
            Downsampling strides: & \begin{tabular}[c]{@{}l@{}}{[}{[}2, 2{]}, {[}2, 2{]}, {[}2, 2{]}, {[}2, 2{]},\\ {[}2, 2{]}, {[}2, 2{]}, {[}2, 2{]}{]}\end{tabular} & \begin{tabular}[c]{@{}l@{}}{[}{[}1, 2, 2{]}, {[}2, 2, 2{]}, {[}2, 2, 2{]}, \\ {[}2, 2, 2{]}, {[}1, 2, 2{]}{]}\end{tabular} & \begin{tabular}[c]{@{}l@{}}{[}{[}2, 2, 2{]}, {[}2, 2, 2{]}, {[}2, 2, 2{]}, \\ {[}2, 2, 2{]}, {[}1, 2, 2{]}{]}\end{tabular} \\
            Convolution kernel sizes: & \begin{tabular}[c]{@{}l@{}}{[}{[}3, 3{]}, {[}3, 3{]}, {[}3, 3{]}, {[}3, 3{]}, \\ {[}3, 3{]}, {[}3, 3{]}, {[}3, 3{]}, {[}3, 3{]}{]}\end{tabular} & \begin{tabular}[c]{@{}l@{}}{[}{[}1, 3, 3{]}, {[}3, 3, 3{]}, {[}3, 3, 3{]}, \\ {[}3, 3, 3{]}, {[}3, 3, 3{]}, {[}3, 3, 3{]}{]}\end{tabular} & \begin{tabular}[c]{@{}l@{}}{[}{[}3, 3, 3{]}, {[}3, 3, 3{]}, {[}3, 3, 3{]}, \\ {[}3, 3, 3{]}, {[}3, 3, 3{]}, {[}3, 3, 3{]}{]}\\[4pt] \end{tabular}
        \end{tabular}}
        \caption{Network configurations generated by nnU-Net for the Colon dataset from the Medical Segmentation Decathlon (D10). For more information on how to decode downsampling strides and kernel sizes into an architecture, see \ref{supplement:architecture_decoding}}
    \end{table}

     \begin{table}[H]
        \renewcommand{\arraystretch}{1.2}
        \footnotesize
        \centering
        \begin{tabular}{lll}
            & colon cancer primaries& mean\\
            \toprule 
            2D& 0.2852& 0.2852\\
            3D\_fullres& 0.4553& 0.4553\\
            3D\_lowres& 0.4538& 0.4538\\
            3D cascade * & 0.4937& 0.4937\\
            Best Ensemble & 0.4853& 0.4853\\
            Postprocessed & 0.4937& 0.4937\\[4pt]  
            Test set & 0.58& 0.58\\[4pt] 
        \end{tabular}
        \caption{Decathlon Colon (D10) results. Note that all reported Dice scores (except the test set) were computed using five fold cross-validation on the training cases. * marks the best performing model selected for subsequent postprocessing (see "Postprocessed") and test set submission (see "Test set"). Note that the Dice scores for the test set are computed with the online platform and only two significant digits are reported. Best ensemble on this dataset was the combination of the 3D U-Net cascade and the 3D full resolution U-Net.}
    \end{table}

\subsection{Multi Atlas Labeling Beyond the Cranial Vault: Abdomen (D11)}
 
    \paragraph{Challenge summary}
    The Multi Atlas Labeling Beyond the Cranial Vault - Abdomen Challenge\footnote{\url{https://www.synapse.org/Synapse:syn3193805/wiki/217752}} \citeSupplement{SM_BCVAbdomenChallenge} (denoted BCV for brevity) comprises 30 CT images for training and 20 for testing. The segmentation target are thirteen different organs in the abdomen. 
    
    \paragraph{Application of nnU-Net to BCV}
    nnU-Net was applied to the BCV challenge without any manual intervention.
    
    \textbf{Normalization:} Clip to $[-958, 327]$, then subtract $82.92$ and finally divide by $136.97$.
    
    \begin{table}[H]
        \renewcommand{\arraystretch}{1.2}
        
        \footnotesize
        \scalebox{0.9}{
        \begin{tabular}{llll}
            \textbf{}  & \textbf{2D U-Net}   & \textbf{3D full resolution U-Net}  & \textbf{3D low resolution U-Net}  \\
            \toprule 
            Target spacing (mm): & NA x 0.76 x 0.76  & 3 x 0.76 x 0.76  & 3.18 x 1.60 x 1.60  \\
            \begin{tabular}[c]{@{}l@{}}Median image shape at \\ target spacing:\end{tabular} & NA x 512 x 512   & 148 x 512 x 512  & 140 x 243 x 243  \\
            Patch size:  & 512 x 512   & 48 x 192 x 192  & 80 x 160 x 160  \\
            Batch size:  & 12   & 2   & 2   \\
            Downsampling strides: & \begin{tabular}[c]{@{}l@{}}{[}{[}2, 2{]}, {[}2, 2{]}, {[}2, 2{]}, {[}2, 2{]},\\ {[}2, 2{]}, {[}2, 2{]}, {[}2, 2{]}{]}\end{tabular} & \begin{tabular}[c]{@{}l@{}}{[}{[}1, 2, 2{]}, {[}2, 2, 2{]}, {[}2, 2, 2{]}, \\ {[}2, 2, 2{]}, {[}1, 2, 2{]}{]}\end{tabular} & \begin{tabular}[c]{@{}l@{}}{[}{[}2, 2, 2{]}, {[}2, 2, 2{]}, {[}2, 2, 2{]}, \\ {[}2, 2, 2{]}, {[}1, 2, 2{]}{]}\end{tabular} \\
            Convolution kernel sizes: & \begin{tabular}[c]{@{}l@{}}{[}{[}3, 3{]}, {[}3, 3{]}, {[}3, 3{]}, {[}3, 3{]}, \\ {[}3, 3{]}, {[}3, 3{]}, {[}3, 3{]}, {[}3, 3{]}{]}\end{tabular} & \begin{tabular}[c]{@{}l@{}}{[}{[}1, 3, 3{]}, {[}3, 3, 3{]}, {[}3, 3, 3{]}, \\ {[}3, 3, 3{]}, {[}3, 3, 3{]}, {[}3, 3, 3{]}{]}\end{tabular} & \begin{tabular}[c]{@{}l@{}}{[}{[}3, 3, 3{]}, {[}3, 3, 3{]}, {[}3, 3, 3{]}, \\ {[}3, 3, 3{]}, {[}3, 3, 3{]}, {[}3, 3, 3{]}{]}\\[4pt] \end{tabular}
        \end{tabular}}
        \caption{Network configurations generated by nnU-Net for the BCV challenge (D131. For more information on how to decode downsampling strides and kernel sizes into an architecture, see \ref{supplement:architecture_decoding}}
    \end{table}
    
    \begin{table}[H]
        \renewcommand{\arraystretch}{1.2}
        \footnotesize
       
        \begin{tabular}{lllllllllllllll}
            & 1& 2& 3& 4& 5& 6& 7& 8\\
            \toprule 
            2D&               0.8860& 0.8131& 0.8357& 0.6406& 0.7724& 0.9453& 0.8405& 0.9128\\
            3D\_fullres&      0.9083& 0.8939& 0.8675& 0.6632& 0.7840& 0.9557& 0.8816& 0.9229\\
            3D\_lowres&       0.9132& 0.9045& 0.9132& 0.6525& 0.7810& 0.9554& 0.8903& 0.9209\\
            3D cascade&       0.9166& 0.9069& 0.9137& 0.7036& 0.7885& 0.9587& 0.9037& 0.9215\\
            Best Ensemble * & 0.9135& 0.9065& 0.8971& 0.6955& 0.7897& 0.9589& 0.9026& 0.9248\\
            Postprocessed&    0.9135& 0.9065& 0.8971& 0.6959& 0.7897& 0.9590& 0.9026& 0.9248\\[4pt]  
            Test set &        0.9721& 0.9182& 0.9578& 0.7528& 0.8411& 0.9769& 0.9220& 0.9290\\[4pt] 
        \end{tabular}
        \begin{tabular}{lllllllllllllll}
            & 9& 10& 11& 12& 13& mean\\
            \toprule 
            2D&               0.8140& 0.7046& 0.7367& 0.6269& 0.5909& 0.7784\\
            3D\_fullres&      0.8638& 0.7659& 0.8176& 0.7148& 0.7238& 0.8279\\
            3D\_lowres&       0.8571& 0.7469& 0.8003& 0.6688& 0.6851& 0.8223\\
            3D cascade&       0.8621& 0.7722& 0.8210& 0.7205& 0.7214& 0.8393\\
            Best Ensemble * & 0.8673& 0.7746& 0.8299& 0.7218& 0.7287& 0.8393\\
            Postprocessed&    0.8673& 0.7746& 0.8299& 0.7262& 0.7290& 0.8397\\[4pt]  
            Test set &        0.8809& 0.8317& 0.8515& 0.7887& 0.7674& 0.8762\\[4pt] 
        \end{tabular}
        \caption{Multi Atlas Labeling Beyond the Cranial Vault Abdomen (D11) results. Note that all reported Dice scores (except the test set) were computed using five fold cross-validation on the training cases. Postprocessing was applied to the model marked with *. This model (incl postprocessing) was used for test set predictions. Note that the Dice scores for the test set are computed with the online platform. Best ensemble on this dataset was the combination of the 3D U-Net cascade and the 3D full resolution U-Net.}
    \end{table}

\subsection{PROMISE12 (D12)}

    \paragraph{Challenge summary}
    The segmentation target of the PROMISE12 challenge \citeSupplement{SM_promise12Challenge} is the prostate in T2 MRI images. 50 training cases with prostate annotations are provided for training. There are 30 test cases which need to be segmented by the challenge participants and are subsequently evaluated on an online platform\footnote{\url{https://promise12.grand-challenge.org/}}.
    
    \paragraph{Application of nnU-Net to PROMISE12}
    nnU-Net was applied to the PROMISE12 challenge without any manual intervention.
    
    \textbf{Normalization:} Each image is normalized independently by subtracting its mean and dividing by its standard deviation.
    
    \begin{table}[H]
        \renewcommand{\arraystretch}{1.2}
        \centering
        \footnotesize
        \scalebox{0.9}{
        \begin{tabular}{llll}
            \textbf{}  & \textbf{2D U-Net}   & \textbf{3D full resolution U-Net}  & \textbf{3D low resolution U-Net}  \\
            \toprule 
            Target spacing (mm): & NA x 0.61 x 0.61  & 2.2 x 0.61 x 0.61  & -  \\
            \begin{tabular}[c]{@{}l@{}}Median image shape at \\ target spacing:\end{tabular} & NA x 327 x 327   & 39 x 327 x 327  & -  \\
            Patch size:  & 384 x 384   & 28 x 256 x 256  & -  \\
            Batch size:  & 22   & 2   & -   \\
            Downsampling strides: & \begin{tabular}[c]{@{}l@{}}{[}{[}2, 2{]}, {[}2, 2{]}, {[}2, 2{]}, {[}2, 2{]},\\ {[}2, 2{]}, {[}2, 2{]}{]}\end{tabular} & \begin{tabular}[c]{@{}l@{}}{[}{[}1, 2, 2{]}, {[}2, 2, 2{]}, {[}2, 2, 2{]}, \\ {[}1, 2, 2{]}, {[}1, 2, 2{]}{]}\end{tabular} & - \\
            Convolution kernel sizes: & \begin{tabular}[c]{@{}l@{}}{[}{[}3, 3{]}, {[}3, 3{]}, {[}3, 3{]}, {[}3, 3{]}, \\ {[}3, 3{]}, {[}3, 3{]}, {[}3, 3{]}{]}\end{tabular} & \begin{tabular}[c]{@{}l@{}}{[}{[}1, 3, 3{]}, {[}3, 3, 3{]}, {[}3, 3, 3{]}, \\ {[}3, 3, 3{]}, {[}3, 3, 3{]}, {[}3, 3, 3{]}{]}\end{tabular} & - \\[4pt]
        \end{tabular}}
        \caption{Network configurations generated by nnU-Net for the PROMISE12 challenge (D12). For more information on how to decode downsampling strides and kernel sizes into an architecture, see \ref{supplement:architecture_decoding}}
    \end{table}
    
     \begin{table}[H]
        \renewcommand{\arraystretch}{1.2}
        \footnotesize
        \centering
        \begin{tabular}{lll}
            & prostate& mean\\
            \toprule 
            2D& 0.8932& 0.8932\\
            3D\_fullres& 0.8891& 0.8891\\
            Best Ensemble *& 0.9029& 0.9029\\
            Postprocessed& 0.9030& 0.9030\\[4pt]  
            Test set & 0.9194& 0.9194\\[4pt] 
        \end{tabular}
        \caption{PROMISE12 (D12) results. Note that all reported Dice scores (except the test set) were computed using five fold cross-validation on the training cases. * marks the best performing model selected for subsequent postprocessing (see "Postprocessed") and test set submission (see "Test set"). Note that the scores for the test set are computed with the online platform. The evaluation score of our test set submission is 89.6507. The test set Dice score reported in the table was computed from the detailed submission results (Detailed results available here \url{https://promise12.grand-challenge.org/evaluation/results/89044a85-6c13-49f4-9742-dea65013e971/}). Best ensemble on this dataset was the combination of the 2D U-Net and the 3D full resolution U-Net.}
    \end{table}

 \subsection{The Automatic Cardiac Diagnosis Challenge (ACDC) (D13)}
 
    \paragraph{Challenge summary}
    The Automatic Cardiac Diagnosis Challenge\footnote{\url{https://acdc.creatis.insa-lyon.fr}} \citeSupplement{SM_ACDCTMI} (ACDC) comprises 100 training patients and 50 test patients. The target structures are the cavity of the right ventricle, the myocardium of the left ventricle and the cavity of the left ventricle. All images are cine MRI sequences of which the enddiastolic (ED) and endsystolic (ES) time points of the cardiac cycle were to be segmented. With two time instances per patient, the effective number of training/test images is 200/100.
    
    \paragraph{Application of nnU-Net to ACDC}
    Since two time instances of the same patient were provided, we manually interfered with the split for the 5-fold cross-validation of our models to ensure mutual exclusivity of patients between folds. A part from that, nnU-Net was applied without manual intervention.
    
    \textbf{Normalization:} Each image is normalized independently by subtracting its mean and dividing by its standard deviation.
    
    \begin{table}[H]
        \renewcommand{\arraystretch}{1.2}
        \centering
        \footnotesize
        \scalebox{0.9}{
        \begin{tabular}{llll}
            \textbf{}  & \textbf{2D U-Net}   & \textbf{3D full resolution U-Net}  & \textbf{3D low resolution U-Net}  \\
            \toprule 
            Target spacing (mm): & NA x 1.56 x 1.56  & 5 x 1.56 x 1.56  & -  \\
            \begin{tabular}[c]{@{}l@{}}Median image shape at \\ target spacing:\end{tabular} & NA x 237 x 208   & 18 x 237 x 208  & -  \\
            Patch size:  & 256 x 224   & 20 x 256 x 224  & -  \\
            Batch size:  & 58   & 3   & -   \\
            Downsampling strides: & \begin{tabular}[c]{@{}l@{}}{[}{[}2, 2{]}, {[}2, 2{]}, {[}2, 2{]}, {[}2, 2{]},\\ {[}2, 2{]}{]}\end{tabular} & \begin{tabular}[c]{@{}l@{}}{[}{[}1, 2, 2{]}, {[}2, 2, 2{]}, {[}2, 2, 2{]}, \\ {[}1, 2, 2{]}, {[}1, 2, 2{]}{]}\end{tabular} & - \\
            Convolution kernel sizes: & \begin{tabular}[c]{@{}l@{}}{[}{[}3, 3{]}, {[}3, 3{]}, {[}3, 3{]}, {[}3, 3{]}, \\ {[}3, 3{]}, {[}3, 3{]}{]}\end{tabular} & \begin{tabular}[c]{@{}l@{}}{[}{[}1, 3, 3{]}, {[}3, 3, 3{]}, {[}3, 3, 3{]}, \\ {[}3, 3, 3{]}, {[}3, 3, 3{]}, {[}3, 3, 3{]}{]}\end{tabular} & - \\[4pt]
        \end{tabular}}
        \caption{Network configurations generated by nnU-Net for the ACDC challenge (D13). For more information on how to decode downsampling strides and kernel sizes into an architecture, see \ref{supplement:architecture_decoding}}
    \end{table}
    
    \begin{table}[H]
        \renewcommand{\arraystretch}{1.2}
        \footnotesize
        \centering
        \begin{tabular}{lllll}
            & RV& MLV& LVC& mean\\
            \toprule 
            2D& 0.9053& 0.8991& 0.9433& 0.9159\\
            3D\_fullres& 0.9059& 0.9022& 0.9458& 0.9179\\
            Best Ensemble *& 0.9145& 0.9059& 0.9479& 0.9227\\
            Postprocessed & 0.9145& 0.9059& 0.9479& 0.9228\\
            Test set & 0.9295 & 0.9183 & 0.9407 & 0.9295\\[4pt] 
        \end{tabular}
        \caption{ACDC results (D13). Note that all reported Dice scores (except the test set) were computed using five fold cross-validation on the training cases. * marks the best performing model selected for subsequent postprocessing (see "Postprocessed") and test set submission (see "Test set"). Note that the Dice scores for the test set are computed with the online platform. The online platform reports the Dice scores for enddiastolic and endsystolic time points separately. We averaged these values for a more condensed presentation. Best ensemble on this dataset was the combination of the 2D U-Net and the 3D full resolution U-Net.}
    \end{table}

    \subsection{Liver and Liver Tumor Segmentation Challenge (LiTS) (D14)}
    
    \paragraph{Challenge summary}
    The Liver and Liver Tumor Segmentation challenge \citeSupplement{SM_litsPaper} provides 131 training CT images with ground truth annotations for the liver and liver tumors. 70 test images are provided without annotations. The predicted segmentation masks of the test cases are evaluated using the LiTS online platform\footnote{\url{https://competitions.codalab.org/competitions/17094}}.
    
    \paragraph{Application of nnU-Net to LiTS}
    nnU-Net was applied to the LiTS challenge without any manual intervention.
    
    \textbf{Normalization:} Clip to $[-17, 201]$, then subtract $99.40$ and finally divide by $39.39$.

    \begin{table}[H]
        \renewcommand{\arraystretch}{1.2}
        \centering
        \footnotesize
        \scalebox{0.9}{
        \begin{tabular}{llll}
        \textbf{}  & \textbf{2D U-Net}   & \textbf{3D full resolution U-Net}  & \textbf{3D low resolution U-Net}  \\
        \toprule 
        Target spacing (mm): & NA x 0.77 x 0.77  & 1 x 0.77 x 0.77  & 2.47 x 1.90 x 1.90  \\
        \begin{tabular}[c]{@{}l@{}}Median image shape at \\ target spacing:\end{tabular} & NA x 512 x 512   & 482 x 512 x 512  & 195 x 207 x 207  \\
            Patch size:  & 512 x 512   & 128 x 128 x 128  & 128 x 128 x 128  \\
            Batch size:  & 12   & 2   & 2   \\
            Downsampling strides: & \begin{tabular}[c]{@{}l@{}}{[}{[}2, 2{]}, {[}2, 2{]}, {[}2, 2{]}, {[}2, 2{]},\\ {[}2, 2{]}, {[}2, 2{]}, {[}2, 2{]}{]}\end{tabular} & \begin{tabular}[c]{@{}l@{}}{[}{[}2, 2, 2{]}, {[}2, 2, 2{]}, {[}2, 2, 2{]}, \\ {[}2, 2, 2{]}, {[}2, 2, 2{]}{]}\end{tabular} & \begin{tabular}[c]{@{}l@{}}{[}{[}2, 2, 2{]}, {[}2, 2, 2{]}, {[}2, 2, 2{]}, \\ {[}2, 2, 2{]}, {[}2, 2, 2{]}{]}\end{tabular} \\
            Convolution kernel sizes: & \begin{tabular}[c]{@{}l@{}}{[}{[}3, 3{]}, {[}3, 3{]}, {[}3, 3{]}, {[}3, 3{]}, \\ {[}3, 3{]}, {[}3, 3{]}, {[}3, 3{]}, {[}3, 3{]}{]}\end{tabular} & \begin{tabular}[c]{@{}l@{}}{[}{[}3, 3, 3{]}, {[}3, 3, 3{]}, {[}3, 3, 3{]}, \\ {[}3, 3, 3{]}, {[}3, 3, 3{]}, {[}3, 3, 3{]}{]}\end{tabular} & \begin{tabular}[c]{@{}l@{}}{[}{[}3, 3, 3{]}, {[}3, 3, 3{]}, {[}3, 3, 3{]}, \\ {[}3, 3, 3{]}, {[}3, 3, 3{]}, {[}3, 3, 3{]}{]}\\[4pt] \end{tabular}
        \end{tabular}}
        \caption{Network configurations generated by nnU-Net for the LiTS challenge (D14). For more information on how to decode downsampling strides and kernel sizes into an architecture, see \ref{supplement:architecture_decoding}}
    \end{table}
 
     \begin{table}[H]
        \renewcommand{\arraystretch}{1.2}
        \footnotesize
        \centering
        \begin{tabular}{llll}
            & liver& cancer& mean\\
            \toprule 
            2D& 0.9547& 0.5603& 0.7575\\
            3D\_fullres& 0.9576& 0.6253& 0.7914\\
            3D\_lowres& 0.9585& 0.6161& 0.7873\\
            3D cascade& 0.9609& 0.6294& 0.7951\\
            Best Ensemble*& 0.9618& 0.6539& 0.8078\\
            Postprocessed& 0.9631& 0.6543& 0.8087\\[4pt]  
            Test set & 0.9670& 0.7630& 0.8650\\[4pt] 
        \end{tabular}
        \caption{LiTS results (D14). Note that all reported Dice scores (except the test set) were computed using five fold cross-validation on the training cases. * marks the best performing model selected for subsequent postprocessing (see "Postprocessed") and test set submission (see "Test set"). Note that the Dice scores for the test set are computed with the online platform. Best ensemble on this dataset was the combination of the 3D low resolution U-Net and the 3D full resolution U-Net.}
    \end{table}

\subsection{Longitudinal multiple sclerosis lesion segmentation challenge (MSLesion) (D15)}
 
    \paragraph{Challenge summary}
    The longitudinal multiple sclerosis lesion segmentation challenge \citeSupplement{SM_mslesionchallenge} provides 5 training patients. For each patient, 4 to 5 images acquired at different time points are provided (4 patients with 4 time points each and one patient with 5 time points for a total of 21 images). Each time point is annotated by two different experts, resulting in 42 training annotations (on 21 images). The test set contains 14 patients, again with several time points each, for a total of 61 MRI acquisitions. Test set predictions are evaluated using the online platform\footnote{\url{https://smart-stats-tools.org/lesion-challenge}}. Each train and test image consists of four MRI modalities: MPRAGE, FLAIR, Proton Density, T2.
    
    \paragraph{Application of nnU-Net to MSLesion}
    We manually interfere with the splits in the cross-validation to ensure mutual exclusivity of patients between folds. Each image was annotated by two different experts. We treat these annotations as separate training images (of the same patient), resulting in a training set size of $2\times 21 = 42$. We do not use the longitudinal nature of the scans and treat each image individually during training and inference.
    
    \textbf{Normalization:} Each image is normalized independently by subtracting its mean and dividing by its standard deviation.
    
    \begin{table}[H]
        \renewcommand{\arraystretch}{1.2}
        \centering
        \footnotesize
        \scalebox{0.9}{
        \begin{tabular}{llll}
            \textbf{}  & \textbf{2D U-Net}   & \textbf{3D full resolution U-Net}  & \textbf{3D low resolution U-Net}  \\
            \toprule 
            Target spacing (mm): & NA x 1 x 1  & 1 x 1 x 1  & -  \\
            \begin{tabular}[c]{@{}l@{}}Median image shape at \\ target spacing:\end{tabular} & NA x 180 x 137   & 137 x 180 x 137  & -  \\
            Patch size:  & 192 x 160   & 112 x 128 x 96  & -  \\
            Batch size:  & 107   & 2   & -   \\
            Downsampling strides: & \begin{tabular}[c]{@{}l@{}}{[}{[}2, 2{]}, {[}2, 2{]}, {[}2, 2{]}, {[}2, 2{]},\\ {[}2, 2{]}{]}\end{tabular} & \begin{tabular}[c]{@{}l@{}}{[}{[}1, 2, 1{]}, {[}2, 2, 2{]}, {[}2, 2, 2{]}, \\ {[}2, 2, 2{]}, {[}2, 2, 2{]}{]}\end{tabular} & - \\
            Convolution kernel sizes: & \begin{tabular}[c]{@{}l@{}}{[}{[}3, 3{]}, {[}3, 3{]}, {[}3, 3{]}, {[}3, 3{]}, \\ {[}3, 3{]}, {[}3, 3{]}{]}\end{tabular} & \begin{tabular}[c]{@{}l@{}}{[}{[}3, 3, 3{]}, {[}3, 3, 3{]}, {[}3, 3, 3{]}, \\ {[}3, 3, 3{]}, {[}3, 3, 3{]}, {[}3, 3, 3{]}{]}\end{tabular} & - \\[4pt]
        \end{tabular}}
        \caption{Network configurations generated by nnU-Net for the MSLesion challenge (D15). For more information on how to decode downsampling strides and kernel sizes into an architecture, see \ref{supplement:architecture_decoding}}
    \end{table}
    
    \begin{table}[H]
        \renewcommand{\arraystretch}{1.2}
        \footnotesize
        \centering
        \begin{tabular}{lll}
            & lesion& mean\\
            \toprule 
            2D& 0.7339& 0.7339\\
            3D\_fullres *& 0.7531& 0.7531\\
            Best Ensemble& 0.7494& 0.7494\\
            Postprocessed& 0.7531& 0.7531\\[4pt] 
            Test set & 0.6785 & 0.6785\\[4pt] 
        \end{tabular}
        \caption[]{MSLesion results (D15). Note that all reported Dice scores (except the test set) were computed using five fold cross-validation on the training cases. * marks the best performing model selected for subsequent postprocessing (see "Postprocessed") and test set submission (see "Test set"). Note that the Dice scores for the test set are computed with the online platform based on the detailed results (which are available here \url{https://smart-stats-tools.org/sites/lesion_challenge/temp/top25/nnUNetV2_12032019_0903.csv}). The ranking is based on a score, which includes other metrics as well (see \citeSupplement{SM_mslesionchallenge} for details). The score of our submission is 92.874. Best ensemble on this dataset was the combination of the 2D U-Net and the 3D full resolution U-Net.}
    \end{table}

\subsection{Combined Healthy Abdominal Organ Segmentation (CHAOS) (D16)}

    \paragraph{Challenge summary}
    The CHAOS challenge \citeSupplement{SM_kavur2020chaos} is divided into five tasks. Here we focused on Tasks 3 (MRI Liver segmentation) and Task 5 (MRI multiorgan segmentation). Tasks 1, 2 and 4 also included the use of CT images, a modality for which plenty of public data is available (see e.g. BCV and LiTS challenge). To isolate the algorithmic performance of nnU-Net relative to other participants we decided to only use the tasks for which a contamination with external data was unlikely.
    The target structures of Task 5 are the liver, the spleen and the left and right kidneys. The CHAOS challenge provides 20 training cases. For each training case, there is a T2 images with a corresponding ground truth annotation as well as a T1 acquisition with its own, separate ground truth annotation. The T1 acquisition has two modalities which are co-registered: T1 in-phase and T1 out-phase. Task 3 is a subset of Task 5 with only the liver being the segmentation target.
    The 20 test cases are evaluated using the online platform\footnote{\url{https://chaos.grand-challenge.org/}}.
    
    \paragraph{Application of nnU-Net to CHAOS}
    nnU-Net only supports images with a constant number of input modalities. The training cases in CHAOS have either one (T2) or two (T1 in \& out phase) modalities. To ensure compatibility with nnU-Net we could have either duplicated the T2 image and trained with two input modalities or use only one input modality and treat T1 in phase and out phase as separate training examples. We opted for the latter because this variant results in more (albeit highly correlated) training images. With 20 training patients being provided, this approach resulted in 60 training images.
    For the cross-validation we ensure that the split is being done on patient level.
    During inference, nnU-Net will generate two separate predictions for T1 in and out phase which need to be consolidated for test set evaluation. We achieve this by simply averaging the softmax probabilities between the two to generate the final segmentation.
    We train nnU-Net only for Task 5. Because task 3 represents a subset of Task 5, we extract the liver from our Task 5 predictions and submit it to Task 3.
    
    \textbf{Normalization:} Each image is normalized independently by subtracting its mean and dividing by its standard deviation.

    \begin{table}[H]
        \renewcommand{\arraystretch}{1.2}
        \centering
        \footnotesize
        \scalebox{0.9}{
        \begin{tabular}{llll}
            \textbf{}  & \textbf{2D U-Net}   & \textbf{3D full resolution U-Net}  & \textbf{3D low resolution U-Net}  \\
            \toprule 
            Target spacing (mm): & NA x 1.66 x 1.66  & 5.95 x 1.66 x 1.66  & -  \\
            \begin{tabular}[c]{@{}l@{}}Median image shape at \\ target spacing:\end{tabular} & NA x 195 x 262   & 45 x 195 x 262  & -  \\
            Patch size:  & 224 x 320   & 40 x 192 x 256  & -  \\
            Batch size:  & 45   & 2   & -   \\
            Downsampling strides: & \begin{tabular}[c]{@{}l@{}}{[}{[}2, 2{]}, {[}2, 2{]}, {[}2, 2{]}, {[}2, 2{]},\\ {[}2, 2{]}, {[}1, 2{]}{]}\end{tabular} & \begin{tabular}[c]{@{}l@{}}{[}{[}1, 2, 2{]}, {[}2, 2, 2{]}, {[}2, 2, 2{]}, \\ {[}2, 2, 2{]}, {[}1, 2, 2{]}, {[}1, 1, 2{]}{]}\end{tabular} & - \\
            Convolution kernel sizes: & \begin{tabular}[c]{@{}l@{}}{[}{[}3, 3{]}, {[}3, 3{]}, {[}3, 3{]}, {[}3, 3{]}, \\ {[}3, 3{]}, {[}3, 3{]}, {[}3, 3{]}{]}\end{tabular} & \begin{tabular}[c]{@{}l@{}}{[}{[}1, 3, 3{]}, {[}3, 3, 3{]}, {[}3, 3, 3{]}, \\ {[}3, 3, 3{]}, {[}3, 3, 3{]}, {[}3, 3, 3{]},\\ {[}3, 3, 3{]}{]}\end{tabular} & - \\[4pt]
        \end{tabular}}
        \caption{Network configurations generated by nnU-Net for the CHAOS challenge (D16). For more information on how to decode downsampling strides and kernel sizes into an architecture, see \ref{supplement:architecture_decoding}.}
    \end{table}

    \begin{table}[H]
        \renewcommand{\arraystretch}{1.2}
        \footnotesize
        \centering
        \begin{tabular}{llllll}
            & liver& right kidney& left kidney& spleen& mean\\
            \toprule 
            2D& 0.9132& 0.8991& 0.8897& 0.8720& 0.8935\\
            3D\_fullres & 0.9202& 0.9274& 0.9209& 0.8938& 0.9156\\
            Best Ensemble *& 0.9184& 0.9283& 0.9255& 0.8911& 0.9158\\
            Postprocessed & 0.9345 & 0.9289 & 0.9212 & 0.894 & 0.9197\\
            Test set & - & - & - & - & -\\[4pt] 
        \end{tabular}
        \caption[]{CHAOS results (D16). Note that all reported Dice scores (except the test set) were computed using five fold cross-validation on the training cases. Postprocessing was applied to the model marked with *. This model (incl postprocessing) was used for test set predictions. Note that the evaluation of the test set was performed with the online platform of the challenge which does not report Dice scores for the individual organs. The score of our submission was 72.44 for Task 5 and 75.10 for Task3 (see \citeSupplement{SM_kavur2020chaos} for details). Best ensemble on this dataset was the combination of the 2D U-Net and the 3D full resolution U-Net.}
    \end{table}

\subsection{Kidney and Kidney Tumor Segmentation (KiTS) (D17)}
    \paragraph{Challenge summary}
    The Kidney and Kidney Tumor Segmentation challenge \citeSupplement{SM_heller2019kits19} was the largest competition (in terms of number of participants) at MICCAI 2019. The target structures are the kidneys and kidney tumors. 210 training and 90 test cases are provided by the challenge organizers. The organizers provide the data both in their original geometry (with voxel spacing varying between cases) as well as interpolated to a common voxel spacing. Evaluation of the test set predictions is done on the online platform\footnote{\url{https://kits19.grand-challenge.org/}}. \\\\
    We participated in the original KiTS 2019 MICCAI challenge with a manually designed residual 3D U-Net. This algorithm, described in \citeSupplement{SM_isensee2019attempt} obtained the first rank in the challenge. For this submission, we did slight modifications to the original training data: Cases 15 and 37 were confirmed to be faulty by the challenge organizers (\url{https://github.com/neheller/kits19/issues/21}) which is why we replaced their respective segmentation masks with predictions of one of our networks. We furthermore excluded cases 23, 68, 125 and 133 because we suspected labeling errors in these cases as well. At the time of conducting the experiments for this publication, no revised segmentation masks were provided by the challenge organizers, which is why we re-used the modified training dataset for training nnU-Net.\\\\
    After the challenge event at MICCAI 2019, an open leaderboard was created. The original challenge leaderboard is retained at \url{http://results.kits-challenge.org/miccai2019/}. All submissions of the original KiTS challenge were mirrored to the open leaderboard. The submission of nnU-Net as performed in the context of this manuscript is done on the open leaderboard, where many more competitors have entered since the challenge. As presented in Figure \ref{fig:quantitative_results}, nnU-Net sets a new state of the art on the open leaderboard, thus also outperforming our initial, manually optimized solution.
    
    \paragraph{Application of nnU-Net to KiTS}
    Since nnU-Net is designed to automatically deal with varying voxel spacings within a dataset, we chose the original, non-interpolated image data as provided by the organizers and let nnU-Net deal with the homogenization of voxel spacing. nnU-Net was applied to the KiTS challenge without any manual intervention.

    \textbf{Normalization:} Clip to $[-79, 304]$, then subtract $100.93$ and finally divide by $76.90$.

    \begin{table}[H]
        \renewcommand{\arraystretch}{1.2}
        \centering
        \footnotesize
        \scalebox{0.9}{
        \begin{tabular}{llll}
            \textbf{}  & \textbf{2D U-Net}   & \textbf{3D full resolution U-Net}  & \textbf{3D low resolution U-Net}  \\
            \toprule 
            Target spacing (mm): & NA x 0.78 x 0.78  & 0.78 x 0.78 x 0.78  & 1.99 x 1.99 x 1.99  \\
            \begin{tabular}[c]{@{}l@{}}Median image shape at \\ target spacing:\end{tabular} & NA x 512 x 512   & 525 x 512 x 512  & 206 x 201 x 201  \\
            Patch size:  & 512 x 512   & 128 x 128 x 128  & 128 x 128 x 128  \\
            Batch size:  & 12   & 2   & 2   \\
            Downsampling strides: & \begin{tabular}[c]{@{}l@{}}{[}{[}2, 2{]}, {[}2, 2{]}, {[}2, 2{]}, {[}2, 2{]},\\ {[}2, 2{]}, {[}2, 2{]}, {[}2, 2{]}{]}\end{tabular} & \begin{tabular}[c]{@{}l@{}}{[}{[}2, 2, 2{]}, {[}2, 2, 2{]}, {[}2, 2, 2{]}, \\ {[}2, 2, 2{]}, {[}2, 2, 2{]}{]}\end{tabular} & \begin{tabular}[c]{@{}l@{}}{[}{[}2, 2, 2{]}, {[}2, 2, 2{]}, {[}2, 2, 2{]}, \\ {[}2, 2, 2{]}, {[}2, 2, 2{]}{]}\end{tabular} \\
            Convolution kernel sizes: & \begin{tabular}[c]{@{}l@{}}{[}{[}3, 3{]}, {[}3, 3{]}, {[}3, 3{]}, {[}3, 3{]}, \\ {[}3, 3{]}, {[}3, 3{]}, {[}3, 3{]}, {[}3, 3{]}{]}\end{tabular} & \begin{tabular}[c]{@{}l@{}}{[}{[}3, 3, 3{]}, {[}3, 3, 3{]}, {[}3, 3, 3{]}, \\ {[}3, 3, 3{]}, {[}3, 3, 3{]}, {[}3, 3, 3{]}{]}\end{tabular} & \begin{tabular}[c]{@{}l@{}}{[}{[}3, 3, 3{]}, {[}3, 3, 3{]}, {[}3, 3, 3{]}, \\ {[}3, 3, 3{]}, {[}3, 3, 3{]}, {[}3, 3, 3{]}{]}\\[4pt] \end{tabular}
        \end{tabular}}
        \caption{Network configurations generated by nnU-Net for the KiTS challenge (D17). For more information on how to decode downsampling strides and kernel sizes into an architecture, see \ref{supplement:architecture_decoding}}
    \end{table}

     \begin{table}[H]
        \renewcommand{\arraystretch}{1.2}
        \footnotesize
        \centering
        \begin{tabular}{llll}
            & Kidney& Tumor& mean\\
            \toprule 
            2D& 0.9613& 0.7563& 0.8588\\
            3D\_fullres& 0.9702& 0.8367& 0.9035\\
            3D\_lowres& 0.9629& 0.8420& 0.9025\\
            3D cascade& 0.9702& 0.8546& 0.9124\\
            Best Ensemble*& 0.9707& 0.8620& 0.9163\\
            Postprocessed& 0.9707& 0.8620& 0.9163\\[4pt]  
            Test set & -& 0.8542& -\\[4pt] 
        \end{tabular}
        \caption{KiTS results (D17). Note that all reported Dice scores (except the test set) were computed using five fold cross-validation on the training cases. Postprocessing was applied to the model marked with *. This model (incl postprocessing) was used for test set predictions. Note that the Dice scores for the test set are computed with the online platform which computes the kidney Dice score based of the union of the kidney and tumor labels whereas nnU-Net always evaluates labels independently, resulting in a missing value for kindey in the table. The reported kindey Dice by the platform (which is not comparable with the value computed by nnU-Net) is 0.9793. Best ensemble on this dataset was the combination of the 3D U-Net cascade and the 3D full resolution U-Net.}
    \end{table}

\subsection{Segmentation of THoracic Organs at Risk in CT images (SegTHOR) (D18)}

    \paragraph{Challenge summary}
    In the Segmentation of THoracic Organs at Risk in CT images \citeSupplement{SM_initialsegthorpaper} challenge, four abdominal organs (the heart, the aorta, the trachea and the esopahgus) are to be segmented in CT images. 40 training images are provided for training and another 20 images are provided for testing. Evaluation of the test images is done using the online platform\footnote{\url{https://competitions.codalab.org/competitions/21145}}.
    
    \paragraph{Application of nnU-Net to SegTHOR}
    nnU-Net was applied to the SegTHOR challenge without any manual intervention.
    
    \textbf{Normalization:} Clip to $[-986, 271]$, then subtract $20.78$ and finally divide by $180.50$.

    \begin{table}[H]
        \renewcommand{\arraystretch}{1.2}
        \centering
        \footnotesize
        \scalebox{0.9}{
        \begin{tabular}{llll}
            \textbf{}  & \textbf{2D U-Net}   & \textbf{3D full resolution U-Net}  & \textbf{3D low resolution U-Net}  \\
            \toprule 
            Target spacing (mm): & NA x 0.89 x 0.89  & 2.50 x 0.89 x 0.89  & 3.51 x 1.76 x 1.76  \\
            \begin{tabular}[c]{@{}l@{}}Median image shape at \\ target spacing:\end{tabular} & NA x 512 x 512   & 171 x 512 x 512  & 122 x 285 x 285  \\
            Patch size:  & 512 x 512   & 64 x 192 x 160  & 80 x 192 x 160  \\
            Batch size:  & 12   & 2   & 2   \\
            Downsampling strides: & \begin{tabular}[c]{@{}l@{}}{[}{[}2, 2{]}, {[}2, 2{]}, {[}2, 2{]}, {[}2, 2{]},\\ {[}2, 2{]}, {[}2, 2{]}, {[}2, 2{]}{]}\end{tabular} & \begin{tabular}[c]{@{}l@{}}{[}{[}1, 2, 2{]}, {[}2, 2, 2{]}, {[}2, 2, 2{]}, \\ {[}2, 2, 2{]}, {[}2, 2, 2{]}, {[}2, 2, 2{]}{]}\end{tabular} & \begin{tabular}[c]{@{}l@{}}{[}{[}2, 2, 2{]}, {[}2, 2, 2{]}, {[}2, 2, 2{]}, \\ {[}2, 2, 2{]}, {[}1, 2, 2{]}{]}\end{tabular} \\
            Convolution kernel sizes: & \begin{tabular}[c]{@{}l@{}}{[}{[}3, 3{]}, {[}3, 3{]}, {[}3, 3{]}, {[}3, 3{]}, \\ {[}3, 3{]}, {[}3, 3{]}, {[}3, 3{]}, {[}3, 3{]}{]}\end{tabular} & \begin{tabular}[c]{@{}l@{}}{[}{[}1, 3, 3{]}, {[}3, 3, 3{]}, {[}3, 3, 3{]}, \\ {[}3, 3, 3{]}, {[}3, 3, 3{]}, {[}3, 3, 3{]}{]}\end{tabular} & \begin{tabular}[c]{@{}l@{}}{[}{[}3, 3, 3{]}, {[}3, 3, 3{]}, {[}3, 3, 3{]}, \\ {[}3, 3, 3{]}, {[}3, 3, 3{]}, {[}3, 3, 3{]}{]}\\[4pt] \end{tabular}
        \end{tabular}}
        \caption{Network configurations generated by nnU-Net for the SegTHOR challenge (D18). For more information on how to decode downsampling strides and kernel sizes into an architecture, see \ref{supplement:architecture_decoding}}
    \end{table}
    
    \begin{table}[H]
        \renewcommand{\arraystretch}{1.2}
        \footnotesize
        \centering
        \begin{tabular}{llllll}
            & esophagus& heart& trachea& aorta& mean\\
            \toprule 
            2D& 0.8181& 0.9407& 0.9077& 0.9277& 0.8986\\
            3D\_fullres& 0.8495& 0.9527& 0.9055& 0.9426& 0.9126\\
            3D\_lowres& 0.8110& 0.9464& 0.8930& 0.9284& 0.8947\\
            3D cascade& 0.8553& 0.9520& 0.9045& 0.9403& 0.9130\\
            Best Ensemble*& 0.8545& 0.9532& 0.9066& 0.9427& 0.9143\\
            Postprocessed& 0.8545& 0.9532& 0.9083& 0.9438& 0.9150\\[4pt]  
            Test set  & 0.8890 & 0.9570 & 0.9228 & 0.9510 & 0.9300\\[4pt] 
        \end{tabular}
        \caption{SegTHOR results (D18). Note that all reported Dice scores (except the test set) were computed using five fold cross-validation on the training cases. Postprocessing was applied to the model marked with *. This model (incl postprocessing) was used for test set predictions. Note that the Dice scores for the test set are computed with the online platform. Best ensemble on this dataset was the combination of the 3D U-Net cascade and the 3D full resolution U-Net.}
    \end{table}

\subsection{Challenge on Circuit Reconstruction from Electron Microscopy Images (CREMI) (D19)}
\label{ref:supplement:d19}
    \paragraph{Challenge summary}
    The Challenge on Circuit Reconstruction from Electron Microscopy Images is subdivided into three tasks. The synaptic cleft segmentation task can be formulated as semantic segmentation (as opposed to e.g. instance segmentation) and is thus compatible with nnU-Net. In this task, the segmentation target is the cell membrane in locations where the cells are forming a synapse. The dataset consists of serial section Transmission Electron Microscopy scans of the \textit{Drosophila melanogaster} brain. Three volumes are provided for training and another three are provided for testing. Test set evaluation is done using the online platform\footnote{\url{https://cremi.org/}}.
    
    \paragraph{Application of nnU-Net to CREMI}
    Since to the number of training images is lower than the number of splits, we cannot run a 5-fold cross-validation. Thus, we trained 5 model instances, each of them on all three training volumes and subsequently ensembled these models for test set prediction. Because this training scheme leaves no validation data, selection of the best of three model configurations as performed by nnU-Net after cross-validation was not possible. Hence, we intervened by only configuring and training the 3D full resolution configuration. 
    
    \textbf{Normalization:} Each image is normalized independently by subtracting its mean and dividing by its standard deviation.

    \begin{table}[H]
        \renewcommand{\arraystretch}{1.2}
        \centering
        \footnotesize
        \scalebox{0.9}{
        \begin{tabular}{llll}
            \textbf{}  & \textbf{2D U-Net}   & \textbf{3D full resolution U-Net}  & \textbf{3D low resolution U-Net}  \\
            \toprule 
            Target spacing (mm):  & - &  40 x 4 x 4 & -   \\
            \begin{tabular}[c]{@{}l@{}}Median image shape at \\ target spacing:\end{tabular} &  - & 125 x 1250 x 1250  & -  \\
            Patch size:  & - & 24 x 256 x256  & -   \\
            Batch size:  & -   & 2   & -   \\
            Downsampling strides: & - & \begin{tabular}[c]{@{}l@{}}{[}{[}1, 2, 2{]}, {[}1, 2, 2{]}, {[}1, 2, 2{]}, \\ {[}2, 2, 2{]}, {[}2, 2, 2{]}, {[}1, 2, 2{]}{]}\end{tabular}  & -  \\
            Convolution kernel sizes: & - & \begin{tabular}[c]{@{}l@{}}{[}{[}1, 3, 3{]}, {[}1, 3, 3{]}, {[}1, 3, 3{]}, \\ {[}3, 3, 3{]}, {[}3, 3, 3{]}, {[}3, 3, 3{]},\\ {[}3, 3, 3{]}{]}\end{tabular} & - \\[4pt]
        \end{tabular}}
        \caption{Network configurations generated by nnU-Net for the CREMI challenge (D19). For more information on how to decode downsampling strides and kernel sizes into an architecture, see \ref{supplement:architecture_decoding}}
    \end{table}
    
    \paragraph{Results} Because our training scheme for this challenge left no validation data, a performance estimate as given for the other datastes is not available for CREMI. The CREMI test set is evaluated by the online platform. The evaluation metric is the so called CREMI score, a description of which is available here \url{https://cremi.org/metrics/}. Dice scores for the test set are not reported. The CREMI score of our test set submission was 74.96 (lower is better).

\section{Using nnU-Net with limited compute resources}
\setcounter{figure}{0} 
\setcounter{table}{0}
    Reduction of computational complexity was one of the key motivations driving the design of nnU-Net. The effort of running all the configurations generated by nnU-Net should be manageable for most users and researchers. There are, however, some shortcuts that can be be taken in case computational resources are extremely scarce. 
    
    \subsection{Reducing the number of network trainings}
    Depending on whether the 3D U-Net cascade is configured for a given dataset, nnU-Net requires 10 (2D and 3D U-Net with 5 models each) or 20 (2d, 3D, 3D cascade (low resolution and high resolution U-Net) with 5 models each) U-Net trainings to run, each of which takes a couple of days on a single GPU. While this approach guarantees the best possible performance, training all models may exceed reasonable computation time if only a single GPU is available. Therefore, we present two strategies to reduce the number of total network trainings when running nnU-Net.
    
    \textbf{Manual selection of U-Net configurations} \\
    Overall, the 3D full resolution U-Net shows the best segmentation results. Thus, this configuration is a good starting point and could simply be selected as default choice. Users can decide whether to train this configuration using all training cases (to train a single model) or run a five-fold cross-validation and ensemble the 5 resulting models for test case predictions.\\\\
    In some scenarios, other configurations than the 3D full resolution U-Net can yield best performance. Identifying such scenarios and selecting the respective most promising configuration, however, requires domain knowledge for the dataset at hand. Datasets with highly anisotropic images (such as D12 PROMISE12), for instance, could be best suited for running a 2D U-Net. There is, however, no guarantee for this relation (see D13 ACDC). On datasets with very large images, the 3D U-Net cascade seems to marginally outperform the 3D full resolution U-Net (for example D11, D14, D17, D18, ...) because it improves the capture of contextual information. Note that this is only true if the target structure requires a large receptive field for optimal recognition. On CREMI (D19) for example, despite large image sizes, only a limited field of view is required, because the target structure are relatively small synapses that can be identified using only local information, which is why we selected the 3D full resolution U-Net for this dataset (see Section \ref{ref:supplement:d19}).
    
    \textbf{Not running all configurations as 5-fold cross-validation} \\
    Another computation shortcut is to not run all models as 5-fold cross-validation. For instance, only one split for each configuration can be run (note, however, that the 3D low resolution U-Net of the cascade is required to be run as a 5-fold cross-validation in order to generate low resolution segmentation maps of all training cases for the second full resolution U-net of the cascade). Even when running multiple configurations to rely on empirical selection of configurations by nnU-Net, this reduces the total number of models to be trained to 2 if no cascade is configured or 8 if the cascade is configured (the cascade requires 6 model trainings: 5 3D low resolution U-Nets and 1 full resolution 3D U-Net training). nnU-Net subsequently bases selection of the best configuration on this single train-val split. Note that this strategy provides less reliable performance estimates and may result in sub optimal configuration choices. Finally, users can decide whether they wish to re-train the selected configuration on the entire training data or run a five-fold cross-validation for this selected configuration. The latter is expected to result in better test set performance because the 5 models can be used as an ensemble.
    
    \subsection{Reduction of GPU memory}
    nnU-Net is configured to utilize 11GB of GPU memory. This requirement is, based on our experience, a realistic requirement for a modern deep-learning capable GPU (such as a Nvidia GTX 1080 ti (11GB), Nvidia RTX 2080 ti (11GB), Nvidia TitanX(p) (12GB), Nvidia P100 (12/16 GB), Nvidia Titan RTX (24GB), Nvidia V100 (16/32 GB), ...). We strongly recommend using nnU-Net with this default configuration, because it has been tested extensively and, as we show in this manuscript, provides excellent segmentation accuracy. Should users still desire to run nnU-Net on a smaller GPU, the amount of GPU memory used for network configuration can be adapted easily. Corresponding instructions are provided along with the source code.

\bibliographystyle{abbrv}

\end{document}